\documentclass[sigconf, screen, table]{acmart}
\usepackage{pifont}
\AtBeginDocument{%
  }

\setcopyright{acmlicensed}
\copyrightyear{2025}
\acmYear{2025}
\acmDOI{XXXXXXX.XXXXXXX}
\acmConference[Conference acronym '25]{Make sure to enter the correct
  conference title from your rights confirmation email}{October 27–31,
  2025}{Dublin, Ireland}
\acmISBN{978-1-4503-XXXX-X/2018/06}

\renewcommand\footnotetextcopyrightpermission[1]{}

\acmSubmissionID{48}

\usepackage{hyperref} 

\usepackage{pifont}
\usepackage{graphicx}
\usepackage{wrapfig}
\usepackage{graphicx}
\usepackage{booktabs}
\usepackage{svg} 
\usepackage{colortbl}  

\usepackage{pifont,makecell}
\usepackage{multirow}
\usepackage{bm}
\usepackage{indentfirst}
\usepackage{threeparttable}
\usepackage{pifont}
\usepackage{arydshln}
\usepackage{arydshln}
\usepackage{tabularx}

\usepackage{makecell}

\usepackage{pifont}
\usepackage{minitoc}

\begin{document}

\title{Human-Activity AGV Quality Assessment: A Benchmark Dataset and an Objective Evaluation Metric}

\author{
    Zhichao Zhang$^{1}$, Wei Sun$^{1,}\text{*}$, Xinyue Li$^{1}$, Yunhao Li$^{1}$, Qihang Ge$^{1}$, Jun Jia$^1$, Zicheng Zhang$^1$, Zhongpeng Ji$^2$, Fengyu Sun$^2$, Shangling Jui$^2$, Xiongkuo Min$^1$, Guangtao Zhai$^{1,}\text{*}$ \\
    $^1$Shanghai Jiao Tong University, Shanghai, China \\
    $^2$Huawei Technologies, Shanghai, China
}
\thanks{\text{*}Corresponding author}

\renewcommand{\shortauthors}{Zhichao Zhang, et al.}


\begin{abstract}

AI-driven video generation techniques have made significant progress in recent years. However, AI-generated videos (AGVs) involving human activities often exhibit substantial visual and semantic distortions, hindering the practical application of video generation technologies in real-world scenarios. To address this challenge, we conduct a pioneering study on human activity AGV quality assessment, focusing on visual quality evaluation and the identification of semantic distortions. First, we construct the \underline{A}I-\underline{G}enerated \underline{Human} activity \underline{V}ideo \underline{Q}uality \underline{A}ssessment (\textbf{\textit{Human}-AGVQA}) dataset, consisting of $6,000$ AGVs derived from $15$ popular text-to-video (T2V) models using $400$ text prompts that describe diverse human activities. We conduct a subjective study to evaluate the \textbf{human appearance quality, action continuity quality, and overall video quality} of AGVs, and identify semantic issues of human body parts. Based on \textit{Human}-AGVQA, we benchmark the performance of T2V models and analyze their strengths and weaknesses in generating different categories of human activities. Second, we develop an objective evaluation metric, named \textit{AI-\underline{G}enerated \underline{H}uman activity \underline{V}ideo \underline{Q}uality metric} (\textbf{GHVQ}), to automatically analyze the quality of human activity AGVs. GHVQ systematically extracts human-focused quality features, AI-generated content-aware quality features, and temporal continuity features, making it a comprehensive and explainable quality metric for human activity AGVs. The extensive experimental results show that GHVQ outperforms existing quality metrics on the \textit{Human}-AGVQA dataset by a large margin, demonstrating its efficacy in assessing the quality of human activity AGVs. The \textit{Human}-AGVQA dataset and GHVQ metric will be released at \href{https://github.com/zczhang-sjtu/GHVQ.git}{https://github.com/zczhang-sjtu/GHVQ.git}.

\end{abstract}



\begin{teaserfigure}
  \includegraphics[width=\textwidth]{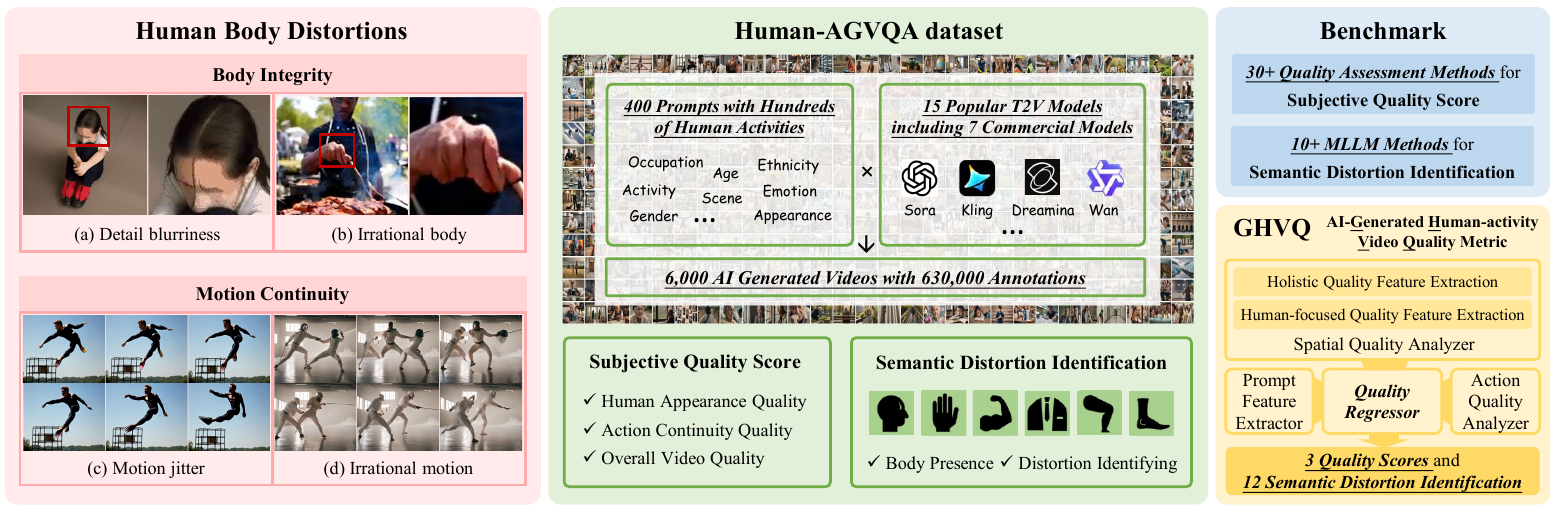}
  \caption{Overview of this work. This study aims to address the distortions in human activity AGVs, with some typical distortions shown in the first column. To achieve this goal, we introduce the Human-AGVQA dataset, consisting of $6,000$ human activity AGVs with quality and distortion identification labels, and we benchmark over $40$ quality assessment and MLLM methods on it. Finally, the proposed GHVQ metric objectively computes the quality of human activity AGVs.
  }
  \label{fig_distortion}
\end{teaserfigure}


\maketitle

\section{Introduction}
\label{sec:intro}

AI-driven video generation, particularly in the form of text-to-video (T2V) generation~\cite{videocrafter2,stablevideo}, can automatically create visual realism videos according to text descriptions. This convenient and cost-effective method of video creation has significant potential for applications in entertainment, art, advertising, education, and various other fields. However, many studies~\cite{VBench,FETV} indicate that current T2V models still struggle to generate realistic human figures and actions. For instance, AI-generated videos (AGVs) may exhibit distorted, incomplete, or abnormal body parts and actions, as illustrated in Figure~\ref{fig_distortion}. These artifacts can significantly affect the user's perceptual quality, as the human body is highly structured, and the human visual system is susceptible to abnormal body parts and irregular movement.

Therefore, accurately assessing the quality of human-activity AGVs is critical to automatically monitor the visual quality of large-scale AGVs in video generation applications, measure the progress of T2V models, and serve as an optimization or reward function to enhance the capability of T2V models. Unfortunately, the quality assessment for AI-generated content (AIGC)~\cite{EvalCrafter} is still in its infancy. The general image/video quality assessment (I/VQA) metrics have been shown to perform poorly in evaluating AGVs~\cite{MA_AGIQA, T2VQA}, while common used metrics in T2V studies, such as Inception Score (IS)~\cite{IS}, Fr\'{e}chet Inception Distance (FID)~\cite{FID}, and Fr\'{e}chet Video Distance (FVD)~\cite{FVD}, evaluate the generation performance and diversity of T2V models by comparing a set of AGVs to real data distributions in feature Inception feature spaces, thus failing to reflect the quality at the individual video level. 



To bridge this gap, we establish the first quality assessment benchmark for human activity AGVs, providing a comprehensive analysis of quality issues in human activity videos generated by mainstream T2V models and serving as a testbed for validating the effectiveness of quality assessment metrics in evaluating human activity AGVs. Specifically, we first construct the \underline{A}I-\underline{G}enerated \underline{Human} activity \underline{V}ideo \underline{Q}uality \underline{A}ssessment (\textbf{\textit{Human}-AGVQA}) dataset, which contains $6,000$ AGVs derived by $15$ state-of-the-art T2V models using $400$ text prompts. We define two practical quality assessment problems for human activity AGVs. The first is quality scoring, which quantifies the specific quality score of AGVs, allowing us to compare the quality of one AGV or T2V model against others. The second one is semantic issue identification, which pinpoints problematic body parts that affect quality scores, providing insights for further optimization of T2V models. Therefore, for the quality scoring task, we invite $80$ subjects to rate the quality scores of AGVs based on three critical dimensions: human appearance quality, action continuity quality, and overall visual quality. For the semantic issue identification task, we invite $5$ experts to label whether the human bodies (\textit{face, hands, arms, torso, legs, and feet}) exhibit semantic issues. Based on these subjectively rated quality labels, we can benchmark the performance of T2V models and analyze their strengths and weaknesses in generating different categories of human activities.

To address the lack of objective metrics for assessing the quality of human activity AGVs, we propose the \textit{AI-\underline{G}enerated \underline{H}uman activity \underline{V}ideo \underline{Q}uality metric} (\textbf{GHVQ}) to automatically evaluate the quality scores of human activity AGVs and identifies their semantic issues. Specifically, GHVQ consists of a spatial quality analyzer, an action quality analyzer, a text feature extractor, and a quality regressor. The spatial quality analyzer extracts human-focused and holistic quality features at the frame level. For human-focused features, we employ body-part segment masks to explicitly extract human body part-aware features, followed by an inner-body distortion analysis module and a cross-body distortion analysis module to refine these features at the individual body part level and the body part interaction level, respectively. Holistic quality features are extracted using a pre-trained AIGC IQA method to better represent complex AIGC artifiacts. The action quality analyzer captures the temporal continuity of AGVs using a pre-trained action recognition model. To ensure semantic alignment, the text feature extractor is employed to capture text-based semantic features. Finally, the four types of features are concatenated to form quality-aware feature representations. Two multi-layer perceptrons (MLPs) are used to simultaneously predict quality scores and binary decisions for each assessment dimension. Experimental results demonstrate that GHVQ outperforms existing metrics across all evaluated quality dimensions on the Human-AGVQA dataset, highlighting its effectiveness as a comprehensive metric for assessing the quality of complex human activity AGVs. We summarize the contributions of this paper as:
\begin{itemize}
\item We establish the \textit{Human}-AGVQA dataset for assessing the quality of human activity AGVs. \textit{Human}-AGVQA comprises $6,000$ videos generated by $15$ T2V models using $400$ text prompts, along with subjectively rated quality labels for $2$ types of assessment tasks. The diversity of text prompts and the richness of quality labels make our benchmark well-suited for investigating the quality assessment of human activity AGVs.
\item We propose the GHVQ metric to automatically evaluate the human appearance quality, action continuity quality, and overall video quality of AGVs, and identify whether six human body parts exhibit semantic issues. Experimental results indicate that GHVQ significantly outperforms existing related quality metrics, demonstrating its effectiveness. 
\end{itemize}

\section{Related Works}
\label{sec:Related Works}
\noindent\textbf{Video Generation Techniques} can be broadly categorized into three types: GAN/VAE-based methods~\cite{pan2017create,li2018video}, autoregressive-based methods~\cite{liang2022nuwa,hong2022cogvideo,villegas2022phenaki}, and diffusion-based methods~\cite{ho2022video,zeng2023make,chen2024videocrafter2}. Among these, diffusion-based methods have seen significant progress, with some commercial models~\cite{Gen2,stablevideo} already being developed or launched for real-world applications. However, several studies~\cite{sunsora,cho2024sora,FETV,VBench} point out that current video generation models still struggle with generating realistic human characteristics and actions, which emphasizes the need for quality assessment study in human activity AGVs.



\vspace{0.2cm}
\noindent\textbf{Video Quality Assessment} studies can be divided into two categories: knowledge-driven methods and data-driven methods. Knowledge-driven VQA methods leverage prior knowledge to extract quality features (\textit{e.g.,} natural scenes statistics features) for quality regression, such as TLVQM~\cite{TLVQM}, VIDEVAL~\cite{VIDEAL}, RAPIQUE~\cite{RAPIQUE}, etc. In contrast, data-driven BVQA methods automatically learn the quality-aware features by training a carefully designed deep neural network (DNN). Popular data-driven VQA methods generally \textbf{1)} use a 2D network to extract key frame features and a 3D network to capture motion features from video chunks, as seen in models like Li22~\cite{BVQA}, SimpleVQA~\cite{simpleVQA}, PatchVQA~\cite{PatchVQ}, MinimalisticVQA~\cite{minimalisticVQA}, etc., or \textbf{2)} employ a 3D network directly to extract video-level features from video chunks, such as FastVQA~\cite{FAST_VQA}. The 2D and 3D backbones can be pre-trained on other computer vision tasks~\cite{ying2020patches,he2016deep} or fine-tuned on VQA datasets in an end-to-end manner. Since current VQA research primarily focuses on natural videos, these methods often show limited performance when assessing the quality of AGVs~\cite{Survey_1}.

\begin{figure}
    \centering
    \includegraphics[width=0.99\linewidth]{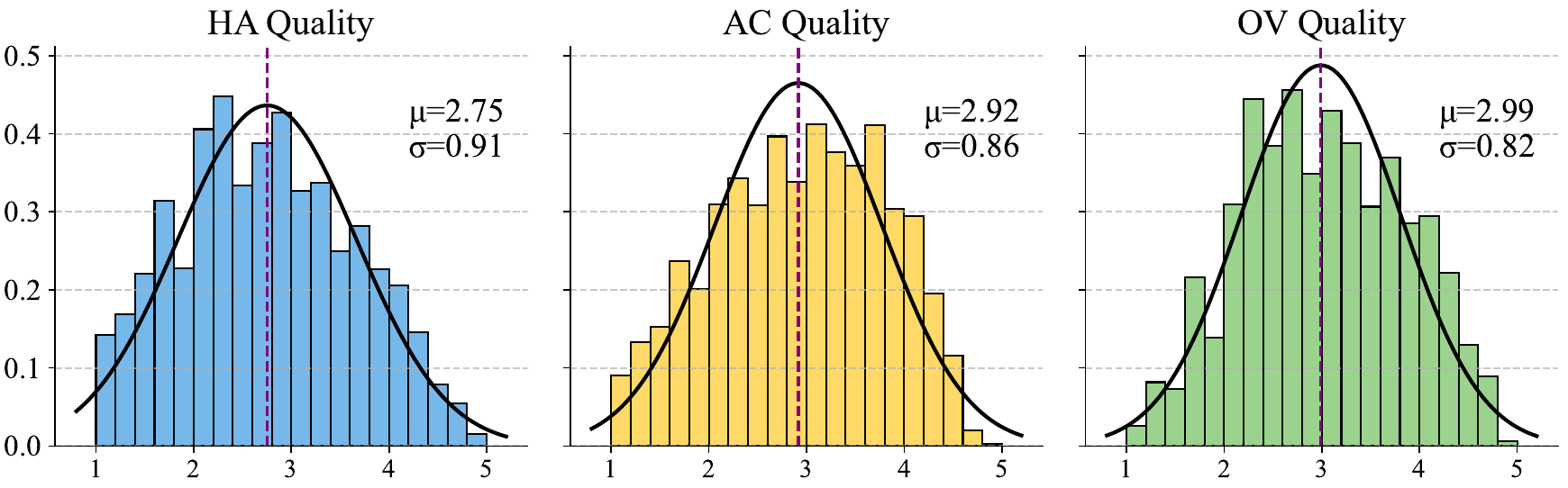}
    \caption{Illustration of the MOSs distribution. HA, AC, and OV quality represent human appearance, action continuity, and overall video, respectively. These abbreviations are used consistently throughout the paper.}
    \label{fig_MOS_density}
\end{figure}


\vspace{0.2cm}
\noindent\textbf{AGV Quality Assessment} has been explored in several studies~\cite{EvalCrafter,VBench,T2VQA}. For instance, FETV~\cite{FETV} benchmarks five T2V models based on static quality, temporal quality, overall alignment, and fine-grained alignment. EvalCrafter~\cite{EvalCrafter} evaluates 17 objective metrics, including visual quality, text-video alignment, motion quality, and temporal consistency, on a subjectively-rated AGV dataset. It also proposes a combined metric that leads to improved performance. VBench~\cite{VBench} introduces 16 additional dimensional metrics to evaluate the performance of T2V models. Moreover, \cite{T2VQA} constructs the T2VQA-DB, a large-scale text-to-video quality assessment dataset containing 10,000 videos, and develops the T2VQA metric, which focuses on the overall quality of AGVs. Despite these advancements, no specific metric has been developed to evaluate the quality of human activity AGVs.

\section{Human-AGVQA Dataset}
\label{sec:Human-AGVQA Dataset}




\subsection{Human activity AGVs Collection}



\noindent\textbf{Text Prompts Selection.} 
Text prompts describe the video content generated by T2V models. To comprehensively analyze the quality assessment problem of human activity AGVs, the selected text prompts in our dataset should cover a wide range of real-world human activities. Therefore, we classify the properties of these words into \textit{ages, genders, ethnicity, occupations, scenes, emotion, appearance, and activities} categories and further divide each category into $44$ subcategories that frequently occur in daily human life. 
We then randomly select one subcategory (\textit{e.g.}, \textbf{\textit{sports}}) from the \textbf{\textit{activity}} category, one subcategory (\textit{e.g.}, \textbf{\textit{outdoor}}) from the \textbf{\textit{scene}} category, and two or three subcategories from human appearance-based categories (\textit{e.g.}, \textbf{\textit{athlete}} from the \textbf{\textit{occupation}} category and \textbf{\textit{male}} from the \textbf{\textit{gender}} category). These keywords are combined into a text prompt using GPT-4~\cite{gpt4} with a prompt: 
\begin{quote}
\#  \textit{Please construct a complete sentence based on the following four concepts: "sports", "outdoor", "athlete", and "male". Note that the specific words "sports", "outdoor", "athlete", and "male" are not required to appear in the sentence, but the sentence should clearly convey the ideas related to these four categories.} \#. 
\end{quote}
\noindent We also use GPT-4 to evaluate the prompt's validity, rejecting any that are deemed unreasonable. Using this method, we generate $400$ distinct text prompts describing various human activities. 
The specific subcategories along with their proportions are listed in the supplementary material.


\begin{figure}
    \centering
    \includegraphics[width=0.99\linewidth]{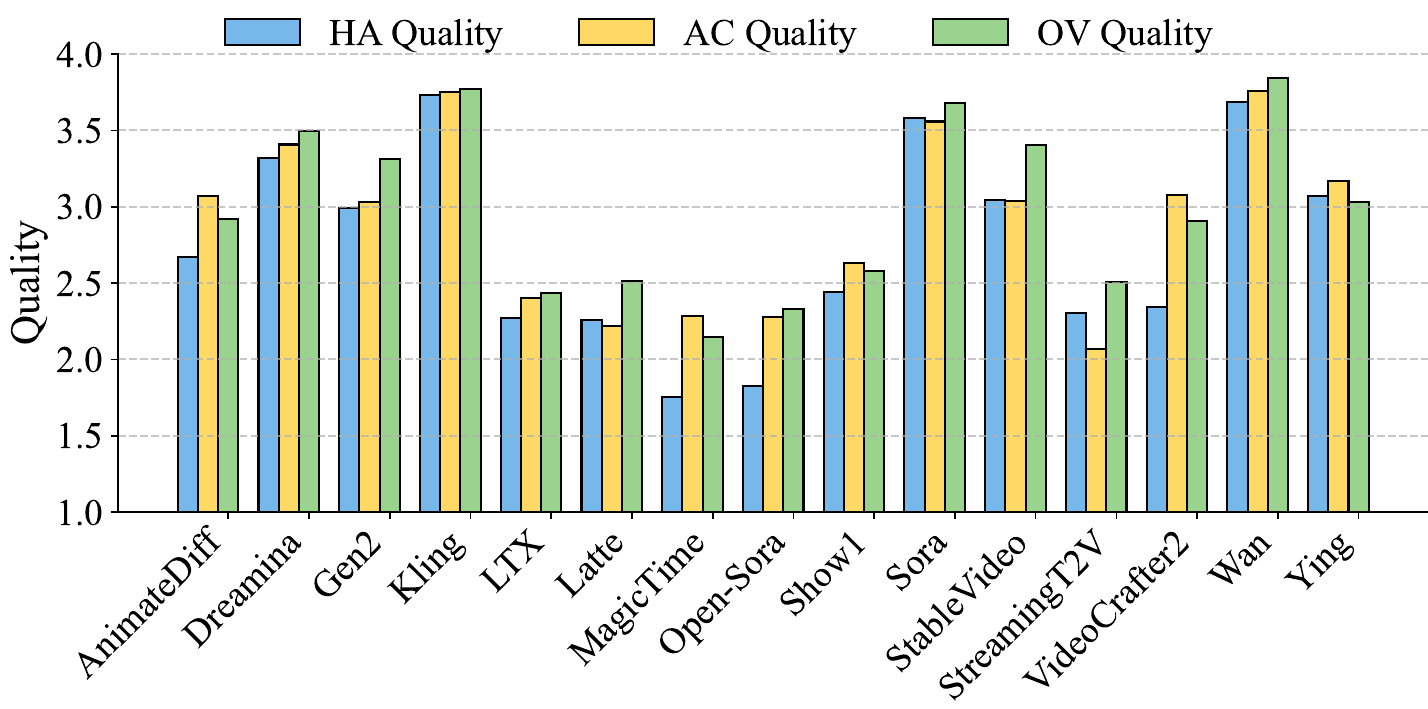}
    \caption{The MOS comparison for $15$ T2V models.}
    \label{fig_radar_overall}
\end{figure}

\vspace{0.2cm}
\noindent\textbf{T2V Models Selection.} 
We select $15$ state-of-the-art (SOTA) T2V models, including Gen-2~\cite{Gen2}, StableVideo~\cite{stablevideo}, StreamingT2V~\cite{StreamingT2V}, Video Crafter2~\cite{videocrafter2}, MagicTime~\cite{MagicTime}, Open Sora~\cite{Open_Sora_Plan}, Latte~\cite{Latte}, and AnimateDiff~\cite{AnimateDiff}, Sora\cite{sora} , Dreamina\cite{jimeng}, Kling\cite{kling}, Ying\cite{ying}, Wan\cite{wanxiang}, LTX-Video\cite{LTXVideo}, Show-1\cite{show1}to generate the video for each prompt. We summary the detailed information about these T2V models in the supplementary material. 

\begin{figure*}
    \centering
    \includegraphics[width=0.99\linewidth]{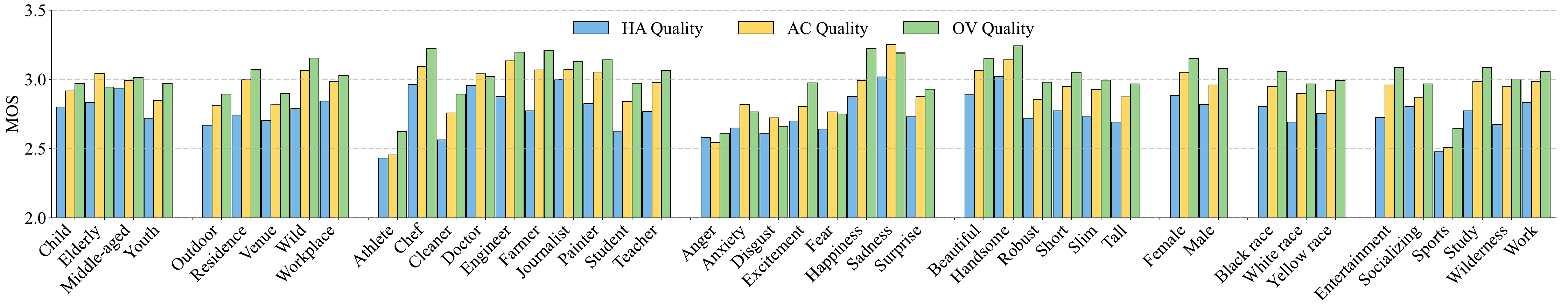}
    \caption{The MOS comparison for $15$ text categories and their $44$ subcategories.}
    \label{fig_bar}
\end{figure*}

In summary, a total of $6,000$ videos were generated by $15$ SOTA T2V models using $400$ diverse text prompts in the \textit{Human}-AGVQA dataset.

\subsection{Quality Assessment Dimensions}
We evaluate the quality of human activity AGVs from two perspectives:

\vspace{0.1cm}
\noindent\textbf{Visual Quality Scoring.}
We assess the quality of AGVs from three key dimensions: \textit{human appearance quality}, \textit{action continuity quality}, and \textit{overall video quality}. Specifically, human appearance quality assesses whether the human bodies in the video are complete, semantically correct, and aligned with the prompt description. Action continuity quality focuses on the coherence of the person's movements, the temporal consistency of limb actions, and whether the actions adhere to the laws of physical motion. Overall video quality evaluates the holistic video quality, including human appearance quality, action continuity quality, as well as background content quality. Participants rate each dimension on a scale from $1$ to $5$, where $1$ represents the lowest quality and $5$ represents the highest.


\vspace{0.1cm}
\noindent\textbf{Semantic Distortion Identification.}
Most quality metrics only provide a specific numerical value to quantify video quality, and cannot indicate which parts of the video are distorted or need further enhancement or optimization. This limitation restricts their applicability in quality artifact analysis and fine-grained video quality optimization. To address this, we label the semantic distorted human body parts in AGVs, providing a detailed explanation of quality scores. Specifically, we first annotate whether $6$ body parts (face, hands, arms, torso, legs, feet) are present in AGVs. Then, for each detected body part, we label whether it contains semantic artifacts. Finally, we provide $12$ binary labels per AGV for semantic distortion identification.

\subsection{Subjective Quality Experiment}
A total of $80$ subjects participated in the visual quality scoring experiment, with ages ranging from $20$ to $30$ years. The group included $46$ males and $34$ females. Given that the semantic artifact identification task is less complex than the visual quality scoring task, we invited $5$ experts in the field of AIGC quality assessment to perform the semantic artifact identification. $6,000$ videos were divided into $15$ groups, each containing $400$ videos that covered all 400 prompts, and each video was rated by $15$ subjects, and labeled by $5$ experts. In total, there are $270,000$ opinion scores and $360,000$ binary labels in the Human-AGVQA dataset. Additional details on the subjective experimental settings can be found in the supplemental material.


\subsection{Data Processing and Analysis}

For the quality scoring task, we follow the recommended method in ~\cite{Methodology} to process the raw subjective ratings into the mean opinion scores (MOSs).
For the distortion identification task, we use a voting method to determine body presence and body distortion. The label with the most votes is selected as the final result. The details about data processing and inter-subject consistency are shown in supplementary material.

\subsubsection{MOS Distribution Analysis}
Figure~\ref{fig_MOS_density} shows the MOS distributions for the three quality dimensions, which follow a Gaussian distribution. This indicates that medium-quality AGVs outnumber both high- and low-quality AGVs.
This suggests that the quality of human activity AGVs still needs improvement to meet visual quality standards. The average MOS for overall video quality is slightly higher than that of action continuity quality, and both are significantly higher than that of human appearance quality. This highlights that generating high-quality realistic humans remains a challenge for current T2V models, underscoring the importance of our study.

\begin{figure}
    \centering
    \includegraphics[width=0.99\linewidth]{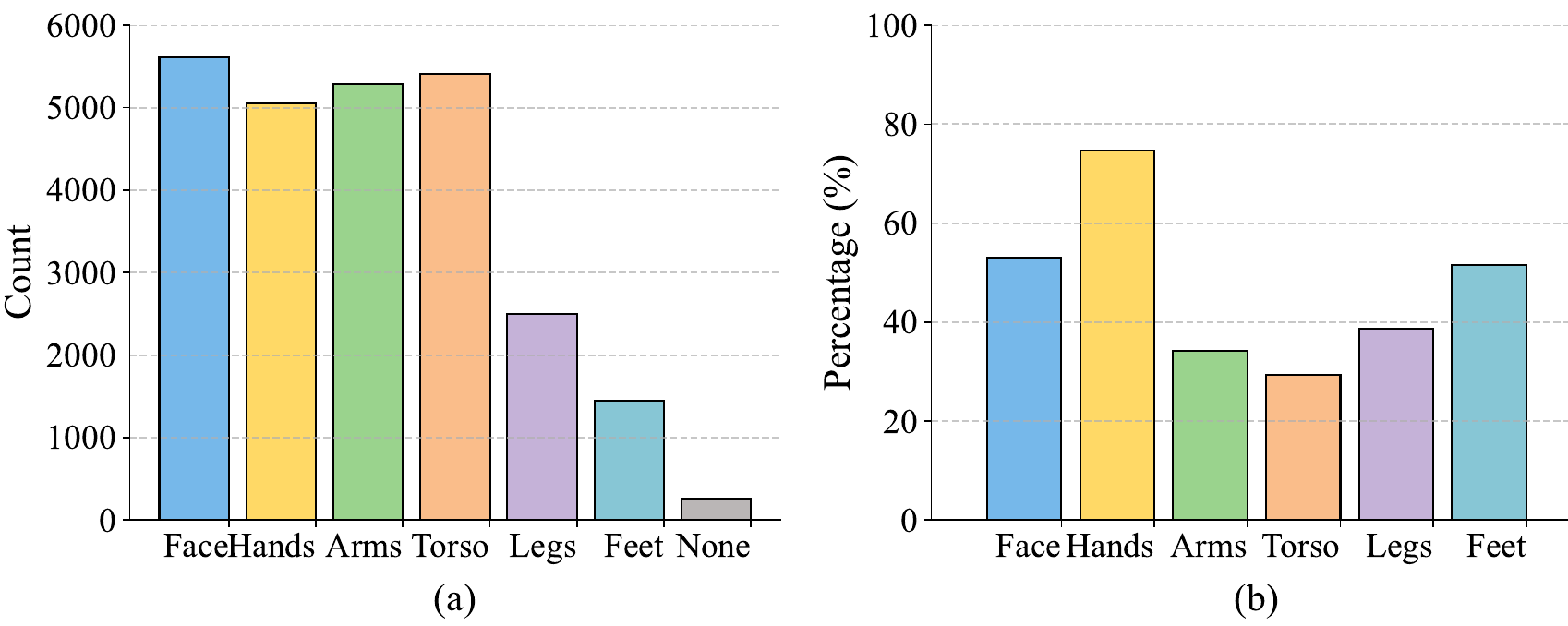}
    \caption{The percentage of AGVs that contain human body parts (a) and the percentage of these body parts exhibit semantic distortions (b).}
    \label{fig_proportion_body_part}
\end{figure}

\begin{figure*}
    \centering
    \includegraphics[width=0.99\linewidth]{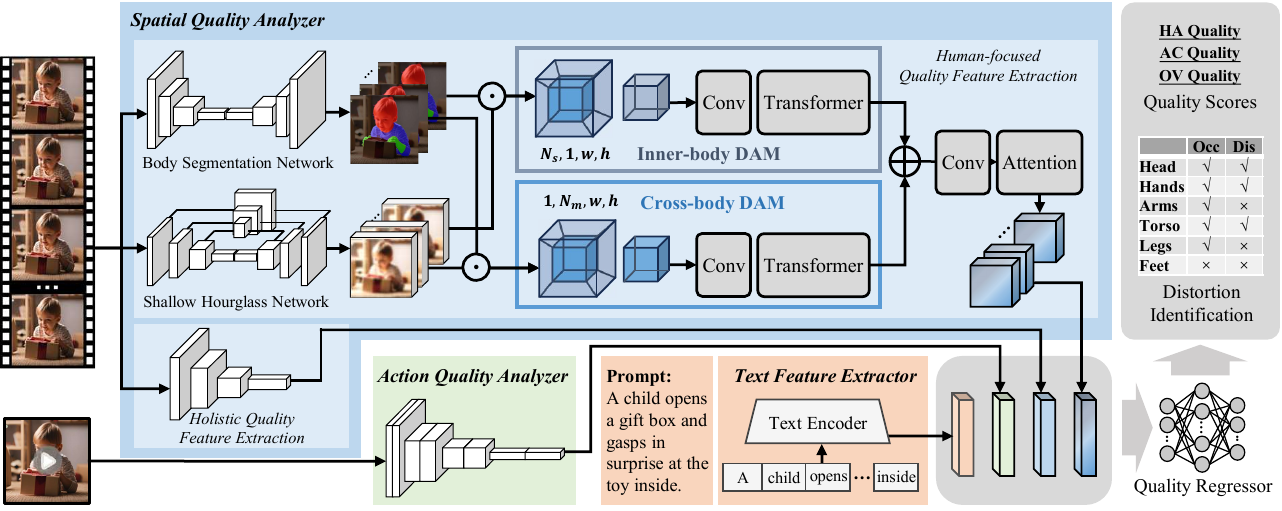}
    \caption{The GHVQ framework comprises four essential modules: a spatial quality analyzer for extracting human-focused and holistic quality features, an action quality analyzer for capturing temporal continuity of the body movement, a text feature extractor for identifying video content that needs to be generated by text prompts, and a quality regressor for mapping these quality-aware features to multi-dimensional quality scores and semantic distortion labels. }
    \label{fig_framework}
\end{figure*}

\subsubsection{MOS Analysis for T2V Models}

We calculate the average MOSs of $15$ T2V models across three quality dimensions to assess their strengths and weaknesses in generating human activity videos. The results are shown in Figure~\ref{fig_radar_overall}. 
Kling, Sora, and Wan are capable of producing highly realistic human appearances and action continuity with excellent overall quality, demonstrating strong generation quality. 
Dreamina, Ying, StableVideo, and Gen-2 also show strong performance. Notably, StableVideo and Gen2 exhibit significantly higher OV quality compared to the other two dimensions.
These above models are well-known commercial T2V tools, reinforcing the trend that commercial algorithms often surpass open-source models in terms of output quality.
LTX-Video, Show-1, AnimateDiff, and VideoCrafter2 show moderate performance across the dimensions, indicating that they have room for improvement.
Latte, MagicTime, OpenSora, and StreamingT2V have lower scores across all three quality dimensions, which suggests that these models may face challenges in generating high-quality videos.



\subsubsection{MOS Analysis for Text Prompts}

Figure~\ref{fig_bar} shows the average MOS for $8$ text prompt categories and their $44$ subcategories. Firstly, for the categories related to human attributes such as age, gender, and race, there is no significant difference in the quality scores among their subcategories, indicating that \textit{\textbf{the evaluated T2V models do not show bias toward human-related attributes}}. Secondly, video quality is generally lower in scenarios involving significant or frequent human movement, such as \textit{outdoor} in the scene category, \textit{athletes} and \textit{cleaners} in the occupation category, and \textit{sports} in the activity category. \textit{\textbf{This may be because frequent body movements cause the degradation of human appearance}}. Thirdly, for the emotion category, simpler emotions such as \textit{sadness} and \textit{happiness} receive relatively high scores, whereas more complex emotions like \textit{anger}, \textit{disgust}, and \textit{anxiety} are not as well-represented, \textit{\textbf{suggesting that current T2V models continue to face challenges in accurately understanding and representing complex emotions}}. Regarding the appearance category, \textit{handsome} and \textit{beautiful} yield satisfactory results. That may be that \textit{\textbf{the prompts like handsome and beautiful will make the generation model generate high-quality humans}}.

\subsubsection{Semantic Distortion Identification}

We counted all body parts appearing in the $6,000$ AGVs of the \textit{Human}-AGVQA dataset and the distortions observed for each body part, as shown in Figure~\ref{fig_proportion_body_part}.

In Figure~\ref{fig_proportion_body_part}(a), we observe that the face, hands, arms, and torso account for the largest proportions, while the legs and feet are significantly less than four parts. Notably, a small portion of the videos lack any visible body parts, indicating that some T2V models struggle to comprehend the text prompts and generate the corresponding human figures.
In Figure~\ref{fig_proportion_body_part}(b), it is evident that the torso exhibits the fewest distortions among the six body parts, whereas the hands, face and feet show more pronounced distortions. This suggests that current T2V models face significant challenges in generating complete and realistic body parts. The performance analysis of T2V models for distortion identification of human bodies can be found in the supplemental material.

\section{Proposed Model}

As illustrated in Figure~\ref{fig_framework}, the proposed AI-generated human activity video quality (GHVQ) method consists of four modules: a spatial quality analyzer, an action quality analyzer, a text feature extractor, and a quality regressor, which is detailed as follows:


\subsection{Spatial Quality Analyzer}
Assume a video $\bm{x}= \{\bm x_i\}_{i=0}^{N-1}$ generated by a text prompt $p$, where each frame $\bm x_i\in\mathbb{R}^{H\times W \times 3}$ denotes the $i$-th frame. Here, $H$ and $W$ are the height and width of each frame, and $N$ is the total number of frames. Given the substantial spatial redundancy between video frames, we first temporally downsample the video, $\bm x$, into a lower frame rate sequence, $\bm{y}= \{\bm y_i\}_{i=0}^{N_s-1}$, where $\bm{y}_i = \bm{x}_{\lfloor N/N_s\times i \rfloor}$, and $N_s$ denotes the total number of frames used for extracting spatial features.

We extract spatial quality features from two perspectives: human-focused and holistic, to more accurately evaluate the human appearance quality and overall video quality.

\subsubsection{Human-focused Quality Feature Extraction}
For frame ${\bm y}_i$, we first apply a human body-part segmentation model~\cite{Sapiens}, to explicitly detect the body-part masks $M_{i}$, where $M_{i} \in\mathbb{R}^{C_m \times H \times W}$ represents six segmentation masks corresponding to the face, arms, torso, legs, feet. Then, we design a shallow hourglass network~\cite{hourglass} to extract the high-resolution feature maps $F_{h,i}$ that preserve the resolution of human appearance while capturing rich low-level features. Here, $F_{h,i} \in\mathbb{R}^{C_h \times H \times W}$, with $C_h$ representing the number of channels of $F_{h,i}$. 

To focus on features specific to human body regions, we multiply the feature maps $F_{h,i}$ with the body masks $M_{i}$ along the channel dimensions to derive the human body-aware feature maps $F_{{\rm body},i}$:
\begin{align}
    F_{{\rm body},i} = F_{h,i} \cdot M_{i},
\end{align}
where $F_{{\rm body},i} \in\mathbb{R}^{C_m \times C_b \times H \times W}$. Next, we develop two modules---the inner-body distortion analysis module (Inner-body DAM) and the cross-body distortion analysis module (Cross-body DAM)---to further extract human-focused quality features at the individual body part level and the body part interaction level, respectively.

\vspace{0.2cm}
\noindent\textbf{Inner-body DAM.}
The inner-body DAM module is designed to capture the quality features related to each individual body part region. This allows the module to analyze each body part separately, helping to assess whether these body parts are presented in the video and which body parts exhibit quality issues. To achieve this, we aggregate the feature maps $F_{{\rm body},i}$ across the channel dimension $ C_h $:
\begin{equation}
    F_{{\rm inner},i} = \sum_{j=0}^{C_h-1} F_{{\rm body},i}[ , j:, :, :],
\end{equation}
where $F_{{\rm inner},i} \in \mathbb{R}^{ C_m \times H \times W}$ contains $C_m$ feature maps, each focusing on quality analysis for one of the $C_m$ body parts. 

\vspace{0.2cm}
\noindent\textbf{Cross-body DAM.} In contrast, the cross-body DAM module focuses on capturing interactive relationships between the body parts. Since a complete human action involves the coordinated movement of multiple body parts, jointly considering these interactions can better represent the quality of the human action.  To achieve this, we aggregate the feature maps $F_{{\rm body},i}$ across the channel dimension $ C_m $:
\begin{equation}
    F_{{\rm cross},i} = \sum_{j=0}^{C_m-1} F_{{\rm body},i}[ j, :, :, :],
\end{equation}
where $F_{{\rm cross},i} \in \mathbb{R}^{ C_s \times H \times W}$ contains $C_s$ feature maps focused on the quality analysis of all body parts.

\vspace{0.2cm}
\noindent\textbf{Feature Refinement and Fusion.}
For $F_{{\rm inner},i}$ and $F_{{\rm cross},i}$, we apply a CNN followed by a Transformer encoder~\cite{transformer} to further refine the inner-body and cross-body features:
\begin{equation}
\begin{aligned}
    F^{\prime}_{{\rm inner},i} = {\rm Transformer}({\rm Conv}(F_{{\rm inner},i})), \\
    F^{\prime}_{{\rm cross},i} = {\rm Transformer}({\rm Conv}(F_{{\rm cross},i})),
\end{aligned}
\end{equation}
where $\rm Transformer$ denotes the Transformer encoder, and the $\rm Conv$ represents a two-layer CNN. Finally, we concatenate them and then apply a CNN layer followed by a self-attention layer~\cite{self_attention} to derive the human-focused quality features:
\begin{equation}
\begin{aligned}
    F_{{\rm bq},i} &= {\rm Attention}({\rm Conv}([F^{\prime}_{{\rm inner},i}, F^{\prime}_{{\rm cross},i}])),
\end{aligned}
\end{equation}
where $\rm Attention$ denotes the self-attention operator and $F_{{\rm bq},i}$ is the human-focused quality features.

\subsubsection{Holistic Quality Feature Extraction}
To extract quality-aware features of the entire frame, we pre-train a ViT~\cite{vit} model on Pick-a-Pic~\cite{PickScore}, a large-scale text-to-image (T2I) quality assessment dataset, to learn the quality-aware feature representation for AIGC images. Then, for frame ${\bm y}_i$, we employ the pre-trained ViT~\cite{vit} as the feature extractor to compute the holistic quality features:
\begin{align}
    F_{{\rm hq},i} = {\rm{ViT}}({\bm y}_i).
\end{align}

\begin{table*}[t]
\scriptsize
\centering
\caption{The performance comparison of existing IQA-based, AQA-based, and VQA-based metrics with our proposed GHVQ metrics on the \textit{Human}-AGVQA dataset. NSC refers to natural scene content.}
\label{tab_benchmark}
\resizebox{0.95\textwidth}{!}{
\begin{tabular}{clllllll}
\toprule
\multirow{2}{*}{Quality Dimension}   & \multicolumn{1}{l}{\multirow{2}{*}{Methods}}   & \multicolumn{1}{l}{\multirow{2}{*}{\makecell[l]{Pre-training /\\Initialization}}} & \multicolumn{1}{l}{\multirow{2}{*}{\makecell[l]{Model\\Type}}}  & \multicolumn{2}{c}{Zero-shot}  & \multicolumn{2}{c}{Fine-tuning} \\
                                                                                                                                                                 \cmidrule(r){5-6}             \cmidrule(r){7-8}   
                           &                                                &                                                 &                                  & SRCC    & PLCC              & SRCC   & PLCC    \\
\midrule                                                                                                                                                                                                              
\multirow{12}{*}{Human Appearance} & NIQE (ISPL, 2012)~\cite{NIQE}          & NA (\textit{handcraft})                         & NSC                              & 0.233   & 0.273             & 0.564  & 0.570   \\
                           & BRISQUE (TIP, 2012)~\cite{BRISQUE}             & NA (\textit{handcraft})                         & NSC                              & 0.258   & 0.314             & 0.594  & 0.619   \\
                           & CNNIQA (CVPR, 2014)~\cite{CNNIQA}              & TID2013~\cite{TID2013}                          & NSC                              & 0.243   & 0.296             & 0.620  & 0.646   \\
                           & HyperIQA (CVPR, 2020)~\cite{HyperIQA}          & TID2013~\cite{TID2013}                          & NSC                              & 0.318   & 0.339             & 0.684  & 0.696   \\
                           & UNIQUE (TIP, 2021)~\cite{UNIQUE}               & KonIQ-10K~\cite{KonIQ_10k}                      & NSC                              & 0.239   & 0.280             & 0.668  & 0.684   \\
                           & MUSIQ (ICCV, 2021)~\cite{MUSIQ}                & KonIQ-10K~\cite{KonIQ_10k}                      & NSC                              & 0.224   & 0.262             & 0.583  & 0.601   \\
                           & StairIQA (JSTSP, 2023)~\cite{StairIQA}         & KonIQ-10K~\cite{KonIQ_10k}                      & NSC                              & 0.339   & 0.393             & 0.661  & 0.663   \\
                           & CLIP-IQA (AAAI, 2023)~\cite{CLIP_IQA}          & KonIQ-10K~\cite{KonIQ_10k}                      & NSC                              & 0.327   & 0.368             & 0.672  & 0.674   \\
                           & LIQE (CVPR, 2023)~\cite{LIQE}                  & KonIQ-10k~\cite{KonIQ_10k}                      & NSC                              & 0.283   & 0.332             & 0.640  & 0.656   \\
                           & MA-AGIQA (ACMMM, 2024)~\cite{MA_AGIQA}         & AGIQA-3k~\cite{AGIQA}                           & AIGC                             & 0.373   & 0.380             & 0.724  & \underline{0.733} \\
                           & Q-Align (ICML, 2024)~\cite{qalign}             & \textit{fused} ~\cite{KonIQ_10k},~\cite{SPAQ},~\cite{PatchVQ},~\cite{Ava} & NSC    & 0.362   & 0.419             & \underline{0.725}  & 0.731 \\
                           & GHVQ (\textit{proposed})                       & ---                                             & AIGC                             & ---       & ----            & \textbf{0.805} & \textbf{0.809} \\
\midrule                                                                                                                                                                                                            
\multirow{9}{*}{Action Continuity} & ACTION-NET (ACMMM, 2020)~\cite{ACTION_NET} & Kinetics~\cite{kinetics}                    & NSC                              & 0.198   & 0.223            & 0.541  & 0.553   \\
                           & USDL (CVPR, 2020)~\cite{USDL}                  & Kinetics~\cite{kinetics}                        & NSC                              & 0.208   & 0.259             & 0.583  & 0.591   \\
                           & CoRe (ICCV, 2021)~\cite{CoRe}                  & Kinetics~\cite{kinetics}                        & NSC                              & 0.177   & 0.210             & 0.562  & 0.577   \\
                           & TSA (CVPR, 2022)~\cite{TSA}                    & Kinetics~\cite{kinetics}                        & NSC                              & 0.204   & 0.256             & 0.602  & 0.613   \\
                           & Motion Smoothness (CVPR, 2024)~\cite{VBench}   & AMT~\cite{Amt}                                  & AIGC                             & 0.250   & 0.275             & ---      & ---   \\   
                           & Temporal Flickering (CVPR, 2024)~\cite{VBench} & RAFT~\cite{RAFT}                                & AIGC                             & 0.137   & 0.239             & ---      & ---   \\
                           & Action-Score (CVPR, 2024)~\cite{EvalCrafter}   & VideoMAE V2~\cite{Videomae_v2}                  & AIGC                             & 0.209   & 0.244             & ---      & ---   \\
                           & Flow-Score (CVPR, 2024)~\cite{EvalCrafter}     & RAFT~\cite{RAFT}                                & AIGC                             & 0.254   & 0.279             & ---      & ---   \\
                           & GHVQ (\textit{proposed})                       & ---                                             & AIGC                             & ---       & ---             & \textbf{0.771} & \textbf{0.778} \\
\midrule                                                                                                                                                                                                                  
\multirow{13}{*}{Overall Video} & TLVQM (TIP, 2019)~\cite{TLVQM}            & NA (\textit{handcraft})                         & NSC                              & 0.272   & 0.312             & 0.603  & 0.617   \\
                           & RAPIQUE (JSP, 2021)~\cite{RAPIQUE}             & NA (handcraft)                                  & NSC                              & 0.313   & 0.351             & 0.621  & 0.637   \\
                           & VIDEAL (TIP, 2021)~\cite{VIDEAL}               & NA (handcraft)                                  & NSC                              & 0.342   & 0.353             & 0.628  & 0.642   \\
                           & PatchVQ (CVPR, 2021)~\cite{PatchVQ}            & LSVQ~\cite{PatchVQ}                             & NSC                              & 0.379   & 0.399             & 0.657  & 0.693   \\
                           & SimpleVQA (ACMMM, 2022)~\cite{simpleVQA}       & LSVQ~\cite{PatchVQ}                             & NSC                              & 0.364   & 0.378             & 0.687  & 0.698   \\
                           & BVQA (TCSVT, 2022)~\cite{BVQA}                 & KoNViD-1k~\cite{KoNViD_1k}                      & NSC                              & 0.349   & 0.374             & 0.672  & 0.702   \\
                           & FastVQA (ECCV, 2023)~\cite{FAST_VQA}           & LSVQ~\cite{PatchVQ}                             & NSC                              & 0.390   & 0.412             & 0.698  & 0.711   \\
                           & DOVER (ICCV, 2023)~\cite{dover}                & DIVIDE~\cite{dover}                             & NSC                              & 0.312   & 0.337             & 0.705  & 0.717   \\
                           & T2VQA (ACMMM, 2024)~\cite{T2VQA}               & T2VQA~\cite{T2VQA}                              & AIGC                             & 0.359   & 0.367             & \underline{0.737}  & 0.742 \\
                           & UGVQ (Arxiv, 2024)~\cite{UGVQ}                 & LGVQ~\cite{UGVQ}                                & AIGC                             & 0.349   & 0.358             & 0.734  & \underline{0.743} \\
                           & EvalCrafter (CVPR, 2024)~\cite{EvalCrafter}    & DIVIDE~\cite{dover}                             & AIGC                             & 0.328   & 0.336             & ---      & ---        \\
                           & Q-Align (ICML, 2024)~\cite{qalign}             & fused ~\cite{KonIQ_10k},~\cite{SPAQ},~\cite{PatchVQ},~\cite{Ava} & AIGC            & 0.422   & 0.481             & 0.715  & 0.723        \\
                           & GHVQ (\textit{proposed})                       & ---                                               & AIGC                           & ---       & ---             & \textbf{0.768}  & \textbf{0.773} \\
\bottomrule
\end{tabular}
}
\end{table*}

Finally, we aggregate the human-focused features and the holistic quality features as the spatial quality features, then average them across all sampled frames:
\begin{equation}
\begin{aligned}
    F_{{\rm sq},i} &= [F_{{\rm bq},i} , F_{{\rm hq},i}], \\
    F_{{\rm sq}} &= \sum^{Ns-1}_{i=0}F_{{\rm sq},i},
\end{aligned}
\end{equation}
where $F_{{\rm sq},i}$ and $F_{{\rm sq}}$ represent the spatial quality features of frame ${\bm y}_i$ and video sequence ${\bm y}$.

\subsection{Action Quality Analyzer}
The action quality analyzer is designed to quantify the temporal continuity of videos, serving as a complement to the spatial quality analyzer in assessing comprehensively the quality of AGVs. To address this, we utilize an action recognition model, as the action quality analyzer to capture the temporal continuity of human body movement. Since action recognition models are trained on large-scale, authentic-captured human action datasets, such as the Kinetics series~\cite{kinetics}, they can effectively extract motion representations of human bodies. Moreover, previous VQA studies~\cite{simpleVQA} have also demonstrated that these features are useful for video quality evaluation. Specifically, for video $\bm x$ and the action recognition network $\rm{SlowFast}$~\cite{SlowFast}, we calculate the action quality features $F_{\rm aq}$ using all frames in $\bm x$.

\subsection{Text Feature Extractor}
Since the purpose of T2V models is to generate video content that aligns with the text prompt, it is necessary to measure the alignment between the video content and the text prompt. Following most T2I and T2V alignment evaluation studies~\cite{VBench,MA_AGIQA}, we utilize the text encoder of CLIP~\cite{PickScore} to extract semantic features $F_{\rm tq}$ of the text prompt $p$.

\subsection{Quality Regressor}

The quality regressor is used to assess the three quality scores and determine whether human bodies are present and if there exist semantic distortions. So, we first integrate the spatial quality features, the action quality features, and the text features into the final quality-aware features $F_q$:
\begin{equation}
  \label{eq_f_q}
    F_q = [F_{\rm sq}, F_{\rm aq}, F_{\rm tq}].
\end{equation}
Then, $F_q$ are fed into two multi-layer perceptron (MLP) to produce the quality scores $\hat{q}$ and distortion labels $\hat{b}$.

The loss function consists of the mean absolute error (MAE) loss, rank loss~\cite{rank_loss}, and binary cross-entropy loss. The MAE loss and rank loss are used to optimize the quality scoring task, and the binary cross-entropy loss is used to optimize the distortion identification task.


\section{Experiments}

\subsection{Experiment Settings}

\noindent\textbf{Compared Quality Metrics.}
Since no specific metrics have been designed to evaluate human appearance quality, we compared $11$ IQA models to assess the quality of video frames as a proxy for human appearance. For action continuity quality and overall video quality, we select $8$ action quality assessment (AQA) and $12$ VQA methods as comparison respectively. For body parts detection and their distortion identification, there are also no specific metrics, we benchmark $11$ general video multi-modal large multi-modality (MLMM) model for this task. A detailed introduction to these methods can be found in the supplemental material.




\vspace{0.1cm}
\noindent\textbf{Dataset Splits.}
We split the \textit{Human}-AGVQA dataset into $70$\% for training, $10$\% for validation, and $20$\% for test. For video quality scoring, the compared metrics are evaluated on the test set using two approaches: zero-shot testing and fine-tuning testing. For body parts detection and distortion identification, we directly test zero-shot performance of compared MLMMs.

\vspace{0.1cm}
\noindent\textbf{Evaluation Criteria.}
We use PLCC and SRCC to evaluate the performance of the test metrics. Additionally, we use classification accuracy to assess the performance of body part detection and distortion identification.

Other experimental settings including the training details of GHVQ can be found in the supplemental material.

\subsection{Performance Comparison}

\subsubsection{Quality Scoring Task}
\noindent\textbf{Human Appearance Quality.} We observe that all compared IQA methods perform poorly in evaluating human appearance quality in the zero-shot setting, which highlights that distortions in human appearance in AGVs represent entirely new artifacts that existing metrics cannot handle. Fine-tuning on the \textit{Human}-AGVQA dataset improves the performance of data-driven methods, but their results remain suboptimal. This may be because these methods focus on learning feature representations from global images, rather than human-focused regions. The proposed GHVQ achieves the best performance, outperforming the second-best method by $0.805$ and $0.809$ in terms of SRCC and PLCC, demonstrating the effectiveness of its design for extracting human appearance-focused features.

\begin{table*}[]
\caption{The accuracy of MLMM models in identifying the occurrences and distortions of body parts in the \textit{Human}-AGVQA dataset.}
\label{tab_acc_of_LMM}
\resizebox{0.97\textwidth}{!}{
\begin{tabular}{l|c|c|c|c|c|c|c}
\toprule
\multirow{2}{*}{Models}                 & \multicolumn{7}{c}{Occurrences / Distortions}                                                                                                                                                                 \\
                                        \cmidrule(r){2-8}                                                                                                                                                                                               
                                        & Face                      & Hands                     & Arms                      & Torso                     & Legs                       & Feet                      & Avg             \\
\midrule                                                                                                                                                                                                                             
LLaMAVID (7B)~\cite{LLaMA_VID}          & 83.25 / 38.24             & 26.94 / 31.99             & 68.61 / 36.00             & 81.72 / 46.40             & 57.91 / 30.60              & 24.99 / 21.39             & 57.24 / 34.10   \\
VideoChatGPT (7B)~\cite{Video_chatgpt}  & 67.11 / 61.58             & 20.09 / \underline{51.13} & 51.75 / 49.65             & 64.52 / 37.54             & 41.36 / 22.93              & 29.11 / 19.26             & 45.66 / 40.35   \\
VideoLLaMA2 (7B)~\cite{VideoLLaMA2}     & 26.70 / 24.77             & 20.64 / 14.82             & 21.64 / 20.29             & 54.41 / 31.13             & 21.74 / 45.72              & 30.33 / 48.24             & 29.24 / 30.83   \\
VILA1.5 (7B)~\cite{VILA}                & 35.65 / 26.52             & 23.46 / 22.39             & 29.79 / 27.38             & 57.82 / 30.63             & 27.52 / 41.63              & 31.19 / 46.66             & 34.24 / 32.54   \\
NeXT-Video (7B)~\cite{Llava_next}       & 60.97 / 42.56             & 35.96 / 27.21             & 45.24 / 31.18             & 64.48 / 51.80             & 32.22 / 34.85              & 45.23 / 57.53             & 47.35 / 40.86   \\
OneVison (7B)~\cite{Llava_onevision}    & 62.21 / 44.06             & 39.21 / 39.05             & 46.17 / 33.50             & 66.90 / 53.13             & 33.27 / 40.93              & 47.11 / 50.17             & 49.15 / 43.47   \\
Qwen2-VL (7B)~\cite{qwen2_vl}           & 72.90 / 52.92             & 47.88 / 31.59             & 59.08 / 36.83             & 71.03 / 55.43             & 47.12 / 65.97              & 54.01 / 64.84             & 58.67 / 51.26   \\
Qwen2.5-VL (7B)~\cite{qwen2.5_vl}       & 87.66 / \underline{68.38} & 65.15 / 38.79             & 72.13 / 47.18             & 81.81 / 63.31             & 59.54 / 71.04              & 69.00 / \underline{72.77} & 72.55 / \underline{60.25} \\
DeepSeek-VL2 (4.1B)~\cite{deepseekvl2}  & 71.80 / 53.56             & 44.81 / 33.38             & 57.45 / \underline{51.60} & 82.67 / 50.67             & 43.89 / 62.97              & 60.95 / 52.76             & 60.26 / 50.82   \\
GPT-4o ~\cite{gpt4}                     & \underline{92.33} / 29.63 & \underline{82.28} / 24.95 & \textbf{88.51} / 19.23    & \underline{84.95} / 60.31 & 62.04 / \underline{73.31}  & \underline{75.96} / 71.02 & \underline{81.01} / 46.41 \\
GPT-4o-mini ~\cite{gpt4}                & 90.50 / 28.29             & 80.35 / 22.60             & 86.41 / 20.03             & 83.55 / \underline{61.49} & \underline{62.52} / 72.24  & 73.94 / 69.16             & 79.55 / 45.64   \\
\midrule                                                                                                                                                                                                                            
GHVQ (\textit{proposed})  & \textbf{92.99 / 78.36}  & \textbf{84.78} / \textbf{62.86}  & \underline{88.50} / \textbf{70.25}  & \textbf{90.23 / 73.13}  & \textbf{64.26 / 83.71} & \textbf{77.37 / 87.35}  & \textbf{83.02 / 75.94} \\
\bottomrule
\end{tabular}
}
\end{table*}

\vspace{0.2cm}
\noindent\textbf{Action Continuity Quality.} Similar to human appearance quality, the compared AVA methods and temporal metrics of AIGC VQA methods show a very low correlation in assessing action continuity quality. This is because AVA methods are typically designed to quantify the completeness of professional sports performances, while the temporal metrics in AIGC VQA methods rely on handcrafted motion descriptors that do not capture semantic content, such as those involving individuals. In the fine-tuning results, GHVQ outperforms the re-trained AVA methods, surpassing the second-best method by $0.771$ and $0.778$ in terms of SRCC and PLCC, demonstrating that the proposed model architecture is better suited to measure action continuity quality.

\vspace{0.2cm}
\noindent\textbf{Overall Video Quality.} The performance of VQA metrics in assessing overall video quality is slightly superior to that of other metrics for human appearance and action continuity quality. Generally, we find that AIGC-based VQA methods outperform general data-driven VQA methods, and both of these are more effective than knowledge-driven VQA methods. Similarly, re-training these methods leads to significant improvements. Since GHVQ comprehensively extracts human-focused, action continuity, and holistic frame quality features, the proposed GHVQ outperforms existing VQA methods, improving performance by $0.768$ and $0.773$ in terms of SRCC and PLCC.

\begin{table}
\caption{Ablation study results for different modules across quality metrics. SQA, AQA, and TFE refer to the spatial quality analyzer, action quality analyzer, and text feature extractor, respectively. The results for the three quality scores are reported using SRCC/PLCC, while distortion identification is presented in terms of occurrence and distortion accuracy.}
\label{tab_ablation1}
\resizebox{0.49\textwidth}{!}{
\begin{tabular}{ccc|cccc}
\toprule
 SQA         & AQA          & TFE         & HA Quality     &  AC Quality    &  OV Quality    &  \makecell[c]{Distortion\\Identification} \\
 \midrule                                                                                                       
             & \ding{52}    & \ding{52}   & 0.625 / 0.631  & 0.611 / 0.626  & 0.717 / 0.730  & 58.36 / 52.16   \\ 
 \ding{52}   &              & \ding{52}   & 0.762 / 0.767  & 0.741 / 0.742  & 0.721 / 0.734  & 75.94 / 71.08   \\ 
 \ding{52}   & \ding{52}    &             & 0.774 / 0.782  & 0.752 / 0.761  & 0.760 / 0.765  & 81.02 / 73.91   \\ \midrule 
 \ding{52}   & \ding{52}    & \ding{52}   & 0.805 / 0.809  & 0.771 / 0.778  & 0.768 / 0.773  & 83.02 / 75.94   \\ 
\bottomrule 
\end{tabular}
}
  \centering
\end{table}

\subsubsection{Body Presence and Distortion Identifying Task}
Identifying the presence and distortions of human body parts is a novel task, and we benchmark the performance of MLMM models for this task in Table~\ref{tab_acc_of_LMM}. The key prompts that we utilized for the evaluation are listed below. Note that replace \textit{\textbf{[body part]}} with one of the following: "face", "hands", "arms", "torso", "legs", or "feet". More details are shown in the supplementary material.

For body part occurrences:
\#  \textit{Is there a person's \textbf{[body part]} appearing in this video? Answer with "Yes" or "No".} \#. 

For body part distortions:
\#  \textit{Is there any incompleteness, unrealistic appearance, or discontinuous movements of the person's \textbf{[body part]} in this video? Answer with "Yes" or "No".} \#. 

As for MLMMS, the GPT-4 series performs best in terms of detecting the presence of body parts, while Qwen2.5-VL achieves the highest accuracy in identifying whether body parts exhibit semantic distortions. We observe that that identifying distortions in arms is the most challenging, followed by facial distortions while detecting distortions in legs and feet is relatively easier. This aligns with the inherent complexity of structures like faces and arms (particularly hands). Overall, our GHVQ model outperforms in both presence detection and distortion identification tasks, improving accuracy by $2.01\%$ and $15.69\%$, respectively.

\subsection{Ablation Study}
We conducted an ablation study to validate the effectiveness of the spatial quality analyzer, action quality analyzer, and text feature extractor. The experimental results are shown in Table~\ref{tab_ablation1}. 

We observe that the performance across all three quality dimensions and distortion identification degrades the most when the spatial quality analyzer is removed, highlighting the importance of extracting human-focused and AIGC-aware quality features. The action quality analyzer is less crucial than the spatial quality analyzer, possibly because the pre-trained action recognition model is less sensitive to motion in AGVs. The text feature extractor has the least impact compared to the spatial and action quality analyzers, as it only captures video content based on text prompts and does not directly engage with AGVs. In summary, the combination of all three modules achieves the highest performance across all quality dimensions.

\section{Conclusion}

In this paper, we present an in-depth quality assessment study on human activity AGVs, which includes the construction of the large-scale human activity AGV dataset Human-AGVQA and the development of the objective quality metric GHVQ for human activity AGVs. The diversity of human activities and the richness of quality labels in Human-AGVQA make it well-suited for developing and validating quality metrics for human activity AGVs. The GHVQ metric has demonstrated strong performance in assessing these videos, making it a valuable tool for T2V studies to measure progress in generating human activity content. We hope these contributions will promote the development and application of T2V models.


\bibliographystyle{ACM-Reference-Format}
\bibliography{main}


\begin{thebibliography}{105}


\ifx \showCODEN    \undefined \def \showCODEN     #1{\unskip}     \fi
\ifx \showDOI      \undefined \def \showDOI       #1{#1}\fi
\ifx \showISBNx    \undefined \def \showISBNx     #1{\unskip}     \fi
\ifx \showISBNxiii \undefined \def \showISBNxiii  #1{\unskip}     \fi
\ifx \showISSN     \undefined \def \showISSN      #1{\unskip}     \fi
\ifx \showLCCN     \undefined \def \showLCCN      #1{\unskip}     \fi
\ifx \shownote     \undefined \def \shownote      #1{#1}          \fi
\ifx \showarticletitle \undefined \def \showarticletitle #1{#1}   \fi
\ifx \showURL      \undefined \def \showURL       {\relax}        \fi
\providecommand\bibfield[2]{#2}
\providecommand\bibinfo[2]{#2}
\providecommand\natexlab[1]{#1}
\providecommand\showeprint[2][]{arXiv:#2}

\bibitem[Met(2002)]%
        {Methodology}
 \bibinfo{year}{2002}\natexlab{}.
\newblock \showarticletitle{Methodology for the subjective assessment of the quality of television pictures}.
\newblock \bibinfo{journal}{\emph{International Telecommunication Union}} (\bibinfo{year}{2002}).
\newblock


\bibitem[AI(2024a)]%
        {stablevideo}
\bibfield{author}{\bibinfo{person}{Stability AI}.} \bibinfo{year}{2024}\natexlab{a}.
\newblock \bibinfo{booktitle}{\emph{StableVideo}}.
\newblock
\urldef\tempurl%
\url{https://www.stablevideo.com/}
\showURL{%
\tempurl}


\bibitem[AI(2024b)]%
        {ying}
\bibfield{author}{\bibinfo{person}{Zhipu AI}.} \bibinfo{year}{2024}\natexlab{b}.
\newblock \showarticletitle{Ying}.
\newblock
\urldef\tempurl%
\url{https://chatglm.cn/video}
\showURL{%
\tempurl}


\bibitem[Bai et~al\mbox{.}(2025)]%
        {qwen2.5_vl}
\bibfield{author}{\bibinfo{person}{Shuai Bai}, \bibinfo{person}{Keqin Chen}, \bibinfo{person}{Xuejing Liu}, \bibinfo{person}{Jialin Wang}, \bibinfo{person}{Wenbin Ge}, \bibinfo{person}{Sibo Song}, \bibinfo{person}{Kai Dang}, \bibinfo{person}{Peng Wang}, \bibinfo{person}{Shijie Wang}, \bibinfo{person}{Jun Tang}, \bibinfo{person}{Humen Zhong}, \bibinfo{person}{Yuanzhi Zhu}, \bibinfo{person}{Mingkun Yang}, \bibinfo{person}{Zhaohai Li}, \bibinfo{person}{Jianqiang Wan}, \bibinfo{person}{Pengfei Wang}, \bibinfo{person}{Wei Ding}, \bibinfo{person}{Zheren Fu}, \bibinfo{person}{Yiheng Xu}, \bibinfo{person}{Jiabo Ye}, \bibinfo{person}{Xi Zhang}, \bibinfo{person}{Tianbao Xie}, \bibinfo{person}{Zesen Cheng}, \bibinfo{person}{Hang Zhang}, \bibinfo{person}{Zhibo Yang}, \bibinfo{person}{Haiyang Xu}, {and} \bibinfo{person}{Junyang Lin}.} \bibinfo{year}{2025}\natexlab{}.
\newblock \showarticletitle{Qwen2.5-VL Technical Report}.
\newblock \bibinfo{journal}{\emph{arXiv preprint arXiv:2502.13923}} (\bibinfo{year}{2025}).
\newblock


\bibitem[Bain et~al\mbox{.}(2021)]%
        {Webvid_10M}
\bibfield{author}{\bibinfo{person}{Max Bain}, \bibinfo{person}{Arsha Nagrani}, \bibinfo{person}{G{\"u}l Varol}, {and} \bibinfo{person}{Andrew Zisserman}.} \bibinfo{year}{2021}\natexlab{}.
\newblock \showarticletitle{Frozen in time: A joint video and image encoder for end-to-end retrieval}. In \bibinfo{booktitle}{\emph{Proceedings of the IEEE/CVF international conference on computer vision}}. \bibinfo{pages}{1728--1738}.
\newblock


\bibitem[ByteDance(2024)]%
        {jimeng}
\bibfield{author}{\bibinfo{person}{ByteDance}.} \bibinfo{year}{2024}\natexlab{}.
\newblock \showarticletitle{Dreamina}.
\newblock
\urldef\tempurl%
\url{https://jimeng.jianying.com/}
\showURL{%
\tempurl}


\bibitem[Cao et~al\mbox{.}(2017)]%
        {OpenPose}
\bibfield{author}{\bibinfo{person}{Zhe Cao}, \bibinfo{person}{Tomas Simon}, \bibinfo{person}{Shih-En Wei}, {and} \bibinfo{person}{Yaser Sheikh}.} \bibinfo{year}{2017}\natexlab{}.
\newblock \showarticletitle{Realtime multi-person 2d pose estimation using part affinity fields}. In \bibinfo{booktitle}{\emph{Proceedings of the IEEE Conference on Computer Vision and Pattern Recognition (CVPR)}}. \bibinfo{pages}{7291--7299}.
\newblock


\bibitem[Carreira and Zisserman(2017a)]%
        {kinetics}
\bibfield{author}{\bibinfo{person}{Joao Carreira} {and} \bibinfo{person}{Andrew Zisserman}.} \bibinfo{year}{2017}\natexlab{a}.
\newblock \showarticletitle{Quo vadis, action recognition? a new model and the kinetics dataset}. In \bibinfo{booktitle}{\emph{proceedings of the IEEE Conference on Computer Vision and Pattern Recognition}}. \bibinfo{pages}{6299--6308}.
\newblock


\bibitem[Carreira and Zisserman(2017b)]%
        {I3D}
\bibfield{author}{\bibinfo{person}{Joao Carreira} {and} \bibinfo{person}{Andrew Zisserman}.} \bibinfo{year}{2017}\natexlab{b}.
\newblock \showarticletitle{Quo vadis, action recognition? A new model and the kinetics dataset}. In \bibinfo{booktitle}{\emph{Proceedings of the IEEE conference on Computer Vision and Pattern Recognition (CVPR)}}. \bibinfo{pages}{6299--6308}.
\newblock


\bibitem[Chen et~al\mbox{.}(2024a)]%
        {videocrafter2}
\bibfield{author}{\bibinfo{person}{Haoxin Chen}, \bibinfo{person}{Yong Zhang}, \bibinfo{person}{Xiaodong Cun}, \bibinfo{person}{Menghan Xia}, \bibinfo{person}{Xintao Wang}, \bibinfo{person}{Chao Weng}, {and} \bibinfo{person}{Ying Shan}.} \bibinfo{year}{2024}\natexlab{a}.
\newblock \showarticletitle{VideoCrafter2: Overcoming Data Limitations for High-Quality Video Diffusion Models}.
\newblock \bibinfo{journal}{\emph{arXiv}} (\bibinfo{year}{2024}).
\newblock
\showeprint{2401.09047}~[cs.CV]


\bibitem[Chen et~al\mbox{.}(2024b)]%
        {chen2024videocrafter2}
\bibfield{author}{\bibinfo{person}{Haoxin Chen}, \bibinfo{person}{Yong Zhang}, \bibinfo{person}{Xiaodong Cun}, \bibinfo{person}{Menghan Xia}, \bibinfo{person}{Xintao Wang}, \bibinfo{person}{Chao Weng}, {and} \bibinfo{person}{Ying Shan}.} \bibinfo{year}{2024}\natexlab{b}.
\newblock \showarticletitle{Videocrafter2: Overcoming data limitations for high-quality video diffusion models}.
\newblock \bibinfo{journal}{\emph{arXiv preprint arXiv:2401.09047}} (\bibinfo{year}{2024}).
\newblock


\bibitem[Chen et~al\mbox{.}(2018)]%
        {Deeplabv3}
\bibfield{author}{\bibinfo{person}{Liang-Chieh Chen}, \bibinfo{person}{Yukun Zhu}, \bibinfo{person}{George Papandreou}, \bibinfo{person}{Florian Schroff}, {and} \bibinfo{person}{Hartwig Adam}.} \bibinfo{year}{2018}\natexlab{}.
\newblock \showarticletitle{Encoder-Decoder with Atrous Separable Convolution for Semantic Image Segmentation}. In \bibinfo{booktitle}{\emph{Proceedings of the European Conference on Computer Vision (ECCV)}}. \bibinfo{pages}{801--818}.
\newblock


\bibitem[Cheng et~al\mbox{.}(2024)]%
        {VideoLLaMA2}
\bibfield{author}{\bibinfo{person}{Zesen Cheng}, \bibinfo{person}{Sicong Leng}, \bibinfo{person}{Hang Zhang}, \bibinfo{person}{Yifei Xin}, \bibinfo{person}{Xin Li}, \bibinfo{person}{Guanzheng Chen}, \bibinfo{person}{Yongxin Zhu}, \bibinfo{person}{Wenqi Zhang}, \bibinfo{person}{Ziyang Luo}, \bibinfo{person}{Deli Zhao}, {et~al\mbox{.}}} \bibinfo{year}{2024}\natexlab{}.
\newblock \showarticletitle{VideoLLaMA 2: Advancing Spatial-Temporal Modeling and Audio Understanding in Video-LLMs}.
\newblock \bibinfo{journal}{\emph{arXiv preprint arXiv:2406.07476}} (\bibinfo{year}{2024}).
\newblock


\bibitem[Cherti et~al\mbox{.}(2023)]%
        {laion}
\bibfield{author}{\bibinfo{person}{Mehdi Cherti}, \bibinfo{person}{Romain Beaumont}, \bibinfo{person}{Ross Wightman}, \bibinfo{person}{Mitchell Wortsman}, \bibinfo{person}{Gabriel Ilharco}, \bibinfo{person}{Cade Gordon}, \bibinfo{person}{Christoph Schuhmann}, \bibinfo{person}{Ludwig Schmidt}, {and} \bibinfo{person}{Jenia Jitsev}.} \bibinfo{year}{2023}\natexlab{}.
\newblock \showarticletitle{Reproducible scaling laws for contrastive language-image learning}. In \bibinfo{booktitle}{\emph{Proceedings of the IEEE/CVF Conference on Computer Vision and Pattern Recognition}}. \bibinfo{pages}{2818--2829}.
\newblock


\bibitem[Cho et~al\mbox{.}(2024)]%
        {cho2024sora}
\bibfield{author}{\bibinfo{person}{Joseph Cho}, \bibinfo{person}{Fachrina~Dewi Puspitasari}, \bibinfo{person}{Sheng Zheng}, \bibinfo{person}{Jingyao Zheng}, \bibinfo{person}{Lik-Hang Lee}, \bibinfo{person}{Tae-Ho Kim}, \bibinfo{person}{Choong~Seon Hong}, {and} \bibinfo{person}{Chaoning Zhang}.} \bibinfo{year}{2024}\natexlab{}.
\newblock \showarticletitle{Sora as an agi world model? a complete survey on text-to-video generation}.
\newblock \bibinfo{journal}{\emph{arXiv preprint arXiv:2403.05131}} (\bibinfo{year}{2024}).
\newblock


\bibitem[Cloud(2025)]%
        {wanxiang}
\bibfield{author}{\bibinfo{person}{Alibaba Cloud}.} \bibinfo{year}{2025}\natexlab{}.
\newblock \showarticletitle{Wanxiang}.
\newblock
\urldef\tempurl%
\url{https://tongyi.aliyun.com/wanxiang}
\showURL{%
\tempurl}


\bibitem[Dosovitskiy et~al\mbox{.}(2021)]%
        {vit}
\bibfield{author}{\bibinfo{person}{Alexey Dosovitskiy}, \bibinfo{person}{Lucas Beyer}, \bibinfo{person}{Alexander Kolesnikov}, \bibinfo{person}{Dirk Weissenborn}, \bibinfo{person}{Xiaohua Zhai}, \bibinfo{person}{Thomas Unterthiner}, \bibinfo{person}{Mostafa Dehghani}, \bibinfo{person}{Matthias Minderer}, \bibinfo{person}{Georg Heigold}, \bibinfo{person}{Sylvain Gelly}, \bibinfo{person}{Jakob Uszkoreit}, {and} \bibinfo{person}{Neil Houlsby}.} \bibinfo{year}{2021}\natexlab{}.
\newblock \bibinfo{title}{An Image is Worth 16x16 Words: Transformers for Image Recognition at Scale}.
\newblock
\newblock
\showeprint[arxiv]{2010.11929}~[cs.CV]
\urldef\tempurl%
\url{https://arxiv.org/abs/2010.11929}
\showURL{%
\tempurl}


\bibitem[Esser et~al\mbox{.}(2023)]%
        {Gen2}
\bibfield{author}{\bibinfo{person}{Patrick Esser}, \bibinfo{person}{Johnathan Chiu}, \bibinfo{person}{Parmida Atighehchian}, \bibinfo{person}{Jonathan Granskog}, {and} \bibinfo{person}{Anastasis Germanidis}.} \bibinfo{year}{2023}\natexlab{}.
\newblock \showarticletitle{Structure and Content-Guided Video Synthesis with Diffusion Models}. In \bibinfo{booktitle}{\emph{Proceedings of the IEEE/CVF International Conference on Computer Vision (ICCV)}}. \bibinfo{pages}{7346--7356}.
\newblock


\bibitem[Fang et~al\mbox{.}(2020)]%
        {SPAQ}
\bibfield{author}{\bibinfo{person}{Yuming Fang}, \bibinfo{person}{Hanwei Zhu}, \bibinfo{person}{Yan Zeng}, \bibinfo{person}{Kede Ma}, {and} \bibinfo{person}{Zhou Wang}.} \bibinfo{year}{2020}\natexlab{}.
\newblock \showarticletitle{Perceptual quality assessment of smartphone photography}. In \bibinfo{booktitle}{\emph{Proceedings of the IEEE/CVF conference on computer vision and pattern recognition}}. \bibinfo{pages}{3677--3686}.
\newblock


\bibitem[Feichtenhofer(2020)]%
        {X3D}
\bibfield{author}{\bibinfo{person}{Christoph Feichtenhofer}.} \bibinfo{year}{2020}\natexlab{}.
\newblock \showarticletitle{X3D: Expanding architectures for efficient video recognition}. In \bibinfo{booktitle}{\emph{Proceedings of the IEEE/CVF Conference on Computer Vision and Pattern Recognition (CVPR)}}. \bibinfo{pages}{2036--2045}.
\newblock


\bibitem[Feichtenhofer et~al\mbox{.}(2019)]%
        {SlowFast}
\bibfield{author}{\bibinfo{person}{Christoph Feichtenhofer}, \bibinfo{person}{Haoqi Fan}, \bibinfo{person}{Jitendra Malik}, {and} \bibinfo{person}{Kaiming He}.} \bibinfo{year}{2019}\natexlab{}.
\newblock \showarticletitle{SlowFast Networks for Video Recognition}. In \bibinfo{booktitle}{\emph{2019 IEEE/CVF International Conference on Computer Vision (ICCV)}}. \bibinfo{pages}{6201--6210}.
\newblock
\urldef\tempurl%
\url{https://doi.org/10.1109/ICCV.2019.00630}
\showDOI{\tempurl}


\bibitem[Gu et~al\mbox{.}(2018)]%
        {Ava}
\bibfield{author}{\bibinfo{person}{Chunhui Gu}, \bibinfo{person}{Chen Sun}, \bibinfo{person}{David~A Ross}, \bibinfo{person}{Carl Vondrick}, \bibinfo{person}{Caroline Pantofaru}, \bibinfo{person}{Yeqing Li}, \bibinfo{person}{Sudheendra Vijayanarasimhan}, \bibinfo{person}{George Toderici}, \bibinfo{person}{Susanna Ricco}, \bibinfo{person}{Rahul Sukthankar}, {et~al\mbox{.}}} \bibinfo{year}{2018}\natexlab{}.
\newblock \showarticletitle{Ava: A video dataset of spatio-temporally localized atomic visual actions}. In \bibinfo{booktitle}{\emph{Proceedings of the IEEE conference on computer vision and pattern recognition}}. \bibinfo{pages}{6047--6056}.
\newblock


\bibitem[Guo et~al\mbox{.}(2024)]%
        {AnimateDiff}
\bibfield{author}{\bibinfo{person}{Yuwei Guo}, \bibinfo{person}{Ceyuan Yang}, \bibinfo{person}{Anyi Rao}, \bibinfo{person}{Zhengyang Liang}, \bibinfo{person}{Yaohui Wang}, \bibinfo{person}{Yu Qiao}, \bibinfo{person}{Maneesh Agrawala}, \bibinfo{person}{Dahua Lin}, {and} \bibinfo{person}{Bo Dai}.} \bibinfo{year}{2024}\natexlab{}.
\newblock \showarticletitle{AnimateDiff: Animate Your Personalized Text-to-Image Diffusion Models without Specific Tuning}.
\newblock \bibinfo{journal}{\emph{International Conference on Learning Representations}} (\bibinfo{year}{2024}).
\newblock


\bibitem[HaCohen et~al\mbox{.}(2024)]%
        {LTXVideo}
\bibfield{author}{\bibinfo{person}{Yoav HaCohen}, \bibinfo{person}{Nisan Chiprut}, \bibinfo{person}{Benny Brazowski}, \bibinfo{person}{Daniel Shalem}, \bibinfo{person}{Dudu Moshe}, \bibinfo{person}{Eitan Richardson}, \bibinfo{person}{Eran Levin}, \bibinfo{person}{Guy Shiran}, \bibinfo{person}{Nir Zabari}, \bibinfo{person}{Ori Gordon}, \bibinfo{person}{Poriya Panet}, \bibinfo{person}{Sapir Weissbuch}, \bibinfo{person}{Victor Kulikov}, \bibinfo{person}{Yaki Bitterman}, \bibinfo{person}{Zeev Melumian}, {and} \bibinfo{person}{Ofir Bibi}.} \bibinfo{year}{2024}\natexlab{}.
\newblock \showarticletitle{LTX-Video: Realtime Video Latent Diffusion}.
\newblock \bibinfo{journal}{\emph{arXiv preprint arXiv:2501.00103}} (\bibinfo{year}{2024}).
\newblock


\bibitem[Han et~al\mbox{.}(2021)]%
        {transformer}
\bibfield{author}{\bibinfo{person}{Kai Han}, \bibinfo{person}{An Xiao}, \bibinfo{person}{Enhua Wu}, \bibinfo{person}{Jianyuan Guo}, \bibinfo{person}{Chunjing Xu}, {and} \bibinfo{person}{Yunhe Wang}.} \bibinfo{year}{2021}\natexlab{}.
\newblock \showarticletitle{Transformer in transformer}.
\newblock \bibinfo{journal}{\emph{Advances in neural information processing systems}}  \bibinfo{volume}{34} (\bibinfo{year}{2021}), \bibinfo{pages}{15908--15919}.
\newblock


\bibitem[He et~al\mbox{.}(2016)]%
        {he2016deep}
\bibfield{author}{\bibinfo{person}{Kaiming He}, \bibinfo{person}{Xiangyu Zhang}, \bibinfo{person}{Shaoqing Ren}, {and} \bibinfo{person}{Jian Sun}.} \bibinfo{year}{2016}\natexlab{}.
\newblock \showarticletitle{Deep residual learning for image recognition}. In \bibinfo{booktitle}{\emph{Proceedings of the IEEE conference on computer vision and pattern recognition}}. \bibinfo{pages}{770--778}.
\newblock


\bibitem[Henschel et~al\mbox{.}(2024)]%
        {StreamingT2V}
\bibfield{author}{\bibinfo{person}{Roberto Henschel}, \bibinfo{person}{Levon Khachatryan}, \bibinfo{person}{Daniil Hayrapetyan}, \bibinfo{person}{Hayk Poghosyan}, \bibinfo{person}{Vahram Tadevosyan}, \bibinfo{person}{Zhangyang Wang}, \bibinfo{person}{Shant Navasardyan}, {and} \bibinfo{person}{Humphrey Shi}.} \bibinfo{year}{2024}\natexlab{}.
\newblock \showarticletitle{StreamingT2V: Consistent, Dynamic, and Extendable Long Video Generation from Text}.
\newblock \bibinfo{journal}{\emph{arXiv preprint arXiv:2403.14773}} (\bibinfo{year}{2024}).
\newblock


\bibitem[Hessel et~al\mbox{.}(2021)]%
        {CLIPScore}
\bibfield{author}{\bibinfo{person}{Jack Hessel}, \bibinfo{person}{Ari Holtzman}, \bibinfo{person}{Maxwell Forbes}, \bibinfo{person}{Ronan Le~Bras}, {and} \bibinfo{person}{Yejin Choi}.} \bibinfo{year}{2021}\natexlab{}.
\newblock \showarticletitle{CLIPScore: A Reference-free Evaluation Metric for Image Captioning}. In \bibinfo{booktitle}{\emph{Proceedings of the 2021 Conference on Empirical Methods in Natural Language Processing (EMNLP)}}. \bibinfo{pages}{7514--7528}.
\newblock


\bibitem[Heusel et~al\mbox{.}(2017)]%
        {FID}
\bibfield{author}{\bibinfo{person}{Martin Heusel}, \bibinfo{person}{Hubert Ramsauer}, \bibinfo{person}{Thomas Unterthiner}, \bibinfo{person}{Bernhard Nessler}, {and} \bibinfo{person}{Sepp Hochreiter}.} \bibinfo{year}{2017}\natexlab{}.
\newblock \showarticletitle{GANs Trained by a Two Time-Scale Update Rule Converge to a Local Nash Equilibrium}. In \bibinfo{booktitle}{\emph{Advances in Neural Information Processing Systems}}, \bibfield{editor}{\bibinfo{person}{I.~Guyon}, \bibinfo{person}{U.~Von Luxburg}, \bibinfo{person}{S.~Bengio}, \bibinfo{person}{H.~Wallach}, \bibinfo{person}{R.~Fergus}, \bibinfo{person}{S.~Vishwanathan}, {and} \bibinfo{person}{R.~Garnett}} (Eds.), Vol.~\bibinfo{volume}{30}. \bibinfo{publisher}{Curran Associates, Inc.}
\newblock


\bibitem[Ho et~al\mbox{.}(2022)]%
        {ho2022video}
\bibfield{author}{\bibinfo{person}{Jonathan Ho}, \bibinfo{person}{Tim Salimans}, \bibinfo{person}{Alexey Gritsenko}, \bibinfo{person}{William Chan}, \bibinfo{person}{Mohammad Norouzi}, {and} \bibinfo{person}{David~J Fleet}.} \bibinfo{year}{2022}\natexlab{}.
\newblock \showarticletitle{Video diffusion models}.
\newblock \bibinfo{journal}{\emph{Advances in Neural Information Processing Systems}}  \bibinfo{volume}{35} (\bibinfo{year}{2022}), \bibinfo{pages}{8633--8646}.
\newblock


\bibitem[Hong et~al\mbox{.}(2022)]%
        {hong2022cogvideo}
\bibfield{author}{\bibinfo{person}{Wenyi Hong}, \bibinfo{person}{Ming Ding}, \bibinfo{person}{Wendi Zheng}, \bibinfo{person}{Xinghan Liu}, {and} \bibinfo{person}{Jie Tang}.} \bibinfo{year}{2022}\natexlab{}.
\newblock \showarticletitle{CogVideo: Large-scale Pretraining for Text-to-Video Generation via Transformers}. In \bibinfo{booktitle}{\emph{The Eleventh International Conference on Learning Representations}}.
\newblock


\bibitem[Hosu et~al\mbox{.}(2017)]%
        {KoNViD_1k}
\bibfield{author}{\bibinfo{person}{Vlad Hosu}, \bibinfo{person}{Franz Hahn}, \bibinfo{person}{Mohsen Jenadeleh}, \bibinfo{person}{Hanhe Lin}, \bibinfo{person}{Hui Men}, \bibinfo{person}{Tam{\'a}s Szir{\'a}nyi}, \bibinfo{person}{Shujun Li}, {and} \bibinfo{person}{Dietmar Saupe}.} \bibinfo{year}{2017}\natexlab{}.
\newblock \showarticletitle{The Konstanz natural video database (KoNViD-1k)}. In \bibinfo{booktitle}{\emph{2017 Ninth international conference on quality of multimedia experience (QoMEX)}}. IEEE, \bibinfo{pages}{1--6}.
\newblock


\bibitem[Hosu et~al\mbox{.}(2020)]%
        {KonIQ_10k}
\bibfield{author}{\bibinfo{person}{Vlad Hosu}, \bibinfo{person}{Hanhe Lin}, \bibinfo{person}{Tamas Sziranyi}, {and} \bibinfo{person}{Dietmar Saupe}.} \bibinfo{year}{2020}\natexlab{}.
\newblock \showarticletitle{KonIQ-10k: An ecologically valid database for deep learning of blind image quality assessment}.
\newblock \bibinfo{journal}{\emph{IEEE Transactions on Image Processing}}  \bibinfo{volume}{29} (\bibinfo{year}{2020}), \bibinfo{pages}{4041--4056}.
\newblock


\bibitem[Huang et~al\mbox{.}(2024)]%
        {VBench}
\bibfield{author}{\bibinfo{person}{Ziqi Huang}, \bibinfo{person}{Yinan He}, \bibinfo{person}{Jiashuo Yu}, \bibinfo{person}{Fan Zhang}, \bibinfo{person}{Chenyang Si}, \bibinfo{person}{Yuming Jiang}, \bibinfo{person}{Yuanhan Zhang}, \bibinfo{person}{Tianxing Wu}, \bibinfo{person}{Qingyang Jin}, \bibinfo{person}{Nattapol Chanpaisit}, \bibinfo{person}{Yaohui Wang}, \bibinfo{person}{Xinyuan Chen}, \bibinfo{person}{Limin Wang}, \bibinfo{person}{Dahua Lin}, \bibinfo{person}{Yu Qiao}, {and} \bibinfo{person}{Ziwei Liu}.} \bibinfo{year}{2024}\natexlab{}.
\newblock \showarticletitle{{VBench}: Comprehensive Benchmark Suite for Video Generative Models}. In \bibinfo{booktitle}{\emph{Proceedings of the IEEE/CVF Conference on Computer Vision and Pattern Recognition}}.
\newblock


\bibitem[Kang et~al\mbox{.}(2014)]%
        {CNNIQA}
\bibfield{author}{\bibinfo{person}{Le Kang}, \bibinfo{person}{Peng Ye}, \bibinfo{person}{Yi Li}, {and} \bibinfo{person}{David Doermann}.} \bibinfo{year}{2014}\natexlab{}.
\newblock \showarticletitle{Convolutional Neural Networks for No-Reference Image Quality Assessment}. In \bibinfo{booktitle}{\emph{Proceedings of the IEEE Conference on Computer Vision and Pattern Recognition (CVPR)}}.
\newblock


\bibitem[Ke et~al\mbox{.}(2021)]%
        {MUSIQ}
\bibfield{author}{\bibinfo{person}{Junjie Ke}, \bibinfo{person}{Qifei Wang}, \bibinfo{person}{Yilin Wang}, \bibinfo{person}{Peyman Milanfar}, {and} \bibinfo{person}{Feng Yang}.} \bibinfo{year}{2021}\natexlab{}.
\newblock \showarticletitle{MUSIQ: Multi-scale Image Quality Transformer}. In \bibinfo{booktitle}{\emph{2021 IEEE/CVF International Conference on Computer Vision (ICCV)}}. \bibinfo{pages}{5128--5137}.
\newblock
\urldef\tempurl%
\url{https://doi.org/10.1109/ICCV48922.2021.00510}
\showDOI{\tempurl}


\bibitem[Khirodkar et~al\mbox{.}(2024)]%
        {Sapiens}
\bibfield{author}{\bibinfo{person}{Rawal Khirodkar}, \bibinfo{person}{Timur Bagautdinov}, \bibinfo{person}{Julieta Martinez}, \bibinfo{person}{Su Zhaoen}, \bibinfo{person}{Austin James}, \bibinfo{person}{Peter Selednik}, \bibinfo{person}{Stuart Anderson}, {and} \bibinfo{person}{Shunsuke Saito}.} \bibinfo{year}{2024}\natexlab{}.
\newblock \bibinfo{title}{Sapiens: Foundation for Human Vision Models}.
\newblock
\newblock
\showeprint[arxiv]{2408.12569}~[cs.CV]


\bibitem[Kirstain et~al\mbox{.}(2023)]%
        {PickScore}
\bibfield{author}{\bibinfo{person}{Yuval Kirstain}, \bibinfo{person}{Adam Poliak}, \bibinfo{person}{Uriel Singer}, {and} \bibinfo{person}{Omer Levy}.} \bibinfo{year}{2023}\natexlab{}.
\newblock \showarticletitle{Pick-a-Pic: An Open Dataset of User Preferences for Text-to-Image Generation}. In \bibinfo{booktitle}{\emph{Advances in Neural Information Processing Systems}}, Vol.~\bibinfo{volume}{36}.
\newblock


\bibitem[Korhonen(2019)]%
        {TLVQM}
\bibfield{author}{\bibinfo{person}{Jari Korhonen}.} \bibinfo{year}{2019}\natexlab{}.
\newblock \showarticletitle{Two-Level Approach for No-Reference Consumer Video Quality Assessment}.
\newblock \bibinfo{journal}{\emph{IEEE Transactions on Image Processing}} \bibinfo{volume}{28}, \bibinfo{number}{12} (\bibinfo{year}{2019}), \bibinfo{pages}{5923--5938}.
\newblock
\urldef\tempurl%
\url{https://doi.org/10.1109/TIP.2019.2923051}
\showDOI{\tempurl}


\bibitem[Kou et~al\mbox{.}(2024)]%
        {T2VQA}
\bibfield{author}{\bibinfo{person}{Tengchuan Kou}, \bibinfo{person}{Xiaohong Liu}, \bibinfo{person}{Zicheng Zhang}, \bibinfo{person}{Chunyi Li}, \bibinfo{person}{Haoning Wu}, \bibinfo{person}{Xiongkuo Min}, \bibinfo{person}{Guangtao Zhai}, {and} \bibinfo{person}{Ning Liu}.} \bibinfo{year}{2024}\natexlab{}.
\newblock \showarticletitle{Subjective-aligned dataset and metric for text-to-video quality assessment}. In \bibinfo{booktitle}{\emph{Proceedings of the 32nd ACM International Conference on Multimedia}}. \bibinfo{pages}{7793--7802}.
\newblock


\bibitem[Lab and etc.(2024)]%
        {Open_Sora_Plan}
\bibfield{author}{\bibinfo{person}{PKU-Yuan Lab} {and} \bibinfo{person}{Tuzhan~AI etc.}} \bibinfo{year}{2024}\natexlab{}.
\newblock \bibinfo{booktitle}{\emph{Open-Sora-Plan}}.
\newblock
\urldef\tempurl%
\url{https://doi.org/10.5281/zenodo.10948109}
\showDOI{\tempurl}


\bibitem[Li et~al\mbox{.}(2022b)]%
        {BVQA}
\bibfield{author}{\bibinfo{person}{Bowen Li}, \bibinfo{person}{Weixia Zhang}, \bibinfo{person}{Meng Tian}, \bibinfo{person}{Guangtao Zhai}, {and} \bibinfo{person}{Xianpei Wang}.} \bibinfo{year}{2022}\natexlab{b}.
\newblock \showarticletitle{Blindly Assess Quality of In-the-Wild Videos via Quality-Aware Pre-Training and Motion Perception}.
\newblock \bibinfo{journal}{\emph{IEEE Transactions on Circuits and Systems for Video Technology}} \bibinfo{volume}{32}, \bibinfo{number}{9} (\bibinfo{year}{2022}), \bibinfo{pages}{5944--5958}.
\newblock
\urldef\tempurl%
\url{https://doi.org/10.1109/TCSVT.2022.3164467}
\showDOI{\tempurl}


\bibitem[Li et~al\mbox{.}(2024a)]%
        {Llava_onevision}
\bibfield{author}{\bibinfo{person}{Bo Li}, \bibinfo{person}{Yuanhan Zhang}, \bibinfo{person}{Dong Guo}, \bibinfo{person}{Renrui Zhang}, \bibinfo{person}{Feng Li}, \bibinfo{person}{Hao Zhang}, \bibinfo{person}{Kaichen Zhang}, \bibinfo{person}{Yanwei Li}, \bibinfo{person}{Ziwei Liu}, {and} \bibinfo{person}{Chunyuan Li}.} \bibinfo{year}{2024}\natexlab{a}.
\newblock \showarticletitle{Llava-onevision: Easy visual task transfer}.
\newblock \bibinfo{journal}{\emph{arXiv preprint arXiv:2408.03326}} (\bibinfo{year}{2024}).
\newblock


\bibitem[Li et~al\mbox{.}(2023a)]%
        {AGIQA}
\bibfield{author}{\bibinfo{person}{Chunyi Li}, \bibinfo{person}{Zicheng Zhang}, \bibinfo{person}{Haoning Wu}, \bibinfo{person}{Wei Sun}, \bibinfo{person}{Xiongkuo Min}, \bibinfo{person}{Xiaohong Liu}, \bibinfo{person}{Guangtao Zhai}, {and} \bibinfo{person}{Weisi Lin}.} \bibinfo{year}{2023}\natexlab{a}.
\newblock \showarticletitle{AGIQA-3K: An Open Database for AI-Generated Image Quality Assessment}.
\newblock \bibinfo{journal}{\emph{IEEE Transactions on Circuits and Systems for Video Technology}} (\bibinfo{year}{2023}), \bibinfo{pages}{1--1}.
\newblock
\urldef\tempurl%
\url{https://doi.org/10.1109/TCSVT.2023.3319020}
\showDOI{\tempurl}


\bibitem[Li et~al\mbox{.}(2024b)]%
        {Llava_next}
\bibfield{author}{\bibinfo{person}{Feng Li}, \bibinfo{person}{Renrui Zhang}, \bibinfo{person}{Hao Zhang}, \bibinfo{person}{Yuanhan Zhang}, \bibinfo{person}{Bo Li}, \bibinfo{person}{Wei Li}, \bibinfo{person}{Zejun Ma}, {and} \bibinfo{person}{Chunyuan Li}.} \bibinfo{year}{2024}\natexlab{b}.
\newblock \showarticletitle{Llava-next-interleave: Tackling multi-image, video, and 3d in large multimodal models}.
\newblock \bibinfo{journal}{\emph{arXiv preprint arXiv:2407.07895}} (\bibinfo{year}{2024}).
\newblock


\bibitem[Li et~al\mbox{.}(2022a)]%
        {BLIP}
\bibfield{author}{\bibinfo{person}{Junnan Li}, \bibinfo{person}{Dongxu Li}, \bibinfo{person}{Caiming Xiong}, {and} \bibinfo{person}{Steven Hoi}.} \bibinfo{year}{2022}\natexlab{a}.
\newblock \showarticletitle{BLIP: Bootstrapping Language-Image Pre-training for Unified Vision-Language Understanding and Generation}. In \bibinfo{booktitle}{\emph{ICML}}.
\newblock


\bibitem[Li et~al\mbox{.}(2018)]%
        {li2018video}
\bibfield{author}{\bibinfo{person}{Yitong Li}, \bibinfo{person}{Martin Min}, \bibinfo{person}{Dinghan Shen}, \bibinfo{person}{David Carlson}, {and} \bibinfo{person}{Lawrence Carin}.} \bibinfo{year}{2018}\natexlab{}.
\newblock \showarticletitle{Video generation from text}. In \bibinfo{booktitle}{\emph{Proceedings of the AAAI conference on artificial intelligence}}, Vol.~\bibinfo{volume}{32}.
\newblock


\bibitem[Li et~al\mbox{.}(2025)]%
        {LLaMA_VID}
\bibfield{author}{\bibinfo{person}{Yanwei Li}, \bibinfo{person}{Chengyao Wang}, {and} \bibinfo{person}{Jiaya Jia}.} \bibinfo{year}{2025}\natexlab{}.
\newblock \showarticletitle{LLaMA-VID: An Image is Worth 2 Tokens in Large Language Models}. In \bibinfo{booktitle}{\emph{Computer Vision -- ECCV 2024}}, \bibfield{editor}{\bibinfo{person}{Ale{\v{s}} Leonardis}, \bibinfo{person}{Elisa Ricci}, \bibinfo{person}{Stefan Roth}, \bibinfo{person}{Olga Russakovsky}, \bibinfo{person}{Torsten Sattler}, {and} \bibinfo{person}{G{\"u}l Varol}} (Eds.). \bibinfo{publisher}{Springer Nature Switzerland}, \bibinfo{address}{Cham}, \bibinfo{pages}{323--340}.
\newblock
\showISBNx{978-3-031-72952-2}


\bibitem[Li et~al\mbox{.}(2023b)]%
        {Amt}
\bibfield{author}{\bibinfo{person}{Zhen Li}, \bibinfo{person}{Zuo-Liang Zhu}, \bibinfo{person}{Ling-Hao Han}, \bibinfo{person}{Qibin Hou}, \bibinfo{person}{Chun-Le Guo}, {and} \bibinfo{person}{Ming-Ming Cheng}.} \bibinfo{year}{2023}\natexlab{b}.
\newblock \showarticletitle{Amt: All-pairs multi-field transforms for efficient frame interpolation}. In \bibinfo{booktitle}{\emph{Proceedings of the IEEE/CVF Conference on Computer Vision and Pattern Recognition}}. \bibinfo{pages}{9801--9810}.
\newblock


\bibitem[Liang et~al\mbox{.}(2022)]%
        {liang2022nuwa}
\bibfield{author}{\bibinfo{person}{Jian Liang}, \bibinfo{person}{Chenfei Wu}, \bibinfo{person}{Xiaowei Hu}, \bibinfo{person}{Zhe Gan}, \bibinfo{person}{Jianfeng Wang}, \bibinfo{person}{Lijuan Wang}, \bibinfo{person}{Zicheng Liu}, \bibinfo{person}{Yuejian Fang}, {and} \bibinfo{person}{Nan Duan}.} \bibinfo{year}{2022}\natexlab{}.
\newblock \showarticletitle{Nuwa-infinity: Autoregressive over autoregressive generation for infinite visual synthesis}.
\newblock \bibinfo{journal}{\emph{Advances in Neural Information Processing Systems}}  \bibinfo{volume}{35} (\bibinfo{year}{2022}), \bibinfo{pages}{15420--15432}.
\newblock


\bibitem[Liang et~al\mbox{.}(2018)]%
        {JPPNet}
\bibfield{author}{\bibinfo{person}{Xiaodan Liang}, \bibinfo{person}{Ke Gong}, \bibinfo{person}{Xiaohui Shen}, {and} \bibinfo{person}{Liang Lin}.} \bibinfo{year}{2018}\natexlab{}.
\newblock \showarticletitle{Look into person: Joint body parsing \& pose estimation network and a new benchmark}. In \bibinfo{booktitle}{\emph{IEEE Transactions on Pattern Analysis and Machine Intelligence (TPAMI)}}.
\newblock


\bibitem[Lin et~al\mbox{.}(2024)]%
        {VILA}
\bibfield{author}{\bibinfo{person}{Ji Lin}, \bibinfo{person}{Hongxu Yin}, \bibinfo{person}{Wei Ping}, \bibinfo{person}{Pavlo Molchanov}, \bibinfo{person}{Mohammad Shoeybi}, {and} \bibinfo{person}{Song Han}.} \bibinfo{year}{2024}\natexlab{}.
\newblock \showarticletitle{VILA: On Pre-training for Visual Language Models}. In \bibinfo{booktitle}{\emph{Proceedings of the IEEE/CVF Conference on Computer Vision and Pattern Recognition (CVPR)}}. \bibinfo{pages}{26689--26699}.
\newblock


\bibitem[Liu et~al\mbox{.}(2024b)]%
        {Survey_1}
\bibfield{author}{\bibinfo{person}{Xiao Liu}, \bibinfo{person}{Xinhao Xiang}, \bibinfo{person}{Zizhong Li}, \bibinfo{person}{Yongheng Wang}, \bibinfo{person}{Zhuoheng Li}, \bibinfo{person}{Zhuosheng Liu}, \bibinfo{person}{Weidi Zhang}, \bibinfo{person}{Weiqi Ye}, {and} \bibinfo{person}{Jiawei Zhang}.} \bibinfo{year}{2024}\natexlab{b}.
\newblock \showarticletitle{A Survey of AI-Generated Video Evaluation}.
\newblock \bibinfo{journal}{\emph{arXiv preprint arXiv:2410.19884}} (\bibinfo{year}{2024}).
\newblock


\bibitem[Liu et~al\mbox{.}(2024a)]%
        {EvalCrafter}
\bibfield{author}{\bibinfo{person}{Yaofang Liu}, \bibinfo{person}{Xiaodong Cun}, \bibinfo{person}{Xuebo Liu}, \bibinfo{person}{Xintao Wang}, \bibinfo{person}{Yong Zhang}, \bibinfo{person}{Haoxin Chen}, \bibinfo{person}{Yang Liu}, \bibinfo{person}{Tieyong Zeng}, \bibinfo{person}{Raymond Chan}, {and} \bibinfo{person}{Ying Shan}.} \bibinfo{year}{2024}\natexlab{a}.
\newblock \showarticletitle{EvalCrafter: Benchmarking and Evaluating Large Video Generation Models}. In \bibinfo{booktitle}{\emph{Proceedings of the IEEE/CVF Conference on Computer Vision and Pattern Recognition (CVPR)}}. \bibinfo{pages}{22139--22149}.
\newblock


\bibitem[Liu et~al\mbox{.}(2023)]%
        {FETV}
\bibfield{author}{\bibinfo{person}{Yuanxin Liu}, \bibinfo{person}{Lei Li}, \bibinfo{person}{Shuhuai Ren}, \bibinfo{person}{Rundong Gao}, \bibinfo{person}{Shicheng Li}, \bibinfo{person}{Sishuo Chen}, \bibinfo{person}{Xu Sun}, {and} \bibinfo{person}{Lu Hou}.} \bibinfo{year}{2023}\natexlab{}.
\newblock \showarticletitle{FETV: A Benchmark for Fine-Grained Evaluation of Open-Domain Text-to-Video Generation}.
\newblock \bibinfo{journal}{\emph{arXiv preprint arXiv: 2311.01813}} (\bibinfo{year}{2023}).
\newblock


\bibitem[Ma et~al\mbox{.}(2024)]%
        {Latte}
\bibfield{author}{\bibinfo{person}{Xin Ma}, \bibinfo{person}{Yaohui Wang}, \bibinfo{person}{Gengyun Jia}, \bibinfo{person}{Xinyuan Chen}, \bibinfo{person}{Ziwei Liu}, \bibinfo{person}{Yuan-Fang Li}, \bibinfo{person}{Cunjian Chen}, {and} \bibinfo{person}{Yu Qiao}.} \bibinfo{year}{2024}\natexlab{}.
\newblock \showarticletitle{Latte: Latent Diffusion Transformer for Video Generation}.
\newblock \bibinfo{journal}{\emph{arXiv preprint arXiv:2401.03048}} (\bibinfo{year}{2024}).
\newblock


\bibitem[Maaz et~al\mbox{.}(2023)]%
        {Video_chatgpt}
\bibfield{author}{\bibinfo{person}{Muhammad Maaz}, \bibinfo{person}{Hanoona Rasheed}, \bibinfo{person}{Salman Khan}, {and} \bibinfo{person}{Fahad~Shahbaz Khan}.} \bibinfo{year}{2023}\natexlab{}.
\newblock \showarticletitle{Video-chatgpt: Towards detailed video understanding via large vision and language models}.
\newblock \bibinfo{journal}{\emph{arXiv preprint arXiv:2306.05424}} (\bibinfo{year}{2023}).
\newblock


\bibitem[Mittal et~al\mbox{.}(2012a)]%
        {BRISQUE}
\bibfield{author}{\bibinfo{person}{Anish Mittal}, \bibinfo{person}{Anush~Krishna Moorthy}, {and} \bibinfo{person}{Alan~Conrad Bovik}.} \bibinfo{year}{2012}\natexlab{a}.
\newblock \showarticletitle{No-reference image quality assessment in the spatial domain}.
\newblock \bibinfo{journal}{\emph{IEEE Transactions on image processing}} \bibinfo{volume}{21}, \bibinfo{number}{12} (\bibinfo{year}{2012}), \bibinfo{pages}{4695--4708}.
\newblock


\bibitem[Mittal et~al\mbox{.}(2012b)]%
        {NIQE}
\bibfield{author}{\bibinfo{person}{Anish Mittal}, \bibinfo{person}{Rajiv Soundararajan}, {and} \bibinfo{person}{Alan~C Bovik}.} \bibinfo{year}{2012}\natexlab{b}.
\newblock \showarticletitle{Making a “completely blind” image quality analyzer}.
\newblock \bibinfo{journal}{\emph{IEEE Signal processing letters}} \bibinfo{volume}{20}, \bibinfo{number}{3} (\bibinfo{year}{2012}), \bibinfo{pages}{209--212}.
\newblock


\bibitem[Newell et~al\mbox{.}(2016)]%
        {hourglass}
\bibfield{author}{\bibinfo{person}{Alejandro Newell}, \bibinfo{person}{Kaiyu Yang}, {and} \bibinfo{person}{Jia Deng}.} \bibinfo{year}{2016}\natexlab{}.
\newblock \bibinfo{title}{Stacked Hourglass Networks for Human Pose Estimation}.
\newblock
\newblock
\showeprint[arxiv]{1603.06937}~[cs.CV]
\urldef\tempurl%
\url{https://arxiv.org/abs/1603.06937}
\showURL{%
\tempurl}


\bibitem[OpenAI(2024a)]%
        {gpt4}
\bibfield{author}{\bibinfo{person}{OpenAI}.} \bibinfo{year}{2024}\natexlab{a}.
\newblock \bibinfo{title}{ChatGPT (Version 4)}.
\newblock
\newblock
\urldef\tempurl%
\url{https://openai.com/chatgpt}
\showURL{%
\tempurl}


\bibitem[OpenAI(2024b)]%
        {sora}
\bibfield{author}{\bibinfo{person}{OpenAI}.} \bibinfo{year}{2024}\natexlab{b}.
\newblock \showarticletitle{Sora}.
\newblock
\urldef\tempurl%
\url{https://sora.com/}
\showURL{%
\tempurl}


\bibitem[Pan et~al\mbox{.}(2017)]%
        {pan2017create}
\bibfield{author}{\bibinfo{person}{Yingwei Pan}, \bibinfo{person}{Zhaofan Qiu}, \bibinfo{person}{Ting Yao}, \bibinfo{person}{Houqiang Li}, {and} \bibinfo{person}{Tao Mei}.} \bibinfo{year}{2017}\natexlab{}.
\newblock \showarticletitle{To create what you tell: Generating videos from captions}. In \bibinfo{booktitle}{\emph{Proceedings of the 25th ACM international conference on Multimedia}}. \bibinfo{pages}{1789--1798}.
\newblock


\bibitem[Ponomarenko et~al\mbox{.}(2015)]%
        {TID2013}
\bibfield{author}{\bibinfo{person}{Nikolay Ponomarenko}, \bibinfo{person}{Lina Jin}, \bibinfo{person}{Oleg Ieremeiev}, \bibinfo{person}{Vladimir Lukin}, \bibinfo{person}{Karen Egiazarian}, \bibinfo{person}{Jaakko Astola}, \bibinfo{person}{Benoit Vozel}, \bibinfo{person}{Kacem Chehdi}, \bibinfo{person}{Marco Carli}, \bibinfo{person}{Federica Battisti}, {et~al\mbox{.}}} \bibinfo{year}{2015}\natexlab{}.
\newblock \showarticletitle{Image database TID2013: Peculiarities, results and perspectives}.
\newblock \bibinfo{journal}{\emph{Signal processing: Image communication}}  \bibinfo{volume}{30} (\bibinfo{year}{2015}), \bibinfo{pages}{57--77}.
\newblock


\bibitem[Radford et~al\mbox{.}(2021a)]%
        {InternVid_10M}
\bibfield{author}{\bibinfo{person}{Alec Radford}, \bibinfo{person}{Jong~Wook Kim}, \bibinfo{person}{Chris Hallacy}, \bibinfo{person}{Aditya Ramesh}, \bibinfo{person}{Gabriel Goh}, \bibinfo{person}{Sandhini Agarwal}, \bibinfo{person}{Girish Sastry}, \bibinfo{person}{Amanda Askell}, \bibinfo{person}{Pamela Mishkin}, \bibinfo{person}{Jack Clark}, {et~al\mbox{.}}} \bibinfo{year}{2021}\natexlab{a}.
\newblock \showarticletitle{Learning transferable visual models from natural language supervision}. In \bibinfo{booktitle}{\emph{International conference on machine learning}}. PmLR, \bibinfo{pages}{8748--8763}.
\newblock


\bibitem[Radford et~al\mbox{.}(2021b)]%
        {CLIP}
\bibfield{author}{\bibinfo{person}{A. Radford}, \bibinfo{person}{J.~W. Kim}, \bibinfo{person}{C. Hallacy}, \bibinfo{person}{A. Ramesh}, \bibinfo{person}{G. Goh}, \bibinfo{person}{S. Agarwal}, \bibinfo{person}{G. Sastry}, \bibinfo{person}{A. Askell}, \bibinfo{person}{P. Mishkin}, \bibinfo{person}{J. Clark}, \bibinfo{person}{G. Krueger}, {and} \bibinfo{person}{I. Sutskever}.} \bibinfo{year}{2021}\natexlab{b}.
\newblock \showarticletitle{Learning Transferable Visual Models From Natural Language Supervision}.
\newblock \bibinfo{journal}{\emph{Proceedings of the 38th International Conference on Machine Learning (ICML)}} (\bibinfo{year}{2021}).
\newblock


\bibitem[Salimans et~al\mbox{.}(2016)]%
        {IS}
\bibfield{author}{\bibinfo{person}{Tim Salimans}, \bibinfo{person}{Ian Goodfellow}, \bibinfo{person}{Wojciech Zaremba}, \bibinfo{person}{Vicki Cheung}, \bibinfo{person}{Alec Radford}, {and} \bibinfo{person}{Xi Chen}.} \bibinfo{year}{2016}\natexlab{}.
\newblock \showarticletitle{Improved techniques for training gans}.
\newblock \bibinfo{journal}{\emph{Advances in neural information processing systems}}  \bibinfo{volume}{29} (\bibinfo{year}{2016}).
\newblock


\bibitem[Shaw et~al\mbox{.}(2018)]%
        {self_attention}
\bibfield{author}{\bibinfo{person}{Peter Shaw}, \bibinfo{person}{Jakob Uszkoreit}, {and} \bibinfo{person}{Ashish Vaswani}.} \bibinfo{year}{2018}\natexlab{}.
\newblock \showarticletitle{Self-attention with relative position representations}.
\newblock \bibinfo{journal}{\emph{arXiv preprint arXiv:1803.02155}} (\bibinfo{year}{2018}).
\newblock


\bibitem[Su et~al\mbox{.}(2020)]%
        {HyperIQA}
\bibfield{author}{\bibinfo{person}{Shaolin Su}, \bibinfo{person}{Qingsen Yan}, \bibinfo{person}{Yu Zhu}, \bibinfo{person}{Cheng Zhang}, \bibinfo{person}{Xin Ge}, \bibinfo{person}{Jinqiu Sun}, {and} \bibinfo{person}{Yanning Zhang}.} \bibinfo{year}{2020}\natexlab{}.
\newblock \showarticletitle{Blindly Assess Image Quality in the Wild Guided by a Self-Adaptive Hyper Network}. In \bibinfo{booktitle}{\emph{Proceedings of the IEEE/CVF Conference on Computer Vision and Pattern Recognition (CVPR)}}.
\newblock


\bibitem[Sun et~al\mbox{.}(2019)]%
        {HRNet}
\bibfield{author}{\bibinfo{person}{Ke Sun}, \bibinfo{person}{Bin Xiao}, \bibinfo{person}{Dong Liu}, {and} \bibinfo{person}{Jingdong Wang}.} \bibinfo{year}{2019}\natexlab{}.
\newblock \showarticletitle{Deep high-resolution representation learning for human pose estimation}. In \bibinfo{booktitle}{\emph{Proceedings of the IEEE/CVF Conference on Computer Vision and Pattern Recognition (CVPR)}}. \bibinfo{pages}{5693--5703}.
\newblock


\bibitem[Sun et~al\mbox{.}({[n.\,d.]})]%
        {sunsora}
\bibfield{author}{\bibinfo{person}{Rui Sun}, \bibinfo{person}{Yumin Zhang}, \bibinfo{person}{Tejal Shah}, \bibinfo{person}{Jiaohao Sun}, \bibinfo{person}{Shuoying Zhang}, \bibinfo{person}{Wenqi Li}, \bibinfo{person}{Haoran Duan}, {and} \bibinfo{person}{Bo Wei}.} \bibinfo{year}{[n.\,d.]}\natexlab{}.
\newblock \showarticletitle{From Sora What We Can See: A Survey of Text-to-Video Generation}.
\newblock  (\bibinfo{year}{[n.\,d.]}).
\newblock


\bibitem[Sun et~al\mbox{.}(2022)]%
        {simpleVQA}
\bibfield{author}{\bibinfo{person}{Wei Sun}, \bibinfo{person}{Xiongkuo Min}, \bibinfo{person}{Wei Lu}, {and} \bibinfo{person}{Guangtao Zhai}.} \bibinfo{year}{2022}\natexlab{}.
\newblock \showarticletitle{A Deep Learning Based No-Reference Quality Assessment Model for UGC Videos}. In \bibinfo{booktitle}{\emph{Proceedings of the 30th ACM International Conference on Multimedia}}. \bibinfo{pages}{856–865}.
\newblock


\bibitem[Sun et~al\mbox{.}(2023)]%
        {StairIQA}
\bibfield{author}{\bibinfo{person}{Wei Sun}, \bibinfo{person}{Xiongkuo Min}, \bibinfo{person}{Danyang Tu}, \bibinfo{person}{Siwei Ma}, {and} \bibinfo{person}{Guangtao Zhai}.} \bibinfo{year}{2023}\natexlab{}.
\newblock \showarticletitle{Blind quality assessment for in-the-wild images via hierarchical feature fusion and iterative mixed database training}.
\newblock \bibinfo{journal}{\emph{IEEE Journal of Selected Topics in Signal Processing}} (\bibinfo{year}{2023}).
\newblock


\bibitem[Sun et~al\mbox{.}(2024)]%
        {minimalisticVQA}
\bibfield{author}{\bibinfo{person}{Wei Sun}, \bibinfo{person}{Wen Wen}, \bibinfo{person}{Xiongkuo Min}, \bibinfo{person}{Long Lan}, \bibinfo{person}{Guangtao Zhai}, {and} \bibinfo{person}{Kede Ma}.} \bibinfo{year}{2024}\natexlab{}.
\newblock \showarticletitle{Analysis of video quality datasets via design of minimalistic video quality models}.
\newblock \bibinfo{journal}{\emph{IEEE Transactions on Pattern Analysis and Machine Intelligence}} (\bibinfo{year}{2024}).
\newblock


\bibitem[Tang et~al\mbox{.}(2020)]%
        {USDL}
\bibfield{author}{\bibinfo{person}{Yansong Tang}, \bibinfo{person}{Zanlin Ni}, \bibinfo{person}{Jiahuan Zhou}, \bibinfo{person}{Danyang Zhang}, \bibinfo{person}{Jiwen Lu}, \bibinfo{person}{Ying Wu}, {and} \bibinfo{person}{Jie Zhou}.} \bibinfo{year}{2020}\natexlab{}.
\newblock \showarticletitle{Uncertainty-Aware Score Distribution Learning for Action Quality Assessment}. In \bibinfo{booktitle}{\emph{Proceedings of the IEEE/CVF Conference on Computer Vision and Pattern Recognition (CVPR)}}.
\newblock


\bibitem[Technology(2024)]%
        {kling}
\bibfield{author}{\bibinfo{person}{Kuaishou Technology}.} \bibinfo{year}{2024}\natexlab{}.
\newblock \showarticletitle{Kling}.
\newblock
\urldef\tempurl%
\url{https://klingai.kuaishou.com/}
\showURL{%
\tempurl}


\bibitem[Teed and Deng(2020)]%
        {RAFT}
\bibfield{author}{\bibinfo{person}{Zachary Teed} {and} \bibinfo{person}{Jia Deng}.} \bibinfo{year}{2020}\natexlab{}.
\newblock \showarticletitle{Raft: Recurrent all-pairs field transforms for optical flow}. In \bibinfo{booktitle}{\emph{Computer Vision--ECCV 2020: 16th European Conference, Glasgow, UK, August 23--28, 2020, Proceedings, Part II 16}}. Springer, \bibinfo{pages}{402--419}.
\newblock


\bibitem[Tran et~al\mbox{.}(2015)]%
        {C3D}
\bibfield{author}{\bibinfo{person}{Du Tran}, \bibinfo{person}{Lubomir Bourdev}, \bibinfo{person}{Rob Fergus}, \bibinfo{person}{Lorenzo Torresani}, {and} \bibinfo{person}{Manohar Paluri}.} \bibinfo{year}{2015}\natexlab{}.
\newblock \showarticletitle{Learning spatiotemporal features with 3D convolutional networks}. In \bibinfo{booktitle}{\emph{Proceedings of the IEEE International Conference on Computer Vision (ICCV)}}. \bibinfo{pages}{4489--4497}.
\newblock


\bibitem[Tu et~al\mbox{.}(2021a)]%
        {VIDEAL}
\bibfield{author}{\bibinfo{person}{Zhengzhong Tu}, \bibinfo{person}{Yilin Wang}, \bibinfo{person}{Neil Birkbeck}, \bibinfo{person}{Balu Adsumilli}, {and} \bibinfo{person}{Alan~C Bovik}.} \bibinfo{year}{2021}\natexlab{a}.
\newblock \showarticletitle{UGC-VQA: Benchmarking blind video quality assessment for user generated content}.
\newblock \bibinfo{journal}{\emph{IEEE Transactions on Image Processing}}  \bibinfo{volume}{30} (\bibinfo{year}{2021}), \bibinfo{pages}{4449--4464}.
\newblock


\bibitem[Tu et~al\mbox{.}(2021b)]%
        {RAPIQUE}
\bibfield{author}{\bibinfo{person}{Zhengzhong Tu}, \bibinfo{person}{Xiangxu Yu}, \bibinfo{person}{Yilin Wang}, \bibinfo{person}{Neil Birkbeck}, \bibinfo{person}{Balu Adsumilli}, {and} \bibinfo{person}{Alan~C. Bovik}.} \bibinfo{year}{2021}\natexlab{b}.
\newblock \showarticletitle{RAPIQUE: Rapid and Accurate Video Quality Prediction of User Generated Content}.
\newblock \bibinfo{journal}{\emph{IEEE Open Journal of Signal Processing}}  \bibinfo{volume}{2} (\bibinfo{year}{2021}), \bibinfo{pages}{425--440}.
\newblock
\urldef\tempurl%
\url{https://doi.org/10.1109/OJSP.2021.3090333}
\showDOI{\tempurl}


\bibitem[Unterthiner et~al\mbox{.}(2018)]%
        {FVD}
\bibfield{author}{\bibinfo{person}{Thomas Unterthiner}, \bibinfo{person}{Sjoerd van Steenkiste}, \bibinfo{person}{Karol Kurach}, \bibinfo{person}{Raphael Marinier}, \bibinfo{person}{Marcin Michalski}, {and} \bibinfo{person}{Sylvain Gelly}.} \bibinfo{year}{2018}\natexlab{}.
\newblock \bibinfo{title}{Towards Accurate Generative Models of Video: A New Metric Challenges}.
\newblock
\newblock
\showeprint[arxiv]{1812.01717}~[cs.CV]


\bibitem[Villegas et~al\mbox{.}(2022)]%
        {villegas2022phenaki}
\bibfield{author}{\bibinfo{person}{Ruben Villegas}, \bibinfo{person}{Mohammad Babaeizadeh}, \bibinfo{person}{Pieter-Jan Kindermans}, \bibinfo{person}{Hernan Moraldo}, \bibinfo{person}{Han Zhang}, \bibinfo{person}{Mohammad~Taghi Saffar}, \bibinfo{person}{Santiago Castro}, \bibinfo{person}{Julius Kunze}, {and} \bibinfo{person}{Dumitru Erhan}.} \bibinfo{year}{2022}\natexlab{}.
\newblock \showarticletitle{Phenaki: Variable length video generation from open domain textual descriptions}. In \bibinfo{booktitle}{\emph{International Conference on Learning Representations}}.
\newblock


\bibitem[Wang et~al\mbox{.}(2023a)]%
        {CLIP_IQA}
\bibfield{author}{\bibinfo{person}{Jianyi Wang}, \bibinfo{person}{Kelvin~C.K. Chan}, {and} \bibinfo{person}{Chen~Change Loy}.} \bibinfo{year}{2023}\natexlab{a}.
\newblock \showarticletitle{Exploring CLIP for Assessing the Look and Feel of Images}.
\newblock \bibinfo{journal}{\emph{Proceedings of the AAAI Conference on Artificial Intelligence}} \bibinfo{volume}{37}, \bibinfo{number}{2} (\bibinfo{date}{Jun.} \bibinfo{year}{2023}), \bibinfo{pages}{2555--2563}.
\newblock
\urldef\tempurl%
\url{https://doi.org/10.1609/aaai.v37i2.25353}
\showDOI{\tempurl}


\bibitem[Wang et~al\mbox{.}(2023b)]%
        {Videomae_v2}
\bibfield{author}{\bibinfo{person}{Limin Wang}, \bibinfo{person}{Bingkun Huang}, \bibinfo{person}{Zhiyu Zhao}, \bibinfo{person}{Zhan Tong}, \bibinfo{person}{Yinan He}, \bibinfo{person}{Yi Wang}, \bibinfo{person}{Yali Wang}, {and} \bibinfo{person}{Yu Qiao}.} \bibinfo{year}{2023}\natexlab{b}.
\newblock \showarticletitle{Videomae v2: Scaling video masked autoencoders with dual masking}. In \bibinfo{booktitle}{\emph{Proceedings of the IEEE/CVF Conference on Computer Vision and Pattern Recognition}}. \bibinfo{pages}{14549--14560}.
\newblock


\bibitem[Wang et~al\mbox{.}(2024a)]%
        {qwen2_vl}
\bibfield{author}{\bibinfo{person}{Peng Wang}, \bibinfo{person}{Shuai Bai}, \bibinfo{person}{Sinan Tan}, \bibinfo{person}{Shijie Wang}, \bibinfo{person}{Zhihao Fan}, \bibinfo{person}{Jinze Bai}, \bibinfo{person}{Keqin Chen}, \bibinfo{person}{Xuejing Liu}, \bibinfo{person}{Jialin Wang}, \bibinfo{person}{Wenbin Ge}, {et~al\mbox{.}}} \bibinfo{year}{2024}\natexlab{a}.
\newblock \showarticletitle{Qwen2-vl: Enhancing vision-language model's perception of the world at any resolution}.
\newblock \bibinfo{journal}{\emph{arXiv preprint arXiv:2409.12191}} (\bibinfo{year}{2024}).
\newblock


\bibitem[Wang et~al\mbox{.}(2024b)]%
        {MA_AGIQA}
\bibfield{author}{\bibinfo{person}{Puyi Wang}, \bibinfo{person}{Wei Sun}, \bibinfo{person}{Zicheng Zhang}, \bibinfo{person}{Jun Jia}, \bibinfo{person}{Yanwei Jiang}, \bibinfo{person}{Zhichao Zhang}, \bibinfo{person}{Xiongkuo Min}, {and} \bibinfo{person}{Guangtao Zhai}.} \bibinfo{year}{2024}\natexlab{b}.
\newblock \showarticletitle{Large Multi-modality Model Assisted AI-Generated Image Quality Assessment}. In \bibinfo{booktitle}{\emph{Proceedings of the 32nd ACM International Conference on Multimedia}}. \bibinfo{pages}{7803--7812}.
\newblock


\bibitem[Wang et~al\mbox{.}(2022)]%
        {viCLIP}
\bibfield{author}{\bibinfo{person}{Yi Wang}, \bibinfo{person}{Kunchang Li}, \bibinfo{person}{Yizhuo Li}, \bibinfo{person}{Yinan He}, \bibinfo{person}{Bingkun Huang}, \bibinfo{person}{Zhiyu Zhao}, \bibinfo{person}{Hongjie Zhang}, \bibinfo{person}{Jilan Xu}, \bibinfo{person}{Yi Liu}, \bibinfo{person}{Zun Wang}, \bibinfo{person}{Sen Xing}, \bibinfo{person}{Guo Chen}, \bibinfo{person}{Junting Pan}, \bibinfo{person}{Jiashuo Yu}, \bibinfo{person}{Yali Wang}, \bibinfo{person}{Limin Wang}, {and} \bibinfo{person}{Yu Qiao}.} \bibinfo{year}{2022}\natexlab{}.
\newblock \showarticletitle{InternVideo: General Video Foundation Models via Generative and Discriminative Learning}.
\newblock \bibinfo{journal}{\emph{arXiv preprint arXiv:2212.03191}} (\bibinfo{year}{2022}).
\newblock


\bibitem[Wen and Wang(2021)]%
        {rank_loss}
\bibfield{author}{\bibinfo{person}{Shaoguo Wen} {and} \bibinfo{person}{Junle Wang}.} \bibinfo{year}{2021}\natexlab{}.
\newblock \showarticletitle{A strong baseline for image and video quality assessment}.
\newblock \bibinfo{journal}{\emph{arXiv preprint arXiv:2111.07104}} (\bibinfo{year}{2021}).
\newblock


\bibitem[Wu et~al\mbox{.}(2022)]%
        {FAST_VQA}
\bibfield{author}{\bibinfo{person}{Haoning Wu}, \bibinfo{person}{Chaofeng Chen}, \bibinfo{person}{Jingwen Hou}, \bibinfo{person}{Liang Liao}, \bibinfo{person}{Annan Wang}, \bibinfo{person}{Wenxiu Sun}, \bibinfo{person}{Qiong Yan}, {and} \bibinfo{person}{Weisi Lin}.} \bibinfo{year}{2022}\natexlab{}.
\newblock \showarticletitle{FAST-VQA: Efficient End-to-End Video Quality Assessment with Fragment Sampling}. In \bibinfo{booktitle}{\emph{Computer Vision – ECCV 2022: 17th European Conference, Tel Aviv, Israel, October 23–27, 2022, Proceedings, Part VI}} (Tel Aviv, Israel). \bibinfo{publisher}{Springer-Verlag}, \bibinfo{address}{Berlin, Heidelberg}, \bibinfo{pages}{538–554}.
\newblock
\showISBNx{978-3-031-20067-0}
\urldef\tempurl%
\url{https://doi.org/10.1007/978-3-031-20068-7_31}
\showDOI{\tempurl}


\bibitem[Wu et~al\mbox{.}(2023b)]%
        {dover}
\bibfield{author}{\bibinfo{person}{Haoning Wu}, \bibinfo{person}{Erli Zhang}, \bibinfo{person}{Liang Liao}, \bibinfo{person}{Chaofeng Chen}, \bibinfo{person}{Jingwen~Hou Hou}, \bibinfo{person}{Annan Wang}, \bibinfo{person}{Wenxiu~Sun Sun}, \bibinfo{person}{Qiong Yan}, {and} \bibinfo{person}{Weisi Lin}.} \bibinfo{year}{2023}\natexlab{b}.
\newblock \showarticletitle{Exploring Video Quality Assessment on User Generated Contents from Aesthetic and Technical Perspectives}. In \bibinfo{booktitle}{\emph{International Conference on Computer Vision (ICCV)}}.
\newblock


\bibitem[Wu et~al\mbox{.}(2023c)]%
        {qalign}
\bibfield{author}{\bibinfo{person}{Haoning Wu}, \bibinfo{person}{Zicheng Zhang}, \bibinfo{person}{Weixia Zhang}, \bibinfo{person}{Chaofeng Chen}, \bibinfo{person}{Liang Liao}, \bibinfo{person}{Chunyi Li}, \bibinfo{person}{Yixuan Gao}, \bibinfo{person}{Annan Wang}, \bibinfo{person}{Erli Zhang}, \bibinfo{person}{Wenxiu Sun}, {et~al\mbox{.}}} \bibinfo{year}{2023}\natexlab{c}.
\newblock \showarticletitle{Q-align: Teaching lmms for visual scoring via discrete text-defined levels}.
\newblock \bibinfo{journal}{\emph{arXiv preprint arXiv:2312.17090}} (\bibinfo{year}{2023}).
\newblock


\bibitem[Wu et~al\mbox{.}(2023a)]%
        {HPSv2}
\bibfield{author}{\bibinfo{person}{Xiaoshi Wu}, \bibinfo{person}{Yiming Hao}, \bibinfo{person}{Keqiang Sun}, \bibinfo{person}{Yixiong Chen}, \bibinfo{person}{Feng Zhu}, \bibinfo{person}{Rui Zhao}, {and} \bibinfo{person}{Hongsheng Li}.} \bibinfo{year}{2023}\natexlab{a}.
\newblock \bibinfo{title}{Human Preference Score v2: A Solid Benchmark for Evaluating Human Preferences of Text-to-Image Synthesis}.
\newblock
\newblock
\showeprint[arxiv]{2306.09341}~[cs.CV]


\bibitem[Wu et~al\mbox{.}(2024)]%
        {deepseekvl2}
\bibfield{author}{\bibinfo{person}{Zhiyu Wu}, \bibinfo{person}{Xiaokang Chen}, \bibinfo{person}{Zizheng Pan}, \bibinfo{person}{Xingchao Liu}, \bibinfo{person}{Wen Liu}, \bibinfo{person}{Damai Dai}, \bibinfo{person}{Huazuo Gao}, \bibinfo{person}{Yiyang Ma}, \bibinfo{person}{Chengyue Wu}, \bibinfo{person}{Bingxuan Wang}, \bibinfo{person}{Zhenda Xie}, \bibinfo{person}{Yu Wu}, \bibinfo{person}{Kai Hu}, \bibinfo{person}{Jiawei Wang}, \bibinfo{person}{Yaofeng Sun}, \bibinfo{person}{Yukun Li}, \bibinfo{person}{Yishi Piao}, \bibinfo{person}{Kang Guan}, \bibinfo{person}{Aixin Liu}, \bibinfo{person}{Xin Xie}, \bibinfo{person}{Yuxiang You}, \bibinfo{person}{Kai Dong}, \bibinfo{person}{Xingkai Yu}, \bibinfo{person}{Haowei Zhang}, \bibinfo{person}{Liang Zhao}, \bibinfo{person}{Yisong Wang}, {and} \bibinfo{person}{Chong Ruan}.} \bibinfo{year}{2024}\natexlab{}.
\newblock \bibinfo{title}{DeepSeek-VL2: Mixture-of-Experts Vision-Language Models for Advanced Multimodal Understanding}.
\newblock
\newblock
\showeprint[arxiv]{2412.10302}~[cs.CV]
\urldef\tempurl%
\url{https://arxiv.org/abs/2412.10302}
\showURL{%
\tempurl}


\bibitem[Xu et~al\mbox{.}(2023)]%
        {ImageReward}
\bibfield{author}{\bibinfo{person}{Jiazheng Xu}, \bibinfo{person}{Xiao Liu}, \bibinfo{person}{Yuchen Wu}, \bibinfo{person}{Yuxuan Tong}, \bibinfo{person}{Qinkai Li}, \bibinfo{person}{Min Ding}, \bibinfo{person}{Jie Tang}, {and} \bibinfo{person}{Yuxiao Dong}.} \bibinfo{year}{2023}\natexlab{}.
\newblock \showarticletitle{ImageReward: Learning and Evaluating Human Preferences for Text-to-Image Generation}. In \bibinfo{booktitle}{\emph{Advances in Neural Information Processing Systems}}.
\newblock


\bibitem[Xu et~al\mbox{.}(2022)]%
        {TSA}
\bibfield{author}{\bibinfo{person}{Jinglin Xu}, \bibinfo{person}{Yongming Rao}, \bibinfo{person}{Xumin Yu}, \bibinfo{person}{Guangyi Chen}, \bibinfo{person}{Jie Zhou}, {and} \bibinfo{person}{Jiwen Lu}.} \bibinfo{year}{2022}\natexlab{}.
\newblock \showarticletitle{FineDiving: A Fine-Grained Dataset for Procedure-Aware Action Quality Assessment}. In \bibinfo{booktitle}{\emph{Proceedings of the IEEE/CVF Conference on Computer Vision and Pattern Recognition (CVPR)}}. \bibinfo{pages}{2949--2958}.
\newblock


\bibitem[Ying et~al\mbox{.}(2021)]%
        {PatchVQ}
\bibfield{author}{\bibinfo{person}{Zhenqiang Ying}, \bibinfo{person}{Maniratnam Mandal}, \bibinfo{person}{Deepti Ghadiyaram}, {and} \bibinfo{person}{Alan Bovik}.} \bibinfo{year}{2021}\natexlab{}.
\newblock \showarticletitle{Patch-VQ: 'Patching Up' the Video Quality Problem}. In \bibinfo{booktitle}{\emph{Proceedings of the IEEE/CVF Conference on Computer Vision and Pattern Recognition (CVPR)}}. \bibinfo{pages}{14019--14029}.
\newblock


\bibitem[Ying et~al\mbox{.}(2020)]%
        {ying2020patches}
\bibfield{author}{\bibinfo{person}{Zhenqiang Ying}, \bibinfo{person}{Haoran Niu}, \bibinfo{person}{Praful Gupta}, \bibinfo{person}{Dhruv Mahajan}, \bibinfo{person}{Deepti Ghadiyaram}, {and} \bibinfo{person}{Alan Bovik}.} \bibinfo{year}{2020}\natexlab{}.
\newblock \showarticletitle{From patches to pictures (PaQ-2-PiQ): Mapping the perceptual space of picture quality}. In \bibinfo{booktitle}{\emph{Proceedings of the IEEE/CVF conference on computer vision and pattern recognition}}. \bibinfo{pages}{3575--3585}.
\newblock


\bibitem[Yu et~al\mbox{.}(2021)]%
        {CoRe}
\bibfield{author}{\bibinfo{person}{Xumin Yu}, \bibinfo{person}{Yongming Rao}, \bibinfo{person}{Wenliang Zhao}, \bibinfo{person}{Jiwen Lu}, {and} \bibinfo{person}{Jie Zhou}.} \bibinfo{year}{2021}\natexlab{}.
\newblock \showarticletitle{Group-Aware Contrastive Regression for Action Quality Assessment}. In \bibinfo{booktitle}{\emph{Proceedings of the IEEE/CVF International Conference on Computer Vision (ICCV)}}. \bibinfo{pages}{7919--7928}.
\newblock


\bibitem[Yuan et~al\mbox{.}(2024)]%
        {MagicTime}
\bibfield{author}{\bibinfo{person}{Shenghai Yuan}, \bibinfo{person}{Jinfa Huang}, \bibinfo{person}{Yujun Shi}, \bibinfo{person}{Yongqi Xu}, \bibinfo{person}{Ruijie Zhu}, \bibinfo{person}{Bin Lin}, \bibinfo{person}{Xinhua Cheng}, \bibinfo{person}{Li Yuan}, {and} \bibinfo{person}{Jiebo Luo}.} \bibinfo{year}{2024}\natexlab{}.
\newblock \showarticletitle{MagicTime: Time-lapse Video Generation Models as Metamorphic Simulators}.
\newblock \bibinfo{journal}{\emph{arXiv preprint arXiv:2404.05014}} (\bibinfo{year}{2024}).
\newblock


\bibitem[Zeng et~al\mbox{.}(2020)]%
        {ACTION_NET}
\bibfield{author}{\bibinfo{person}{Ling-An Zeng}, \bibinfo{person}{Fa-Ting Hong}, \bibinfo{person}{Wei-Shi Zheng}, \bibinfo{person}{Qi-Zhi Yu}, \bibinfo{person}{Wei Zeng}, \bibinfo{person}{Yao-Wei Wang}, {and} \bibinfo{person}{Jian-Huang Lai}.} \bibinfo{year}{2020}\natexlab{}.
\newblock \showarticletitle{Hybrid Dynamic-static Context-aware Attention Network for Action Assessment in Long Videos}. In \bibinfo{booktitle}{\emph{Proceedings of the 28th ACM International Conference on Multimedia}} (Seattle, WA, USA) \emph{(\bibinfo{series}{MM '20})}. \bibinfo{publisher}{Association for Computing Machinery}, \bibinfo{address}{New York, NY, USA}, \bibinfo{pages}{2526–2534}.
\newblock
\showISBNx{9781450379885}
\urldef\tempurl%
\url{https://doi.org/10.1145/3394171.3413560}
\showDOI{\tempurl}


\bibitem[Zeng et~al\mbox{.}(2023)]%
        {zeng2023make}
\bibfield{author}{\bibinfo{person}{Yan Zeng}, \bibinfo{person}{Guoqiang Wei}, \bibinfo{person}{Jiani Zheng}, \bibinfo{person}{Jiaxin Zou}, \bibinfo{person}{Yang Wei}, \bibinfo{person}{Yuchen Zhang}, {and} \bibinfo{person}{Hang Li}.} \bibinfo{year}{2023}\natexlab{}.
\newblock \showarticletitle{Make pixels dance: High-dynamic video generation}.
\newblock \bibinfo{journal}{\emph{arXiv preprint arXiv:2311.10982}} (\bibinfo{year}{2023}).
\newblock


\bibitem[Zhang et~al\mbox{.}(2023a)]%
        {show1}
\bibfield{author}{\bibinfo{person}{David~Junhao Zhang}, \bibinfo{person}{Jay~Zhangjie Wu}, \bibinfo{person}{Jia-Wei Liu}, \bibinfo{person}{Rui Zhao}, \bibinfo{person}{Lingmin Ran}, \bibinfo{person}{Yuchao Gu}, \bibinfo{person}{Difei Gao}, {and} \bibinfo{person}{Mike~Zheng Shou}.} \bibinfo{year}{2023}\natexlab{a}.
\newblock \showarticletitle{Show-1: Marrying Pixel and Latent Diffusion Models for Text-to-Video Generation}.
\newblock \bibinfo{journal}{\emph{arXiv preprint arXiv:2309.15818}} (\bibinfo{year}{2023}).
\newblock


\bibitem[Zhang et~al\mbox{.}(2021)]%
        {UNIQUE}
\bibfield{author}{\bibinfo{person}{Weixia Zhang}, \bibinfo{person}{Kede Ma}, \bibinfo{person}{Guangtao Zhai}, {and} \bibinfo{person}{Xiaokang Yang}.} \bibinfo{year}{2021}\natexlab{}.
\newblock \showarticletitle{Uncertainty-Aware Blind Image Quality Assessment in the Laboratory and Wild}.
\newblock \bibinfo{journal}{\emph{IEEE Transactions on Image Processing}}  \bibinfo{volume}{30} (\bibinfo{year}{2021}), \bibinfo{pages}{3474--3486}.
\newblock
\urldef\tempurl%
\url{https://doi.org/10.1109/TIP.2021.3061932}
\showDOI{\tempurl}


\bibitem[Zhang et~al\mbox{.}(2023b)]%
        {LIQE}
\bibfield{author}{\bibinfo{person}{Weixia Zhang}, \bibinfo{person}{Guangtao Zhai}, \bibinfo{person}{Ying Wei}, \bibinfo{person}{Xiaokang Yang}, {and} \bibinfo{person}{Kede Ma}.} \bibinfo{year}{2023}\natexlab{b}.
\newblock \showarticletitle{Blind Image Quality Assessment via Vision-Language Correspondence: A Multitask Learning Perspective}. In \bibinfo{booktitle}{\emph{Proceedings of the IEEE/CVF Conference on Computer Vision and Pattern Recognition (CVPR)}}. \bibinfo{pages}{14071--14081}.
\newblock


\bibitem[Zhang et~al\mbox{.}(2024)]%
        {UGVQ}
\bibfield{author}{\bibinfo{person}{Zhichao Zhang}, \bibinfo{person}{Xinyue Li}, \bibinfo{person}{Wei Sun}, \bibinfo{person}{Jun Jia}, \bibinfo{person}{Xiongkuo Min}, \bibinfo{person}{Zicheng Zhang}, \bibinfo{person}{Chunyi Li}, \bibinfo{person}{Zijian Chen}, \bibinfo{person}{Puyi Wang}, \bibinfo{person}{Zhongpeng Ji}, {et~al\mbox{.}}} \bibinfo{year}{2024}\natexlab{}.
\newblock \showarticletitle{Benchmarking AIGC Video Quality Assessment: A Dataset and Unified Model}.
\newblock \bibinfo{journal}{\emph{arXiv preprint arXiv:2407.21408}} (\bibinfo{year}{2024}).
\newblock


\end{thebibliography}

\clearpage
\appendix
\twocolumn[{\section*{\centering Supplemental Materials\\}}\vspace*{10mm}]






In this supplementary file, we provide more details of \textit{Human}-AGVQA dataset in Section~\ref{sec_dataset_details}, including text prompts selection, T2V model details, and subjective assessment experiments. We then provide data processing details and further analysis in Section~\ref{sec_data_processing}, including MOS analysis for T2V models, distortion identification analysis, and the relationship between distortion identification and visual quality. The detailed experiment results are discussed in Section~\ref{sec_exp_detail}, including evaluation criteria in our experiments, compared quality metrics in the proposed benchmark, training details of proposed GHVQ metric, and ablation study of different backbones.

\section{\textit{Human}-AGVQA Dataset}
\label{sec_dataset_details}
\subsection{Text Prompts Selection}

\begin{figure}
    \centering
    \includegraphics[width=0.99\linewidth]{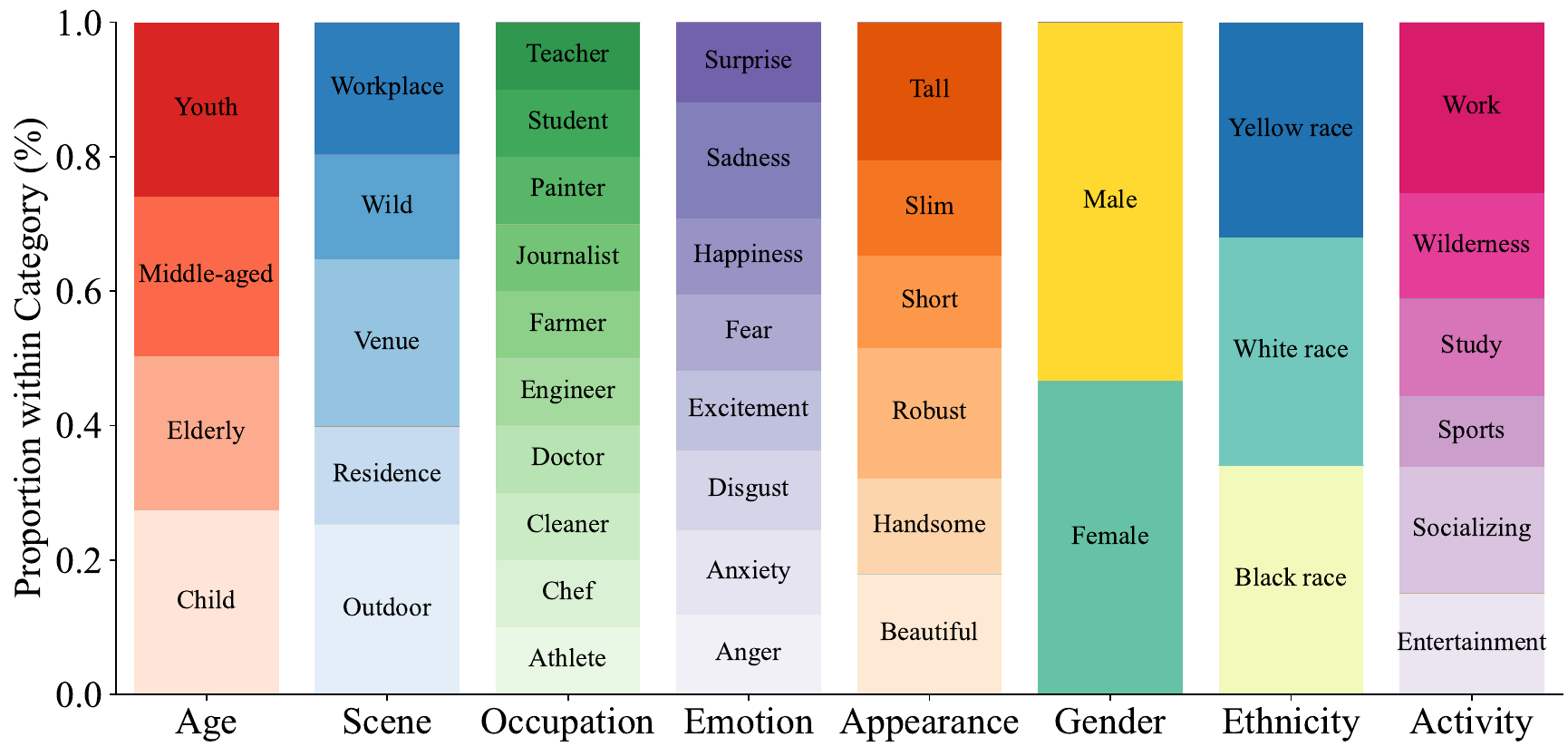}
    \caption{The distribution of 44 subcategories in 8 categories of the 400 prompts.
    }
    \label{fig_prompt_cls_bar}
\end{figure}

To ensure that the text prompts in our dataset encompass a wide range of real-world human activities, we perform a comprehensive and systematic classification of the words used to create human activity-oriented text prompts. 
Specifically, the categories of human activities are derived from the AIGC VQA dataset, Vbench\cite{VBench}, FETV\cite{FETV}, and real-world video caption datasets, InternVid-10M\cite{InternVid_10M} and Webvid-10M\cite{Webvid_10M}. We use GPT-4 to identify high-frequency activities within these datasets, and finally we divide the words into $8$ categories according to their properties: \textbf{\textit{ages, genders, races, occupations, scenes, emotion, appearance, and activities}}. Each of these categories is then further subdivided into more specific subcategories that are commonly observed in daily human life, as illustrated in Table~\ref{tab_prompt_cls}.
The prompts require at least one person to be present. 60\% of the prompts specify a single person, while 40\% involve multiple individuals.



\begin{figure}
    \centering
    \includegraphics[width=0.99\linewidth]{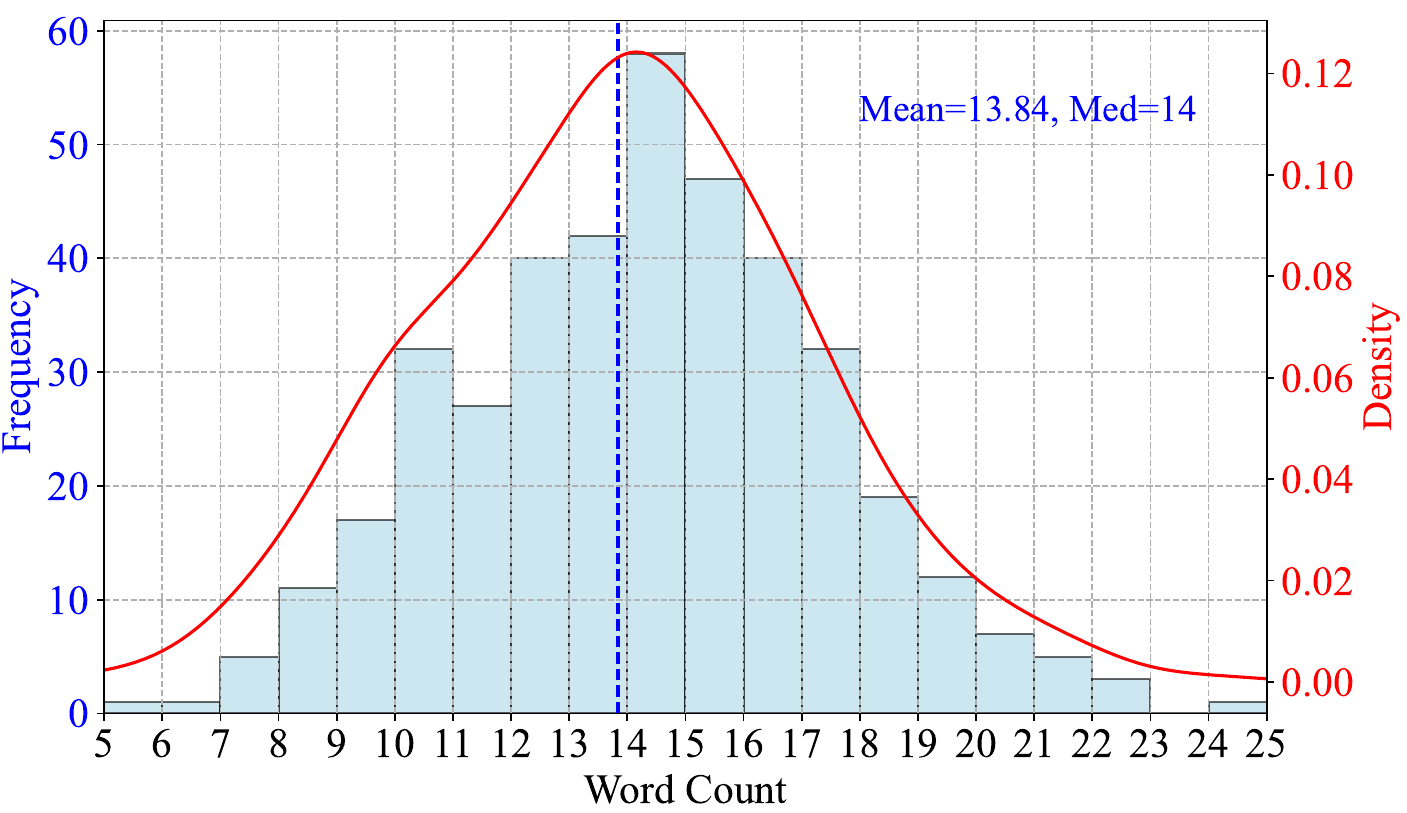}
    \caption{Histogram and density plot of word count per prompt. }
    \label{fig_prompt_length}
\end{figure}

The proportions of each subcategory within their respective categories are illustrated in Figure~\ref{fig_prompt_cls_bar}. It can be observed that the proportions of subcategories in the \textbf{\textit{age}}, \textbf{\textit{occupation}}, \textbf{\textit{gender}}, and \textbf{\textit{ethnicity}} categories are identical. While for other categories, the subcategories \textbf{\textit{venue}} and \textbf{\textit{outdoor}} in \textbf{\textit{scene}}, \textbf{\textit{sadness}} in \textbf{\textit{emotion}}, \textbf{\textit{robust}} in \textbf{\textit{appearance}}, and \textbf{\textit{work}} and \textbf{\textit{socializing}} in \textbf{\textit{activity}} have larger proportions than their respective subcategories. In addition, we analyzed the length of text prompts in the \textit{Human}-AGVQA dataset, and the distribution of word counts are presented in Figure~\ref{fig_prompt_length}. The average text prompt length is $13.85$, with a median of $14$ words.

\begin{table}
\centering
\caption{Subcategories of \textit{ages, genders, races, occupations, scenes, emotion, appearance, and activities}.}
\resizebox{0.49\textwidth}{!}{
\begin{tabular}{|l|l|}
\toprule
\textbf{Category} & \textbf{Subcategories} \\ \midrule
age & child, youth, middle-aged, elderly \\ \midrule
gender & male, female \\ \midrule
scene & outdoor, residence, venue, wild, workplace \\ \midrule
appearance & tall, slim, short, handsome, robust, beautiful \\ \midrule
occupation & \makecell[l]{teacher, doctor, engineer, journalist, chef, \\ painter, farmer, athlete, cleaner, student} \\ \midrule
emotion & \makecell[l]{happiness, sadness, surprise, anger, \\ fear, anxiety, excitement, disgust} \\ \midrule
ethnicity & white race, yellow race, black race \\ \midrule
activity & \makecell[l]{study, work, socializing, entertainment, \\ wilderness, sports} \\
\bottomrule
\end{tabular}
}
\label{tab_prompt_cls}
\end{table}

\begin{table*}
\centering
  \caption{Video formats generated by the $15$ T2V models in the \textit{Human}-AGVQA dataset.}
  \label{tab_t2v}
  \resizebox{0.95\textwidth}{!}{
  \begin{tabular}{lccccccc}
    \toprule
    Methods & Duration (s) & FPS & Resolution & Launch Month & Utilized Version  & Version Release Month & Open Source  \\
    \midrule                                                                                                                                    
    Show-1~\cite{show1}                  & $3.4$   & $8$  & $1914 \times 951$  & 2023.10   & -                       & 2023.10   & \ding{52}  \\
    VideoCrafter2~\cite{videocrafter2}   & $1.6$   & $10$ & $512 \times 320$   & 2023.04   & -                       & 2024.01   & \ding{52}  \\
    StreamingT2V~\cite{StreamingT2V}     & $5.0$   & $8$  & $1280 \times 720$  & 2024.03   & -                       & 2024.03   & \ding{52}  \\
    AnimateDiff~\cite{AnimateDiff}       & $2.0$   & $8$  & $512 \times 512$   & 2024.03   & AnimateDiff-Lightning   & 2024.03   & \ding{52}  \\
    MagicTime~\cite{MagicTime}           & $2.1$   & $23$ & $512 \times 512$   & 2024.04   & -                       & 2024.04   & \ding{52}  \\
    Open-sora-plan~\cite{Open_Sora_Plan} & $2.7$   & $24$ & $512 \times 512$   & 2024.03   & v1.1.0                  & 2024.05   & \ding{52}  \\
    Latte~\cite{Latte}                   & $2.0$   & $8$  & $512 \times 512$   & 2024.01   & Latte-1                 & 2024.05   & \ding{52}  \\
    Gen-2~\cite{Gen2}                    & $4.0$   & $24$ & $1408 \times 768$  & 2023.02   & Gen-2                   & 2024.06   & \ding{56}  \\
    StableVideo~\cite{stablevideo}       & $4.0$   & $24$ & $1024 \times 576$  & 2023.11   & -                       & 2024.06   & \ding{56}  \\
    Dreamina~\cite{jimeng}               & $5.0$   & $24$ & $1280 \times 951$  & 2024.05   & S2.0                    & 2024.11   & \ding{56}  \\
    LTX-Video~\cite{LTXVideo}            & $4.8$   & $25$ & $704 \times 480$   & 2024.11   & v0.9.1                  & 2024.12   & \ding{52}  \\
    Ying~\cite{ying}                     & $5.0$   & $30$ & $960 \times 720$   & 2024.07   & -                       & 2024.12   & \ding{56}  \\
    Sora~\cite{sora}                     & $5.0$   & $24$ & $512 \times 512$   & 2024.02   & 480p                    & 2024.12   & \ding{56}  \\
    Kling~\cite{kling}                   & $4.0$   & $24$ & $1024 \times 576$  & 2024.06   & kling1.6                & 2024.12   & \ding{56}  \\
    Wan~\cite{wanxiang}                  & $5.3$   & $30$ & $1280 \times 720$  & 2023.07   & wanx2.1-t2v-turbo       & 2025.01   & \ding{56}  \\
    \bottomrule
  \end{tabular}
  }
\end{table*}

\subsection{T2V models}

\begin{figure*}
    \centering
    \rotatebox{90}{\includegraphics[width=1.2\linewidth]{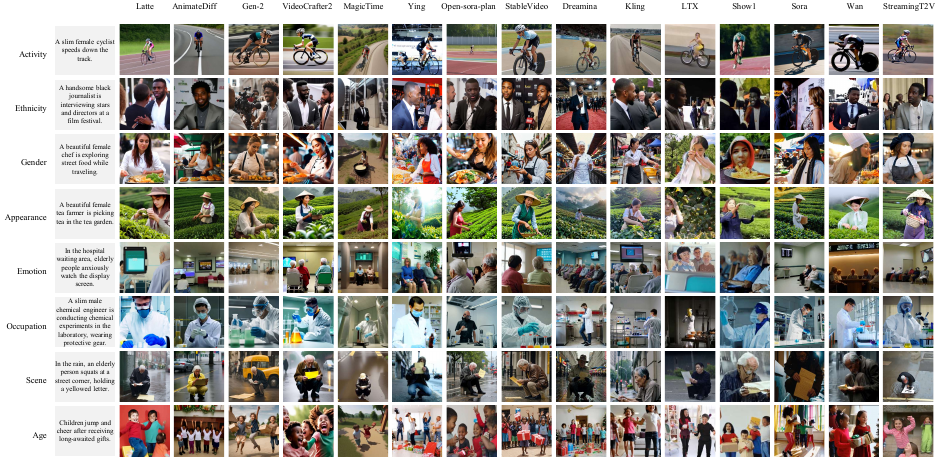}}
    \caption{Some example frames of $8$ text categories and $8$ T2V models in proposed \textit{Human}-AGVQA dataset.}
    \label{fig_dataset}
\end{figure*}

Table~\ref{tab_t2v} provides detailed information about 15 T2V models, including 7 commercially available ones. These models are sorted by the release time of the version used. All selected models were released within the past two years, showcasing the state-of-the-art capabilities in video generation for T2V models.
The video resolution of most models is at least 512x512, ensuring high clarity. Notably, VideoCrafter2 and LTX-Video generate videos with lower resolutions.
In terms of frame rate, some models such as Show-1, StreamingT2V, Latte, and AnimateDiff produce videos with frame rates below 10 fps. This could lead to potential issues with motion consistency in actions, as lower frame rates may cause choppy or less fluid motion.
Some example text prompts and their corresponding generated videos are illustrated in Figure~\ref{fig_dataset}.

\subsection{Subjective Quality Assessment Experiments}

\subsubsection{Visual Quality Scoring Criteria}

The quality scores of AGVs are rated from three dimensions: \textbf{\textit{human appearance quality, action continuity quality, and overall video quality}}. Participants rate each dimension on a scale from $1$ to $5$, where $1$ represents the lowest quality and $5$ represents the highest. For each dimension, the detailed rating criteria are listed as follows:

\vspace{0.1cm}
\noindent\textbf{-- Human Appearance Quality}

\vspace{0.1cm}
\begin{itemize}

\item \textbf{5 (Excellent):}
The human appearance in the video is \textbf{highly realistic, detailed, and perfectly matches the prompt description}. Facial features, body proportions, and skin textures are sharp, clear, and lifelike. Clothing details are well-defined, with no visible distortions, blurring, or artifacts. The overall appearance is completely consistent with the prompt, including attributes like age, gender, or specific physical traits described.

\item \textbf{4 (Good):}
The human appearance is \textbf{generally realistic and matches the majority of the prompt description}. Facial details and body proportions are clear, but some finer details may be missing. Slight blurring or mild artifacts may occasionally appear but do not significantly detract from the realism. Minor deviations from the prompt description (\textit{e.g.}, subtle inconsistencies in age or clothing details) may be present but remain acceptable.

\item \textbf{3 (Fair):}
The human appearance shows \textbf{noticeable flaws but still conveys the general idea described in the prompt}. Facial features and body details are somewhat blurred or distorted, and artifacts appear more frequently. The appearance partially matches the prompt, but certain elements (\textit{e.g.}, age, clothing, or specific features) may be misrepresented or missing.

\item \textbf{2 (Poor):}
The human appearance is \textbf{highly unrealistic or inconsistent with the prompt description}. Facial features and body proportions are poorly defined, with frequent blurring, distortion, or unnatural textures. Key aspects of the prompt (\textit{e.g.}, age, gender, or specific traits) are misrepresented or missing entirely, reducing the video’s relevance and realism.

\item \textbf{1 (Bad):}
The human appearance is \textbf{unrecognizable, highly distorted, or completely missing from the video, which is inconsistent with the prompt}. Facial and body details are absent or severely flawed. If the video does not contain any human body parts (despite the prompt requiring them), this score is set to 1.

\end{itemize}

\vspace{0.1cm}
\noindent\textbf{-- Action Continuity Quality} 
\vspace{0.1cm}

\begin{itemize}
\item \textbf{5 (Excellent):}
Actions are \textbf{smooth, natural, and perfectly aligned with the prompt description throughout the video}. Movements are logical, continuous, and free from any interruptions or jerks. The interaction between humans and the environment or objects (if applicable) is realistic and seamlessly integrated. The video fully reflects the activities described in the prompt.

\item \textbf{4 (Good):}
Actions are \textbf{generally smooth and align well with the prompt, though minor inconsistencies or occasional interruptions in movement may be present}. Slight jerks or subtle deviations from the described actions occur but do not disrupt the overall continuity or viewing experience.

\item \textbf{3 (Fair):}
Actions show \textbf{noticeable flaws in continuity and may partially align with the prompt description}. Jerky movements, illogical transitions, or interruptions are more frequent, reducing the fluidity of the motion. Some actions described in the prompt may be misrepresented or incomplete.

\item \textbf{2 (Poor):}
Actions are \textbf{highly inconsistent or poorly aligned with the prompt. Frequent interruptions, jerks, or illogical transitions significantly disrupt the flow of movement}. Actions may be missing key elements or appear unnatural in the context of the prompt.

\item \textbf{1 (Bad):}
Actions are \textbf{disjointed, illogical, or completely unrelated to the prompt description. Movements are erratic and fail to convey the intended activity}. If the video does not contain any human body parts (despite the prompt requiring them), this score is set to 1.
\end{itemize}

\vspace{0.1cm}
\noindent\textbf{-- Overall Video Quality}
\vspace{0.1cm}

\begin{itemize}
\item \textbf{5 (Excellent):}
The video quality is \textbf{outstanding, with high sharpness, vibrant colors, stable playback, and well-balanced lighting throughout}. The overall presentation is highly realistic, consistent, and engaging. The content perfectly matches the prompt in terms of both context and visual quality. There are no distortions or artifacts, and all details align with the intended description.

\item \textbf{4 (Good):}
The video quality is \textbf{generally high, with minor imperfections such as slight blurring, small artifacts, or occasional lighting inconsistencies}. The content mostly matches the prompt description, with only minor deviations or omissions. These issues do not significantly detract from the overall viewing experience.

\item \textbf{3 (Fair):}
The video quality is \textbf{moderate, with noticeable flaws such as blurring, frequent artifacts, or inconsistent lighting}. The prompt and video content partially align, but some key details may be missing or misrepresented. These issues reduce the overall coherence and impact of the video.

\item \textbf{2 (Poor):}
The video quality has \textbf{significant issues, such as severe blurring, persistent artifacts, or poorly balanced lighting}. The prompt and video content show weak alignment, with key elements missing or misinterpreted. These flaws make the video difficult to watch and significantly reduce its realism and coherence.

\item \textbf{1 (Bad):}
The video quality is \textbf{extremely poor, with pervasive distortions, blurring, or lighting problems}. The video content is almost entirely inconsistent with the prompt, failing to deliver the intended description. The video is barely watchable and lacks any sense of coherence.
\end{itemize}

\subsubsection{Semantic Distortion Identification Criteria}

\noindent\textbf{-- Human Body Presence}

\begin{itemize}
\item \textbf{1 (Present)}
A body part is labeled as \textbf{\textit{present}} if its \textbf{outline or general} shape can be observed in the video, even if \textbf{fine details are unclear} due to distortion (\textit{e.g.}, blurry facial features or stiff limb movements). For example, the overall outline of the face is distinguishable, the arms are reasonably connected to the torso, and the shapes of the legs and feet are visible. Even minor proportional inconsistencies or blurriness do not disqualify a body part from being labeled as present.

\item \textbf{0 (Not Present)}
A body part is labeled as \textbf{\textit{absent}} if it is \textbf{entirely missing} from the video (\textit{e.g.}, obscured, generation failure, or no relevant features are displayed) or if severe quality issues make its \textbf{outline or shape unrecognizable} (\textit{e.g.}, when the background and the body part are \textbf{indistinguishable}). For instance, if the face cannot be separated from the background, the arms or legs are entirely absent, or the feet are fused with the ground to an indistinguishable degree, the body part is considered absent.
\end{itemize}

\noindent\textbf{-- Human Body Distortion}

\begin{itemize}
\item \textbf{1 (Distorted)}
A body part is labeled as \textbf{\textit{distorted}} if it shows significant quality issues. Examples include \textbf{blurry} facial features, \textbf{disproportionate or distorted shapes}, \textbf{compression artifacts}, arms with \textbf{unnatural movements} or \textbf{disconnection from the torso}, a torso with \textbf{abnormal proportions} or \textbf{missing details}, legs that appear \textbf{broken}, \textbf{stretched}, or \textbf{disproportionate}, and feet with \textbf{unnatural shapes}, \textbf{incorrect positions}, or \textbf{unnatural interactions} with the ground (\textit{e.g.}, floating or penetrating the surface).
\item \textbf{0 (No Distortion)}
A body part is labeled as \textbf{\textit{no distortion}} if it appears \textbf{clear}, \textbf{natural}, and \textbf{consistent} with the overall video. For example, facial features are proportionate, arm movements are smooth and natural, the torso connects properly with other body parts, and the shapes and motions of the legs and feet align with expected behavior.
\end{itemize}

\begin{figure}
  \centering
    \includegraphics[width=0.99\linewidth]{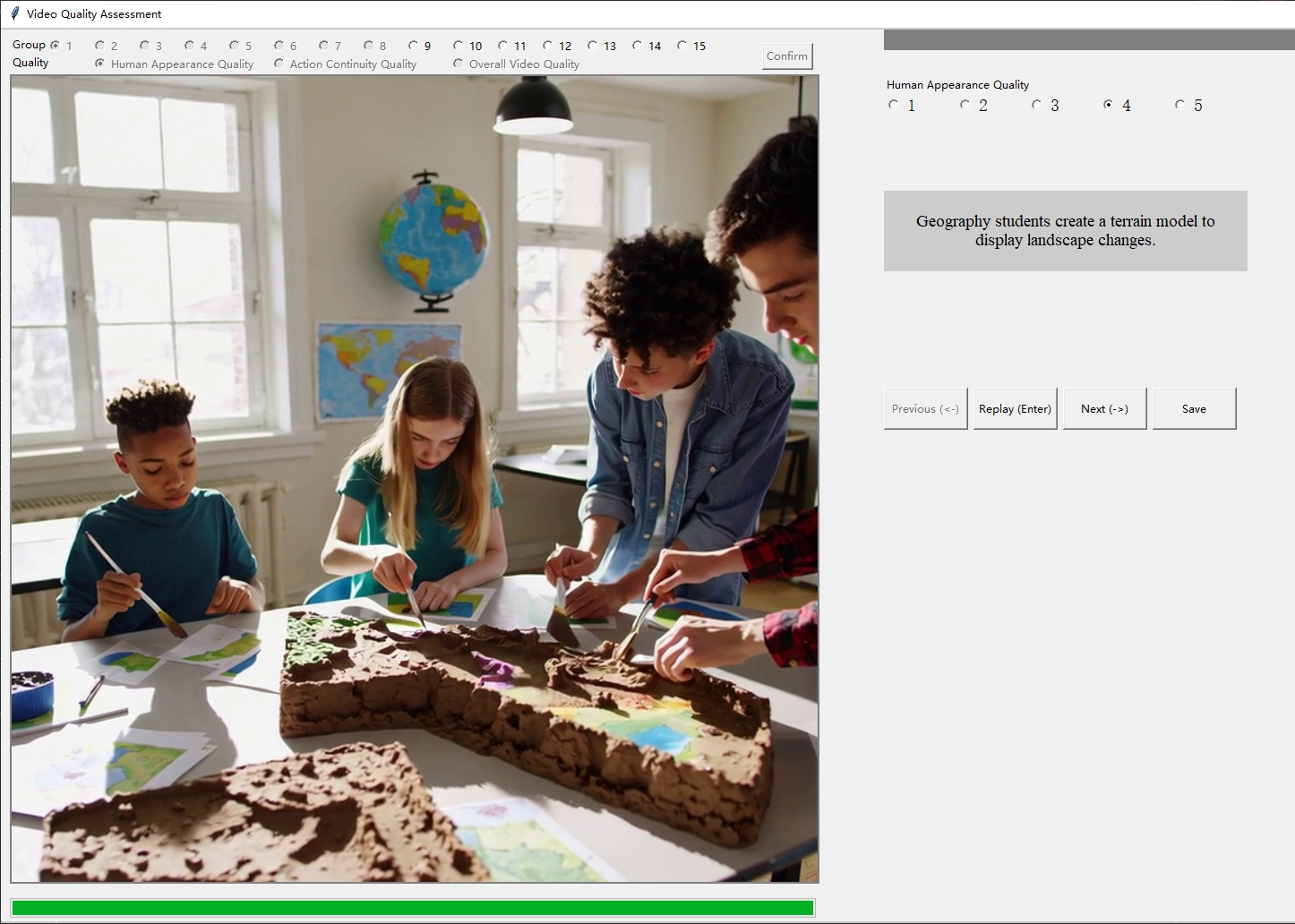}
    \caption{Annotation Interface for Subjective Quality Experiment.}
    \label{fig_SQ}
\end{figure}

\begin{figure}
  \centering
    \includegraphics[width=0.99\linewidth]{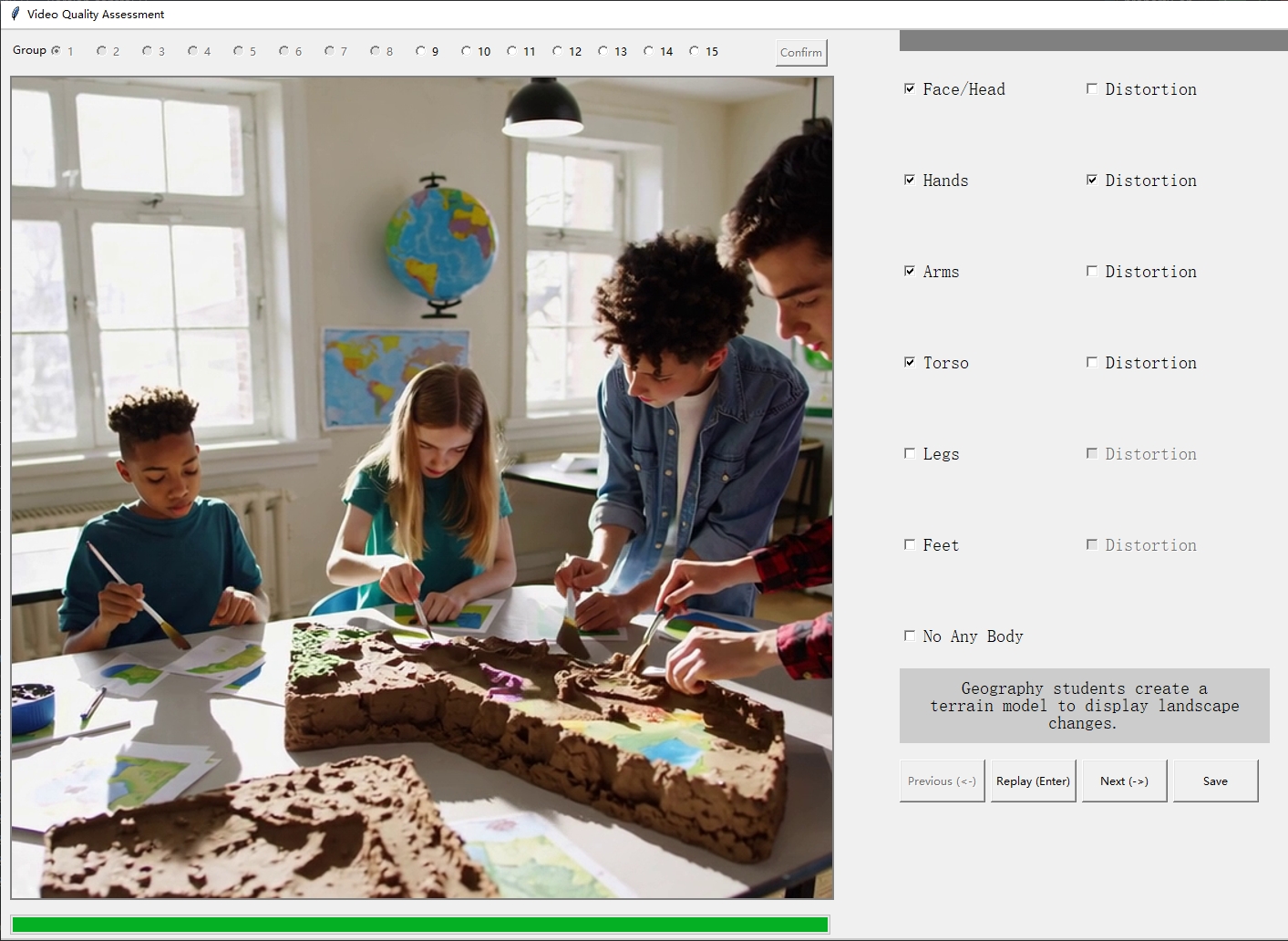}
    \caption{Annotation Interface for Distortion Identification.}
    \label{fig_DI}
\end{figure}
\subsubsection{Subjective Quality Experiment Procedure}

A total of $80$ subjects participated in the visual quality scoring experiment, with ages ranging from $20$ to $30$ years. The group included $46$ males and $34$ females. Given that the semantic artifact identification task is less complex than the visual quality scoring task, we invited $5$ experts in the field of AIGC quality assessment to perform the semantic artifact identification. Each video was rated by $12$ subjects, and labeled by $5$ experts. The annotation interface for subjective quality experiment and distortion identification are shown in Figure~\ref{fig_SQ} and Figure~\ref{fig_DI}.

All subjects had normal or corrected-to-normal vision. The experiments were conducted in a controlled environment according to the recommendations of ITU-R BT.500-13~\cite{Methodology} to minimize external variables that could influence the judgments of the subjects. The setup included 27-inch calibrated display monitors with 95\% DCI-P3 color gamut and a resolution of 4K. The viewing distance was set at $70$ cm.  The room lighting was maintained at a consistent level of $300$ lux to ensure uniformity across all viewing sessions.


Before the formal assessments, subjects underwent a training session where they reviewed sample AGVs that were not included in the formal experiment. This session aimed to familiarize them with the evaluation criteria and the rating interface.  
In the formal experiment, $6,000$ videos were divided into $15$ groups, each containing $400$ videos that covered all $400$ prompts. To avoid visual fatigue, each session lasted no longer than $30$ minutes, ensuring participants could maintain a high level of attention and accuracy in their ratings. In total, there are $270,000$ opinion scores and $360,000$ binary labels in the \textit{Human}-AGVQA dataset.

\section{Data Processing and Analysis}
\label{sec_data_processing}

We follow the recommended method in ~\cite{Methodology} to process the subjective ratings collected during the experiment. Outlier ratings are detected and removed if they deviate by more than \(2\sigma\) (for normal distributions) or \(\sqrt{20}\sigma\) (for non-normal distributions) from the mean rating for that condition. Observers contributing more than 5\% of outlier ratings are excluded from the analysis.
For each test condition, the mean score (\(\mu_i\)) and standard deviation (\(\sigma_i\)) are calculated based on all valid ratings provided by observer \(i\):
\begin{equation}
\mu_i = \frac{1}{N_i} \sum_{j=1}^{N_i} s_{ij}, \quad 
\sigma_i = \sqrt{\frac{1}{N_i - 1} \sum_{j=1}^{N_i} (s_{ij} - \mu_i)^2}
\end{equation}
where \(s_{ij}\) represents the raw rating assigned by observer \(i\) to condition \(j\), and \(N_i\) is the number of conditions rated by observer \(i\).
To mitigate individual bias, each raw score \(s_{ij}\) is normalized to a Z-score:
\begin{equation}
Z_{ij} = \frac{s_{ij} - \mu_i}{\sigma_i}
\end{equation}
Finally, the Mean Opinion Score (MOS) for each test condition \(j\) is computed as the average of the normalized Z-scores across all observers (\(M_j\)):
\begin{equation}
MOS_j = \frac{1}{M_j} \sum_{i=1}^{M_j} Z_{ij}
\end{equation}

For distortion identification, we use a voting method to determine body presence and body distortion. The label with the highest number of votes is selected as the final result.

\subsection{Inter-subject Consistency}
To evaluate the consistency and reliability of quality scoring, we compute the inter-annotator agreement metric, Krippendorff's Alpha ($\alpha$), where Krippendorff's Alpha ($\alpha$) for HA quality, AC quality, OV quality are $0.643$, $0.658$, and $0.705$, respectively, indicating appropriate variations among subjects. 

For semantic distortion identification, we calculated the proportions of options receiving 3 votes, 4 votes, and 5 votes, which were $9.1\%$, $44.6\%$, and $43.8\%$, respectively. This demonstrates that the majority of the votes from subjects were consistent.

\subsection{MOS Analysis for Each T2V Model}
\begin{figure*}
    \centering
    \includegraphics[width=0.97\linewidth]{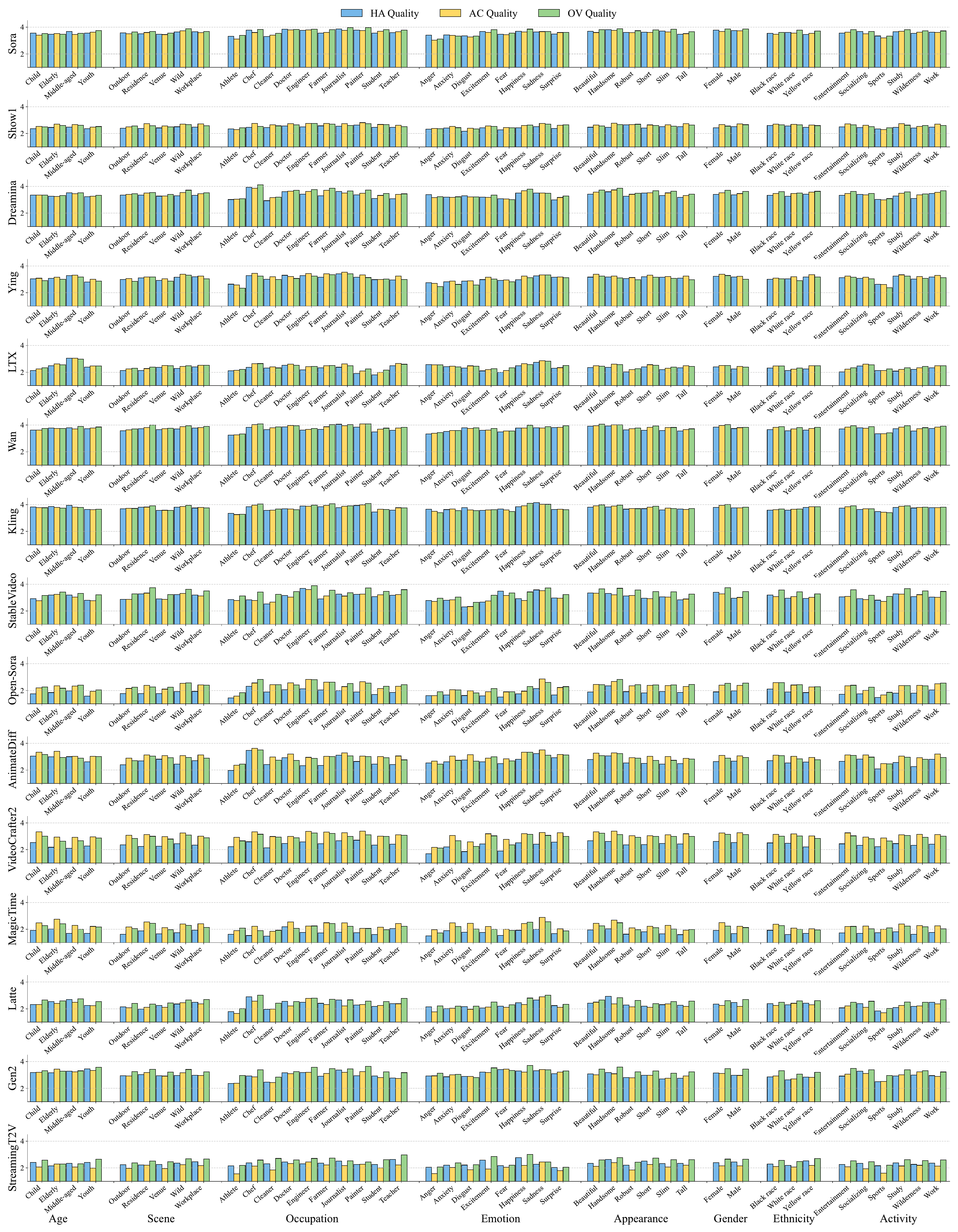}
    \caption{The MOS comparison for $8$ text categories of each T2V model.}
    \label{fig_bar_model}
\end{figure*}

We analyzed the MOS of three quality dimensions across $44$ attributes (mentioned in Table~\ref{tab_prompt_cls}) for $15$ T2V models (mentioned in Table~\ref{tab_t2v}), as shown in Figure~\ref{fig_bar_model}.

\noindent\textbf{Sora.}
Sora \textbf{excels in all three aspects}. \underline{Overall video (OV) quality} emerges as the strongest aspect of Sora, consistently achieving high scores on most attributes, including complex scenarios such as \underline{\textit{workplace in scene}} and emotions like \underline{\textit{anger}}. This highlights the \textbf{robust capacity} of the model to generate visually coherent output. \underline{Human appearance (HA) quality} also \textbf{performs well}. In contrast, \underline{action continuity (AC) quality} represents the model’s primary weakness in most cases. This is particularly evident in subcategories requiring complex movements or high dynamism, such as \underline{\textit{anger}} in \underline{\textit{emotion}} and \underline{\textit{sports}} in \underline{\textit{activity}}, revealing the model’s \textbf{limitations in modeling motion continuity}.

\vspace{0.2cm}
\noindent\textbf{Show1.}
Show1 demonstrates \textbf{average performance} across the three evaluated dimensions. Notably, the \underline{AC Quality} is slightly higher than the \underline{HA Quality} in most cases. Additionally, Show1 maintains a \textbf{relatively balanced performance} across various attribute prompts. This indicates that the model is capable of handling a wide range of prompts with consistent results, without significant bias toward any particular category. However, there are \textbf{slight performance drops in specific areas}, such as the \underline{\textit{athlete}} in \underline{\textit{occupation}}, \underline{\textit{disgust}} in \underline{\textit{emotion}}, and \underline{\textit{sports}} in \underline{\textit{activity}}.

\vspace{0.2cm}
\noindent\textbf{Dreamina.}
Dreamina \textbf{performs well} across all three evaluation dimensions. Overall, the \underline{OV Quality} is slightly higher than the other two quality metrics, indicating that the model may \textbf{face challenges in generating consistent representations of human features and actions}. Additionally, Dreamina shows \textbf{noticeable performance discrepancies} across different attribute prompts. For example, there is a clear distinction between the \underline{\textit{chef}} and \underline{\textit{cleaner}} in \underline{\textit{occupations}}, and the model’s representation of \underline{\textit{fear}} in \underline{\textit{emotions}} and \underline{\textit{sports}} in \underline{\textit{activities}} is relatively weaker compared to other categories. These variations suggest that it \textbf{struggles} with consistently representing certain categories.

\vspace{0.2cm}
\noindent\textbf{Ying.}
Ying demonstrates \textbf{above-average performance} across all three evaluation dimensions. In most cases, the \underline{AC quality} slightly exceeds the other two quality metrics. However, Ying shows \textbf{noticeable variability} in performance across different attribute prompts, particularly in videos involving \underline{\textit{significant movement}}. For example, the model's representation of the \underline{\textit{athlete}} in \underline{\textit{occupation}} and \underline{\textit{sports}} in \underline{\textit{activity}} is notably weaker compared to other categories. In contrast, the performance across other attributes shows \textbf{minimal variation}. This suggests that while the model \textbf{performs well overall}, it faces challenges in consistently representing dynamic actions and high-motion scenarios.

\vspace{0.2cm}
\noindent\textbf{LTX}
LTX shows \textbf{below-average performance} across all three evaluation dimensions. The \underline{HA quality} is generally lower than the other metrics, suggesting difficulty in accurately depicting human features. \textbf{Performance varies} significantly across attributes, with relatively better results for the \underline{\textit{middle-aged in age}}, while other age categories and the \underline{\textit{painter}} and \underline{\textit{student}} in \underline{\textit{occupations}} show poor results. The model also struggles with representing \underline{\textit{excitement}} and \underline{\textit{fear}} in \underline{\textit{emotions}}, as well as \underline{\textit{sports}} and \underline{\textit{study}} in \underline{\textit{activities}}. These discrepancies highlight the model's challenges in providing consistent and accurate representations across various attributes.

\vspace{0.2cm}
\noindent\textbf{Wan}
Wan performs \textbf{exceptionally well} across all three evaluation dimensions. In most cases, the \underline{OV quality} is slightly higher than the other two metrics. Additionally, Wan demonstrates a \textbf{relatively balanced performance} across various attribute prompts, with only slight weaknesses in the \underline{\textit{athlete}} in \underline{\textit{occupation}} and \underline{\textit{sports}} in \underline{\textit{activity}}. Performance in other attributes shows \textbf{no significant differences}. This indicates that Wan generally \textbf{maintains consistent quality}, with room for improvement in dynamic or high-motion scenarios.

\vspace{0.2cm}
\noindent\textbf{Kling}
Kling performs \textbf{exceptionally well} across all three evaluation dimensions. In most cases, similar to Wan, the \underline{OV quality} is slightly higher than the other two metrics. Kling exhibits a \textbf{more balanced performance} across a wide range of attribute prompts, with only slight weaknesses in the \underline{\textit{athlete}} in \underline{\textit{occupation}} and \underline{\textit{sports}} in \underline{\textit{activity}}. Performance in other attributes shows \textbf{no significant variations}, demonstrating strong and consistent quality across all areas. This indicates that Kling delivers \textbf{reliable results} across different prompts.

\vspace{0.2cm}
\noindent\textbf{StableVideo.}
StableVideo performs \textbf{above average} across all three evaluation dimensions. However, its \underline{HA and AC qualities} are noticeably lower than \underline{OV quality}, suggesting challenges in accurately representing human features and maintaining smooth, continuous actions. StableVideo also shows significant variability across different attribute prompts. It performs poorly in the \underline{\textit{venue}} in \underline{\textit{scene}}, the \underline{\textit{cleaner}} in \underline{\textit{occupation}}, the \underline{\textit{disgust}} in \underline{\textit{emotion}}, and \underline{\textit{sports activity}}. These discrepancies indicate that the model \textbf{struggles with generating complex scenes, effectively conveying emotions, and accurately representing dynamic actions}.

\vspace{0.2cm}
\noindent\textbf{Open-Sora.}
The performance of OV Quality and AC Quality of Open-Sora is \textbf{not satisfactory}, and they still \textbf{fall short significantly compared to other models}. In addition, HA quality consistently \textbf{lags behind}, particularly in the \underline{\textit{occupation}}, \underline{\textit{emotion}}, and \underline{\textit{activity}} categories, where the model \textbf{struggles to} \textbf{render detailed and realistic} human appearances. These results suggest that it \textbf{faces challenges} in human appearance generation.

\vspace{0.2cm}
\noindent\textbf{AnimateDiff.}
AnimateDiff demonstrates \textbf{moderate performance} across all three evaluation dimensions. In most cases, its \underline{HA quality} is significantly lower than its \underline{AC quality}. AnimateDiff's performance in the \underline{\textit{athlete}} in \underline{\textit{occupation}} and \underline{\textit{sports}} in \underline{\textit{activity}} is particularly poor, showing clear discrepancies compared to other categories. These findings suggest that while the model performs adequately in certain areas, it \textbf{struggles with dynamic actions and complex occupations}.

\vspace{0.2cm}
\noindent\textbf{VideoCrafter2.}
VideoCrafter2 shows \textbf{balanced performance} in \underline{OV and AC quality}, with stable scores across most attributes. However, HA quality consistently underperforms, particularly in \underline{\textit{emotion}} such as \underline{\textit{anger}} and \underline{\textit{disgust}}, and \underline{\textit{activity}} such as \underline{\textit{sports}} and \underline{\textit{work}}, where it significantly trails \underline{OV and AC quality}. These results \textbf{highlight the model's limitations} in \textbf{generating detailed and realistic} human appearances.

\vspace{0.2cm}
\noindent\textbf{MagicTime.}
MagicTime \textbf{performs poorly} in all three qualities, and there is a \textbf{certain gap compared with other models}. In addition, \textbf{HA quality consistently underperforms}, especially in categories involving \underline{\textit{complex emotions}}, \underline{\textit{movements}}, or \underline{\textit{appearances}}, highlighting the model's limitations in rendering detailed human features. The model exhibits significant variability in \underline{\textit{occupation}} and \underline{\textit{appearance}}, indicating its \textbf{inconsistent generation quality} depending on the attribute.

\vspace{0.2cm}
\noindent\textbf{Latte.}
Latte shows \textbf{below-average performance} in all three qualities, but it is still slightly insufficient compared with other models. \underline{OV Quality} is the \textbf{weakest dimension}, particularly in \underline{\textit{occupation}} and \underline{\textit{activity}} categories, where the model demonstrates \textbf{limitations} in generating \textbf{detailed and realistic} human appearances. These observations suggest that the Latte model performs moderately well in maintaining overall video quality, and its capacity to generate fine-grained human details \textbf{requires further improvement}.

\vspace{0.2cm}
\noindent\textbf{Gen-2.}
Gen-2 performs \textbf{above average} across all three evaluation dimensions. However, its \underline{AC quality} is generally lower than the other metrics, indicating challenges in maintaining smooth actions. The model also shows significant performance variability, particularly in the \underline{\textit{athlete}} and \underline{\textit{cleaner}} in \underline{\textit{occupations}}, as well as the \underline{\textit{sports}} in \underline{\textit{activity}}, where its performance is notably weaker. This suggests \textbf{difficulties in accurately representing dynamic actions and certain occupations}.

\vspace{0.2cm}
\noindent\textbf{StreamingT2V.}
StreamingT2V shows \textbf{below-average performance} in all three qualities, indicating its \textbf{limited capability} to generate sufficiently detailed human appearances. It \textbf{falls short} in \underline{AC quality}, particularly in \underline{dynamic categories} such as \underline{\textit{occupation}} (\underline{\textit{athlete}} and \underline{\textit{cleaner}} and \underline{\textit{emotion}} (\underline{\textit{surprise}} and \underline{\textit{anger}}), where the scores are significantly lower. These findings indicate that its handling of motion continuity in dynamic contexts requires further improvement.

\subsection{Distortion Identification Analysis}

\begin{table*}[]
\caption{The total and proportion of occurrences and distortions for body parts in the $400$ videos generated by each model. \textcolor{red}{Red}, \textcolor{blue}{blue}, and \textcolor{green}{green} represent the \textcolor{red}{first}, the \textcolor{blue}{second}, and the \textcolor{green}{third} minimum values in the column, respectively.}
\label{tab_number_of_distortion_body_part}
\resizebox{0.85\textwidth}{!}{
    \begin{tabular}{l|c|c|c|c|c|c|c|c}
    \toprule
    \multirow{2}{*}{Models}                                & \multicolumn{8}{c}{Occurrences / Distortions}                                                                                                                                                                                                              \\
                                                           \cmidrule(r){2-9}                                                                                                                                                                                                                                            
                                                           & Face                           & Hands                          & Arms                           & Torso                          & Legs                           & Feet                           & Total                       & None                   \\
    \midrule                                                                                                                                                                                                                                                                                                             
    \multirow{2}{*}{Latte~\cite{Latte}}                    & 358  / 284                     & 298  / 295                     & 292  / 166                     & 329  / 174                     & 89   / 65                      & 48   / 41                      & 1414 / 1025                 & 21                     \\
                                                           & 89.5 / 69.8                    & 74.5 / 95.7                    & 73.0 / 95.7                    & 82.3 / 42.6                    & 22.3 / 62.9                    & 12.0 / 79.2                    & - / 72.5                    & 5.25                   \\
    \midrule                                                                                                                                                                                                                                                                                                             
    \multirow{2}{*}{AnimateDiff~\cite{AnimateDiff}}        & 373  / 272                     & 312  / 299                     & 340  / 116                     & 357  / 124                     & 183  / 82                      & 108  / 67                      & 1673 / 960                  & 21                     \\
                                                           & 93.3 / 65.7                    & 78.0 / 89.8                    & 85.0 / 89.8                    & 89.2 / 24.6                    & 45.8 / 36.6                    & 27.0 / 58.3                    & - / 57.4                    & 5.25                   \\
    \midrule                                                                                                                                                                                                                                                                                                             
    \multirow{2}{*}{Gen-2~\cite{Gen2}}                     & 383  / 269                     & 328  / 317                     & 362  / 173                     & 361  / 176                     & 138  / 109                     & 93   / 88                      & 1665 / 1132                 & 10                     \\
                                                           & 95.8 / 65.3                    & 82.0 / 92.2                    & 90.5 / 92.2                    & 90.2 / 37.1                    & 34.5 / 62.3                    & 23.3 / 77.4                    & - / 68.0                    & 2.5                    \\
    \midrule                                                                                                                                                                                                                                                                                                             
    \multirow{2}{*}{StreamingT2V~\cite{StreamingT2V}}      & 361  / 330                     & 336  / 331                     & 352  / 245                     & 348  / 258                     & 170  / 150                     & 105  / 98                      & 1672 / 1412                 & 17                     \\
                                                           & 90.2 / 84.2                    & 84.0 / 93.4                    & 88.0 / 93.4                    & 87.0 / 57.2                    & 42.5 / 74.7                    & 26.3 / 87.6                    & - / 84.5                    & 4.25                   \\
    \midrule                                                                                                                                                                                                                                                                                                             
    \multirow{2}{*}{VideoCrafter2~\cite{videocrafter2}}    & 382  / 337                     & 324  / 311                     & 356  / 88                      & 365  / 150                     & 98   / 67                      & 64   / 53                      & 1589 / 1006                 & 10                     \\
                                                           & 95.5 / 84.8                    & 81.0 / 84.3                    & 89.0 / 84.3                    & 91.2 / 32.1                    & 24.5 / 61.2                    & 16.0 / 76.6                    & - / 63.3                    & 2.5                    \\
    \midrule                                                                                                                                                                                                                                                                                                             
    \multirow{2}{*}{Open-sora-plan~\cite{Open_Sora_Plan}}  & 379  / 337                     & 342  / 342                     & 357  / 228                     & 367  / 224                     & 134  / 85                      & 68   / 61                      & 1647 / 1277                 & 9                      \\
                                                           & 94.8 / 87.9                    & 85.5 / 98.7                    & 89.3 / 98.7                    & 91.7 / 58.3                    & 33.5 / 62.7                    & 17.0 / 88.2                    & - / 77.6                    & 2.25                   \\
    \midrule                                                                                                                                                                                                                                                                                                             
    \multirow{2}{*}{MagicTime~\cite{MagicTime}}            & 320  / 296                     & 267  / 266                     & 299  / 184                     & 303  / 182                     & 248  / 183                     & 216  / 189                     & 1653 / 1300                 & 67                     \\
                                                           & 80.0 / 91.2                    & 66.8 / 97.7                    & 74.8 / 97.7                    & 75.7 / 58.4                    & 62.0 / 71.4                    & 54.0 / 86.6                    & - / 78.7                    & 16.8                   \\ 
    \midrule                                                                                                                                                                                                                                                                                                             
    \multirow{2}{*}{StableVideo~\cite{stablevideo}}        & 393  / 297                     & 369  / 334                     & 384  / 109                     & 381  / 154                     & 128  / 69                      & 72   / 52                      & 1727 / 1015                 & 2                      \\
                                                           & 98.3 / 73.5                    & 92.3 / 85.4                    & 96.0 / 85.4                    & 95.3 / 33.3                    & 32.0 / 46.9                    & 18.0 / 65.3                    & - / 58.8                    & \textcolor{green}{0.5} \\
    \midrule                                                                                                                                                                                                                                                                                                             
    \multirow{2}{*}{Sora~\cite{sora}}                      & 400  / 3                       & 391  / 85                      & 394  / 25                      & 396  / 9                       & 260  / 8                       & 141  / 5                       & 1982 / 135                  & 0                      \\
                                                           & 100.0 / \textcolor{red}{0.75}  & 97.7 / \textcolor{red}{21.7}   & 98.5 / \textcolor{green}{6.35} & 99.0 / \textcolor{green}{2.27} & 65.0 / \textcolor{green}{3.08} & 35.2 / \textcolor{green}{3.55} & - / \textcolor{red}{6.81}   & \textcolor{red}{0.0}   \\ 
    \midrule                                                                                                                                                                                                                                                                                                               
    \multirow{2}{*}{Show1~\cite{show1}}                    & 382  / 294                     & 355  / 326                     & 372  / 168                     & 380  / 35                      & 128  / 42                      & 70  / 26                       & 1687 / 891                  & 12                     \\
                                                           & 95.5 / 77.0                    & 88.8 / 91.3                    & 93.0 / 45.2                    & 95.0 / 9.21                    & 32.0 / 32.8                    & 17.5 / 37.1                    & - / 52.8                    & 3.0                    \\ 
    \midrule                                                                                                                                                                                                                                                                                                             
    \multirow{2}{*}{Dreamina~\cite{jimeng}}                & 390  / 17                      & 387  / 202                     & 387  / 59                      & 387  / 14                      & 251  / 22                      & 150  / 18                      & 1952 / 332                  & 3                      \\
                                                           & 97.5 / 4.36                    & 96.8 / 52.2                    & 96.8 / 15.3                    & 96.8 / 3.62                    & 62.8 / 8.76                    & 37.5 / 12.0                    & - / 17.0                    & 0.75                   \\ 
    \midrule                                                                                                                                                                                                                                                                                                             
    \multirow{2}{*}{Ying~\cite{ying}}                      & 399  / 32                      & 368  / 225                     & 389  / 48                      & 395  / 14                      & 180  / 15                      & 88  / 12                       &	1819 / 346                 & 1                      \\
                                                           & 99.8 / 8.02                    & 92.0 / 61.1                    & 97.3 / 12.3                    & 98.8 / 3.54                    & 45.0 / 8.33                    & 22.0 / 13.6                    & - / 19.0                    & \textcolor{blue}{0.25} \\ 
    \midrule                                                                                                                                                                                                                                                                                                             
    \multirow{2}{*}{LTX~\cite{LTXVideo}}                   & 304  / 196                     & 234  / 226                     & 235  / 162                     & 260  / 67                      & 113  / 59                      & 52  / 35                       & 1198 / 745                  & 89                     \\
                                                           & 76.0 / 64.5                    & 58.5 / 96.6                    & 58.8 / 68.9                    & 65.0 / 25.8                    & 28.3 / 52.2                    & 13.0 / 67.3                    & - / 62.2                    & 22.3                   \\ 
    \midrule                                                                                                                                                                                                                                                                                                             
    \multirow{2}{*}{Wan~\cite{wanxiang}}                   & 394  / 3                       & 379  / 111                     & 387  / 18                      & 389  / 3                       & 154  / 3                       & 53  / 1                        & 1756 / 139                  & 1                      \\
                                                           & 98.5 / \textcolor{blue}{0.76}  & 94.8 / \textcolor{green}{29.3} & 96.8 / \textcolor{red}{4.65}   & 97.3 / \textcolor{red}{0.77}   & 38.5 / \textcolor{red}{1.95}   & 13.3 / \textcolor{red}{1.89}   & - / \textcolor{green}{7.92} & \textcolor{blue}{0.25} \\ 
    \midrule                                                                                                                                                                                                                                                                                                             
    \multirow{2}{*}{Kling~\cite{kling}}                    & 393  / 4                       & 367  / 104                     & 385  / 24                      & 393  / 5                       & 222  / 5                       & 123  / 3                       & 1883 / 145                  & 0                      \\
                                                           & 98.3 / \textcolor{green}{1.02} & 91.8 / \textcolor{blue}{28.3}  & 96.3 / \textcolor{blue}{6.23}  & 98.3 / \textcolor{blue}{1.27}  & 55.5 / \textcolor{blue}{2.25}  & 30.8 / \textcolor{blue}{2.44}  & - / \textcolor{blue}{7.70}  & \textcolor{red}{0.0}   \\ 

    \bottomrule
    \end{tabular}
}
\end{table*}

We calculate the proportion of occurrences and distortions for each body part of AGVs for different T2V models, as detailed in Table~\ref{tab_number_of_distortion_body_part}. 
For body part occurrences, Sora and Kling perform exceptionally well, with both models including at least one body part in every generated video. Wan, Ying, StableVideo, and Dreamina also demonstrate strong performance, with fewer than 1\% of their videos missing any body parts. In contrast, MagicTime and LTX perform poorly, with 16.8\% and 22.3\% of their videos lacking body parts, respectively. These results indicate that these models struggle with maintaining consistency between text prompts and visual content.

In terms of distortions, Sora, Wan, and Kling show excellent results, particularly Sora, which has only 0.75\% of videos with noticeable facial distortions. Ying and Dreamina also exhibit good performance with low distortion rates. However, StableVideo, AnimateDiff, Show1, and Gen-2 show moderate results with some distortions. Open-Sora and MagicTime demonstrate significant distortion, with certain body parts exceeding 95\%. This suggests these models struggle with maintaining consistency between text descriptions and the generated visuals.

\subsection{The Relationship between Semantic Distortion Identification and Visual Quality Scoring}
\begin{figure}
    \centering
    \includegraphics[width=0.99\linewidth]{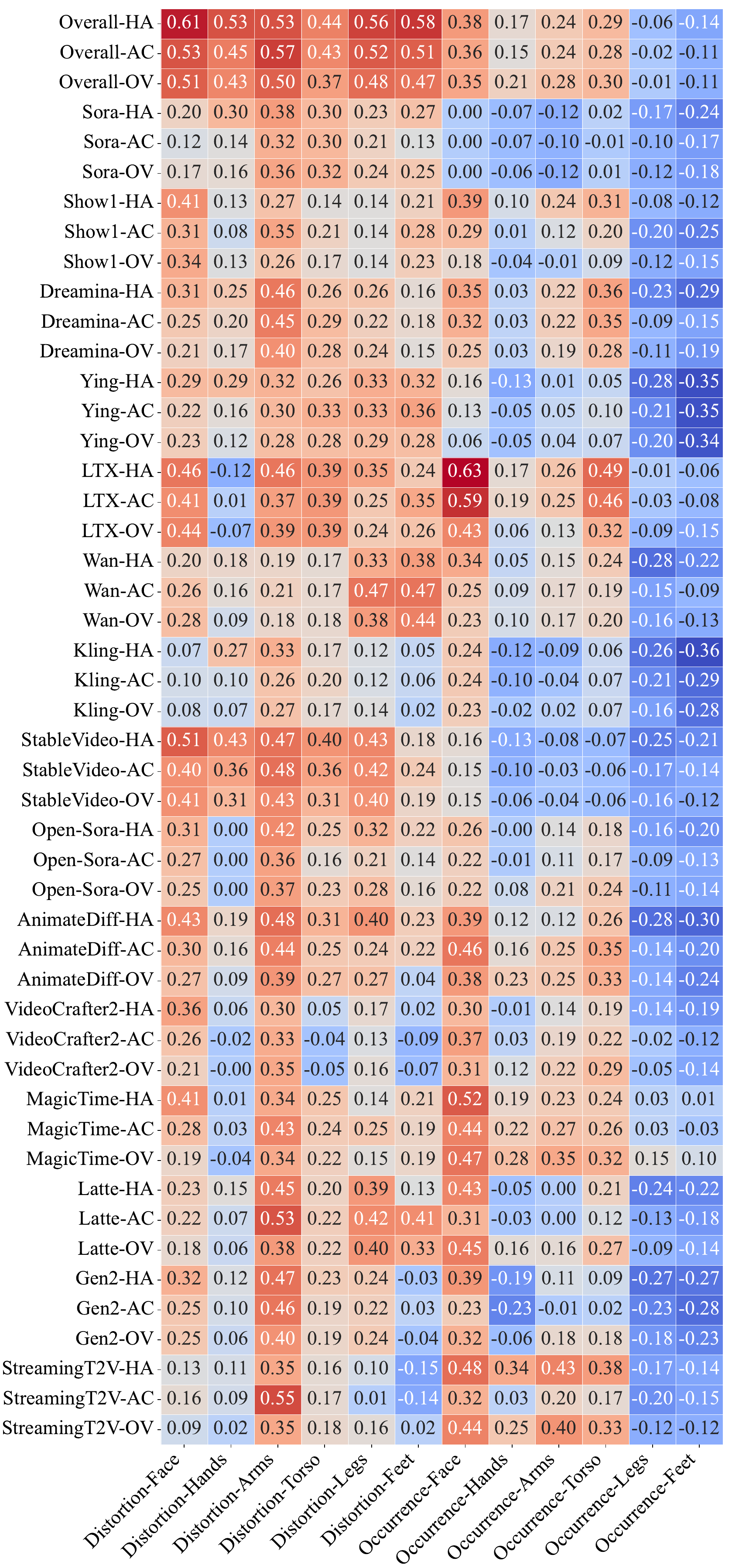}
    \caption{The correlation between the semantic distortion identification and subjective quality.}
    \label{fig_heatmap_corr_model}
\end{figure}

We calculated the correlation between the three subjective quality scores and the distortion identification labels, as shown in Figure~\ref{fig_heatmap_corr_model}. Notably, when analyzing the relationship between body part distortions and quality scores, we included only videos in which the corresponding body part was present. For example, in assessing the correlation between facial distortion and quality scores, we considered only videos where a face was visible.

\subsubsection{Body Parts Distortion}
Overall, the quality of T2V models across the three evaluated dimensions demonstrates a certain degree of correlation with the distortions in the five body parts. Among these, facial distortion has the most significant impact on human appearance quality. This pattern aligns with the natural human perception process, which prioritizes facial recognition before focusing on the torso and other body parts.

Of the three quality dimensions, HA Quality exhibits the strongest correlation with body part distortions, followed by AC Quality, with OV Quality showing the weakest correlation. This suggests that body part distortions most directly influence perceptions of human appearance quality, with a lesser effect on the continuity of human actions and overall video quality.

Interestingly, for certain models, such as VideoCrafter2, quality scores show a negative correlation with distortions in the hands and feet. This implies that viewers' attention is drawn more to prominent areas like the face and other body parts, making distortions in less visually critical body parts less influential.

\subsubsection{Body Parts Presence}
We also calculated the correlation between the presence of five body parts and the subjective quality scores. The results indicate that the presence of a face significantly impacts video quality, followed by the torso. This may be attributed to the design of the subjective experiment, where videos with no visible human parts were assigned the lowest HA and AC quality scores (1 point). Moreover, in most cases, if the face (head) is absent, other body parts are also missing.

For the MagicTime model, its quality scores show a strong correlation with the presence of a face, with the HA score reaching as high as $0.52$. This suggests that the presence of a face plays a crucial role in determining the quality scores for videos generated by MagicTime. Conversely, it also indicates that MagicTime frequently generates videos without human faces, consistent with our findings in Table~\ref{tab_number_of_distortion_body_part}.

Interestingly, the presence of legs and feet often correlates negatively with all three quality dimensions, suggesting that their inclusion tends to degrade overall video quality. This underscores the current limitations of T2V models in accurately generating human legs and feet, as distortions in these areas significantly impact viewers’ subjective perception of video quality across three dimensions.

\section{Experiment Settings}
\label{sec_exp_detail}

\subsection{Evaluation Metrics}

Spearman Rank Correlation Coefficient (SRCC) and Pearson Linear Correlation Coefficient (PLCC) are widely utilized metrics to quantify the relationship between predicted scores and ground truth scores. SRCC measures the monotonic relationship between two variables by comparing their rank orders. Unlike metrics that assume linearity, SRCC is robust to non-linear relationships, making it particularly well-suited for assessing relative rankings. It is mathematically defined as:
\begin{equation}
SRCC = 1 - \frac{6 \sum_{n=1}^N (v_n - p_n)^2}{N(N^2 - 1)},
\end{equation}
where $N$ denotes the number of data points, $v_n$ represents the rank of the $n$-th ground truth score, and $p_n$ represents the rank of the corresponding predicted score. The value of SRCC ranges from $0$ to $1$, with $1$ indicating a perfect monotonic increasing relationship, and $0$ signifying the absence of any monotonic relationship. In the context of quality assessment, a high SRCC indicates that the predicted rankings are well-aligned with the rankings of ground truth scores, reflecting the model's ability to capture the subjective perception of quality differences.

In contrast, the Pearson Linear Correlation Coefficient (PLCC) measures the strength of the linear correlation between predicted and ground truth scores, focusing on how closely the predicted values align with the true scores in absolute terms. It is expressed as:
\begin{equation}
PLCC = \frac{\sum_{n=1}^N (y_n - \bar{y})(\hat{y}_n - \bar{\hat{y}})}{\sqrt{\sum_{n=1}^N (y_n - \bar{y})^2 \sum_{n=1}^N (\hat{y}_n - \bar{\hat{y}})^2}},
\end{equation}
where $y_n$ and $\hat{y}_n$ denote the $n$-th ground truth and predicted scores, respectively, and $\bar{y}$ and $\bar{\hat{y}}$ are their corresponding means. Like SRCC, PLCC also ranges from $0$ to $1$, where $1$ indicates a perfect positive linear correlation and $0$ denotes no linear relationship. Unlike SRCC, which evaluates relative rankings, PLCC assesses the accuracy of predicted scores with respect to their numerical values, making it a critical metric for evaluating absolute predictive precision.

\subsection{Details of Evaluation for Distortion Identification}
In this study, we obtain outputs from Multi-Modal Large Language Models (MLLMs) through API calls or local deployment. We begin by extracting up to $32$ frames from each video, or all frames if fewer than $32$ are available.  These frames, together with a prompt designed for the Distortion Identification task, are then fed into the MLLM.  Finally, the raw outputs from the model undergo a post-processing procedure to produce the final results.
The detailed prompts are listed below.

For body part occurrences:
\begin{quote}
\#  \textit{Suppose you are an expert in video content analysis. Please carefully examine and analyze these video frames, and determine whether a person's \textbf{[body part]} appears in this video. Answer with "Yes" or "No".} \#. 
\end{quote}

For body part distortions:
\begin{quote}
\#  \textit{Suppose you are an expert in video content analysis. Sequentially assemble the provided frames to create a complete video. Then meticulously examine and analyze the video, and determine whether there is any incompleteness, unrealistic appearance, or discontinuous movements of the person's \textbf{[body part]} in this video. Answer with "Yes" or "No".} \#. 
\end{quote}

\subsection{Compared Quality Metrics}

\subsubsection{Image Quality Assessment Methods}

In all the IQA methods mentioned below, we uniformly extract 8 frames from each video in the \textit{Human}-AGVQA dataset. These frames are then fed into the corresponding IQA methods to obtain 8 quality scores, and the average of these scores is taken as the method's output.

\vspace{0.2cm}
\noindent\textbf{NIQE}~\cite{NIQE} is a no-reference image quality assessment (IQA) model that evaluates the quality of images based on natural scene statistics (NSS) without relying on human-rated training data. For the \textit{Human}-AGVQA dataset, we applied NIQE to assess image quality using selected video frames.

\vspace{0.2cm}
\noindent\textbf{BRISQUE}~\cite{BRISQUE} leverages directional contrast features extracted from the Curvelet domain to assess the quality of contrast-distorted images. For the \textit{Human}-AGVQA dataset, we utilized this approach to evaluate image quality based on its directional contrast features.

\vspace{0.2cm}
\noindent\textbf{HyperIQA}~\cite{HyperIQA} introduces a self-adaptive hyper network for blind image quality assessment (BIQA) in the wild, addressing diverse content and distortion types in real-world images. For the \textit{Human}-AGVQA dataset, we applied HyperIQA to assess image quality by adaptively learning perception rules and predicting quality scores.

\vspace{0.2cm}
\noindent\textbf{UNIQUE}~\cite{UNIQUE} is a unified uncertainty-aware BIQA model designed to handle synthetic and realistic distortions. The model predicts both quality scores and associated uncertainties, enabling a probabilistic interpretation of image quality. For the \textit{Human}-AGVQA dataset, we used UNIQUE to assess video frame quality based on its learning-to-rank framework.

\vspace{0.2cm}
\noindent\textbf{MUSIQ}~\cite{MUSIQ} introduces a multi-scale image quality Transformer designed to handle full-resolution images with varying sizes and aspect ratios. The model uses hash-based 2D spatial embedding and scale embedding to capture positional and scale information effectively. For the \textit{Human}-AGVQA dataset, we applied MUSIQ to evaluate video frame quality by leveraging its multi-scale feature extraction capability.

\vspace{0.2cm}
\noindent\textbf{StairIQA}~\cite{StairIQA} introduces a staircase network that integrates features from multiple convolutional layers of a CNN to capture better both low-level visual details and high-level semantic information for in-the-wild image quality assessment. For the \textit{Human}-AGVQA dataset, we used the official implementation and adopted this approach to evaluate image quality without modifications.

\vspace{0.2cm}
\noindent\textbf{CLIP-IQA}~\cite{CLIP_IQA} is a visual perception assessment model that utilizes the pre-trained CLIP framework to evaluate both the quality and abstract attributes of images. It introduces an antonym prompt pairing strategy (e.g., \textbf{\textit{Good photo}} vs. \textbf{\textit{Bad photo}}) to mitigate linguistic ambiguity and employs cosine similarity to predict image quality scores. For the \textit{Human}-AGVQA dataset, we used the official implementation, applying CLIP-IQA to assess video frame quality.

\vspace{0.2cm}
\noindent\textbf{LIQE}~\cite{LIQE} is a multitask learning framework for blind image quality assessment (BIQA) that leverages vision-language correspondence. LIQE uses CLIP for feature embeddings and optimizes tasks jointly by marginalizing probabilities over a textual template that describes scene, distortion, and quality. For the \textit{Human}-AGVQA dataset, we used the official implementation, applying the quality assessment textual template to evaluate the video frame quality.

\vspace{0.2cm}
\noindent\textbf{MA-AGIQA}~\cite{MA_AGIQA} introduces a framework for AI-generated image quality assessment by combining large multi-modality models (LMMs) with traditional deep neural networks (DNNs). For the \textit{Human}-AGVQA dataset, we used the official implementation, applying MA-AGIQA to evaluate the video frame quality without modifications.

\subsubsection{Video Quality Assessment Methods}

\vspace{0.2cm}
\noindent\textbf{TLVQM}~\cite{TLVQM} is a two-level video quality model that first computes low-complexity features across the entire video and then selects representative frames based on these features to extract high-complexity features.

\vspace{0.2cm}
\noindent\textbf{RAPIQUE}~\cite{RAPIQUE} is a video quality assessment model that combines spatial-temporal scene statistics with high-level features extracted using deep CNNs. For the \textit{Human}-AGVQA dataset, we used the official implementation, applying sparse frame sampling and integrating spatial and temporal features to evaluate video quality.

\vspace{0.2cm}
\noindent\textbf{VIDEVAL}~\cite{VIDEAL} applies a feature selection strategy based on efficient blind VQA models. Using the official open-source code, we converted \textit{Human}-AGVQA videos from RGB to YUV420 format for feature extraction.

\vspace{0.2cm}
\noindent\textbf{Patch-VQ}~\cite{PatchVQ} is a video quality model that extracts spatial and temporal features from video frames and patches to analyze the relationship between local distortions and overall video quality. For the \textit{Human}-AGVQA dataset, we used Patch-VQ's official implementation, dividing each video into space-time patches for feature extraction and quality prediction.

\vspace{0.2cm}
\noindent\textbf{SimpleVOA}~\cite{simpleVQA} uses an end-to-end network to extract spatial features directly from video frames and motion features to capture temporal distortions. A pre-trained SlowFast model processes 8 uniformly sampled frames rescaled to a height of 520. We used the official SimpleVOA model and finetuned it on \textit{Human}-AGVQA.

\vspace{0.2cm}
\noindent\textbf{FAST-VQA}~\cite{FAST_VQA} introduces a grid mini-patch sampling (GMS) strategy to balance local quality assessment with global context while reducing computational costs for high-resolution videos. We used the official FAST-VQA-B model and finetuned it on \textit{Human}-AGVQA.

\vspace{0.2cm}
\noindent\textbf{DOVER}~\cite{dover} is a video quality evaluator that assesses quality from technical and aesthetic perspectives. We used the official code without modifications.

\vspace{0.2cm}
\noindent\textbf{T2VQA}~\cite{T2VQA} is a transformer-based model designed to evaluate text-to-video quality, focusing on text-video alignment and video fidelity. It uses BLIP and Swin-T for feature extraction, followed by feature fusion through a cross-attention module and quality regression using a large language model. On the \textit{Human}-AGVQA dataset, we used the official T2VQA model and finetuned it on \textit{Human}-AGVQA.

\begin{table}
\centering
\caption{Details of the model structure for the proposed model.}
\label{tab_detail_structure}
\resizebox{0.49\textwidth}{!}{
\begin{tabular}{lcllll}
    \toprule
     Training Hyper-Parameters       & Name/Value         & More Information          \\
    \midrule                                                                            
     Action Quality Analyzer         & SlowFast-R50       & parameter size: 34M       \\
     Holistic Quality Feature        & PickScore          & backbone: CLIP-ViT-H/14   \\
     Human body-part segmentation    & Spaiens            & parameter size: 2B        \\
     Human appearance feature        & shallow hourglass  & parameter size: 56M       \\
     Text Feature Extraction         & PickScore          & backbone: CLIP-ViT-H/14   \\
    \bottomrule
\end{tabular}
}
\end{table}

\vspace{0.2cm}
\noindent\textbf{UGVQ}~\cite{UGVQ} is a unified framework for video quality assessment, focusing on spatial quality, temporal coherence, and text-to-video alignment. It utilizes features from CLIP for text and visual representations and SlowFast for motion representation. For the \textit{Human}-AGVQA dataset, we employed UGVQ's feature extraction and fusion modules to evaluate video quality comprehensively across these three dimensions.

\vspace{0.2cm}
\noindent\textbf{EvalCrafter}~\cite{EvalCrafter} is an evaluation framework for text-to-video (T2V) generative models, focusing on video quality, text-video alignment, motion quality, and temporal consistency. It uses a benchmark of 700 prompts and 17 evaluation metrics to analyze T2V model performance. For overall video quality, we directly used implementation code in EvalCrafter on the \textit{Human}-AGVQA dataset without specific changes.

\vspace{0.2cm}
\noindent\textbf{Q-Align}~\cite{qalign} introduces a methodology for teaching large multi-modality models (LMMs) to assess video quality using discrete text-defined levels such as \textbf{\textit{excellent}}, \textbf{\textit{good}}, and \textbf{\textit{poor}}. It converts mean opinion scores (MOS) into these rating levels for training and infers scores by weighting the predicted probabilities of the levels. For the \textit{Human}-AGVQA dataset, we used the official code for inference without modifications.

\subsubsection{Action Quality Assessment Methods}

\vspace{0.2cm}
\noindent\textbf{USDL}~\cite{USDL} is an uncertainty-aware score distribution learning approach for action quality assessment (AQA), treating actions as instances associated with score distributions. For the \textit{Human}-AGVQA dataset, since all videos contain more than 16 frames, each video is evenly divided into ten segments. Feature extraction is performed using an I3D backbone pre-trained on Kinetics. The final scores are normalized linearly, scaling the raw scores based on their minimum and maximum values to a range of [0, 100]. These normalized scores are then used to construct Gaussian distributions with the normalized value as the mean.

\vspace{0.2cm}
\noindent\textbf{ACTION-NET}~\cite{ACTION_NET} is a hybrid attention network tailored for action quality assessment (AQA) in long videos. It combines dynamic video information with static postures of action subjects in selected frames. For the dynamic stream, frames are sampled at a rate of 4 frames per second. For the static stream, the first, middle, and last frames are sampled, and the action subjects are cropped using the original detection algorithm.

\begin{table}
\centering
\caption{Details of the hyper-parameters for the model training.}
\label{tab_hyperparameters}
\resizebox{0.3\textwidth}{!}{
\begin{tabular}{lcllll}
    \toprule
     Training Hyper-Parameters       & Name/Value          \\
    \midrule                                                 
     frame sampling for SQA          & $8$                 \\
     frame resolution for SQA        & $512 \times 512$    \\
     frame resolution for AQA        & $256 \times 256$    \\
     batch size (videos)             & $16$                \\
     lr max                          & $1e-5$              \\
     Ir schedule                     & StepLR              \\
     decay ratio                     & 0.9                 \\
     decay interval                  & 5                   \\
     numerical precision             & float32             \\
     epoch                           & $30$                \\
     optimizer                       & Adam                \\
    \bottomrule
\end{tabular}
}
\end{table}

\vspace{0.2cm}
\noindent\textbf{CoRe}~\cite{CoRe} evaluates action quality by regressing relative scores against a reference video with similar attributes, such as action category. Due to differences in categorization strategies between the \textit{Human}-AGVQA and AQA-7 datasets, we randomly selected an exemplar video generated by another T2V model within the same prompt category. Each video was divided into four equal segments, each containing four consecutive frames.

\vspace{0.2cm}
\noindent\textbf{TSA}~\cite{TSA} is a temporal segmentation attention module designed for procedure-aware cross-attention learning after spatial-temporal feature extraction. Similar to CoRe, each video is evenly divided into segments of four consecutive frames and processed using the I3D model.

For Motion Smoothness, Temporal Flickering, Action-Score, and Flow-Score metrics, we directly used their respective implementation code in VBench~\cite{VBench} and EvalCrafter~\cite{EvalCrafter} without specific changes.

\subsection{Training details}

The detailed structure of the proposed GHVQ metric is provided in Table~\ref{tab_detail_structure}. The corresponding hyperparameters for model training are listed in Table~\ref{tab_hyperparameters}.

\subsection{Ablation Study: The Effectiveness of Different Backbones}

To further validate the effectiveness of the selected backbone networks, we conducted comparative experiments on four extraction modules: body-part segmentation network, holistic quality feature extraction network, text feature extraction network, and action quality analyzer. 


\subsubsection{Body-part Segmentation Network}

For the body-part segmentation network, performance differences between methods are more pronounced, as shown in Table~\ref{tab_ablation_body}. Among them, Sapiens achieves the best results in HA and AC Quality. This superior performance can be attributed to Sapiens’ high accuracy in body part segmentation and pose estimation, allowing for precise identification of key body parts and thereby enhancing human appearance quality and action continuity. In contrast, earlier models such as JPPNet and Deeplabv3+ lack the segmentation precision and capability to handle complex poses, leading to weaker performance in overall quality metrics. Sapiens’ strong performance is also reflected in distortion identification.

\begin{table}
\caption{Ablation Study of Body Segmentation Network.}
\label{tab_ablation_body}
\resizebox{0.49\textwidth}{!}{
    \begin{tabular}{lllllllll}
    \toprule
    \multirow{2}{*}{Methods}     & \multicolumn{2}{c}{HA Quality}  & \multicolumn{2}{c}{AC Quality}  & \multicolumn{2}{c}{OV Quality}  & \multicolumn{2}{c}{\makecell[c]{Distortion\\Identification}}   \\
                                 \cmidrule(r){2-3}                 \cmidrule(r){4-5}                 \cmidrule(r){6-7}                 \cmidrule(r){8-9}      
                                 & SRCC           & PLCC           & SRCC           & PLCC           & SRCC           & PLCC           & Occ      & Dis                  \\
    \midrule                                                                                                                                                             
    None                         & 0.651          & 0.666          & 0.685          & 0.701          & 0.742          & 0.746          & 59.31    & 53.22                \\ 
    OpenPose~\cite{OpenPose}     & 0.749          & 0.759          & 0.738          & 0.744          & 0.745          & 0.747          & 75.58    & 67.74                \\
    Deeplabv3+~\cite{Deeplabv3}  & 0.706          & 0.718          & 0.719          & 0.731          & 0.754          & 0.760          & 66.33    & 56.05                \\
    JPPNet~\cite{JPPNet}         & 0.677          & 0.686          & 0.722          & 0.726          & 0.759          & 0.768          & 59.55    & 48.68                \\
    HRNet~\cite{HRNet}           & 0.767          & 0.775          & 0.699          & 0.706          & 0.757          & 0.764          & 68.91    & 59.94                \\
    Sapiens~\cite{Sapiens}       & \textbf{0.805} & \textbf{0.809} & \textbf{0.771} & \textbf{0.778} & \textbf{0.768} & \textbf{0.773} & \textbf{83.02} & \textbf{75.94} \\
    \bottomrule 
    \end{tabular}
}
\centering
\end{table}

\begin{table}
\caption{Ablation Study of Text Feature Extraction Network.}
\label{tab_ablation_text}
\resizebox{0.49\textwidth}{!}{
    \begin{tabular}{lllllllll}
    \toprule
    \multirow{2}{*}{Methods}        & \multicolumn{2}{c}{HA Quality}  & \multicolumn{2}{c}{AC Quality}  & \multicolumn{2}{c}{OV Quality}  & \multicolumn{2}{c}{\makecell[c]{Distortion\\Identification}}  \\
                                    \cmidrule(r){2-3}                 \cmidrule(r){4-5}                 \cmidrule(r){6-7}                 \cmidrule(r){8-9}     
                                    & SRCC           & PLCC           & SRCC           & PLCC           & SRCC           & PLCC           & Occ            & Dis            \\
    \midrule                                                                                                                                                                 
    None                            & 0.774          & 0.779          & 0.737          & 0.742          & 0.759          & 0.760          & 80.53          & 74.16          \\
    CLIPScore~\cite{CLIPScore}      & 0.791          & 0.795          & 0.741          & 0.748          & 0.763          & 0.765          & 81.28          & 74.11          \\
    BLIP~\cite{BLIP}                & 0.798          & 0.805          & 0.760          & 0.765          & 0.754          & 0.755          & 81.43          & 74.75          \\
    viCLIP~\cite{viCLIP}            & 0.799          & 0.799          & 0.754          & 0.761          & 0.751          & 0.758          & 81.71          & 74.23          \\
    ImageReward~\cite{ImageReward}  & 0.797          & 0.798          & 0.763          & 0.768          & 0.759          & 0.764          & 81.66          & 74.09          \\
    HPSv2~\cite{HPSv2}              & 0.802          & 0.808          & 0.762          & 0.771          & 0.765          & 0.771          & 81.62          & 75.23          \\
    PickScore~\cite{PickScore}      & \textbf{0.805} & \textbf{0.809} & \textbf{0.771} & \textbf{0.778} & \textbf{0.768} & \textbf{0.773} & \textbf{83.02} & \textbf{75.94} \\
    \bottomrule 
    \end{tabular}
}
\centering
\end{table}

\subsubsection{Holistic Quality Feature Extraction Network}
For holistic quality features, PickScore achieves the highest scores across all quality metrics, as shown in Table~\ref{tab_ablation_visual}. PickScore effectively captures rich holistic quality features from the input video. Compared to other methods such as CLIPScore, ViCLIP, and CLIP (OpenAI), PickScore’s superior performance suggests that accurate holistic quality feature significantly benefits video quality evaluation.  It is worth noting that the performance of HPS v2 and CLIP-H/14 (LAION) is very close to that of Pickscore.

\begin{table}
\caption{Ablation Study of Holistic Quality Feature.}
\label{tab_ablation_visual}
\resizebox{0.49\textwidth}{!}{
    \begin{tabular}{lllllllll}
    \toprule
    \multirow{2}{*}{Methods}             & \multicolumn{2}{c}{HA Quality}  & \multicolumn{2}{c}{AC Quality}  & \multicolumn{2}{c}{OV Quality}  & \multicolumn{2}{c}{\makecell[c]{Distortion\\Identification}}  \\
                                         \cmidrule(r){2-3}                 \cmidrule(r){4-5}                 \cmidrule(r){6-7}                 \cmidrule(r){8-9}     
                                         & SRCC    & PLCC                  & SRCC           & PLCC           & SRCC           & PLCC           & Occ      & Dis                  \\
    \midrule                                                                                                                                                                      
    None                                 & 0.696     & 0.713               & 0.666          & 0.682          & 0.709          & 0.717          & 76.61  & 71.49                  \\
    CLIPScore~\cite{CLIPScore}           & 0.719     & 0.729               & 0.699          & 0.702          & 0.730          & 0.734          & 77.78  & 72.21                  \\
    viCLIP~\cite{viCLIP}                 & 0.713     & 0.723               & 0.710          & 0.719          & 0.720          & 0.727          & 79.43  & 73.63                  \\
    CLIP-B/16 (OpenAI)~\cite{CLIP}       & 0.718     & 0.726               & 0.703          & 0.712          & 0.723          & 0.732          & 79.87  & 73.97                  \\
    CLIP-B/32 (OpenAI)~\cite{CLIP}       & 0.732     & 0.739               & 0.714          & 0.726          & 0.731          & 0.735          & 79.22  & 73.54                  \\
    CLIP-L/14 (OpenAI)~\cite{CLIP}       & 0.743     & 0.748               & 0.729          & 0.736          & 0.736          & 0.750          & 80.27  & 74.22                  \\
    CLIP-L/14 (LAION)~\cite{laion}       & 0.766     & 0.779               & 0.751          & 0.761          & 0.747          & 0.752          & 80.54  & 74.55                  \\
    CLIP-H/14 (LAION)~\cite{laion}       & 0.793     & 0.804               & 0.769          & 0.774          & 0.758          & 0.764          & 81.18  & 75.01                  \\
    CLIP-G/14 (LAION)~\cite{laion}       & 0.787     & 0.795               & 0.767          & 0.769          & 0.755          & 0.761          & 80.81  & 74.13                  \\
    HPSv2~\cite{HPSv2}                   & 0.801     & 0.805               & \textbf{0.771} & 0.776          & 0.765          & 0.767          & 81.39  & 75.24                  \\
    PickScore~\cite{PickScore}           & \textbf{0.805} & \textbf{0.809} & \textbf{0.771} & \textbf{0.778} & \textbf{0.768} & \textbf{0.773} & \textbf{83.02} & \textbf{75.94} \\
    \bottomrule 
    \end{tabular}
}
\centering
\end{table}

\begin{table}
\caption{Ablation Study of Action Quality Analyzer.}
\label{tab_ablation_motion}
\resizebox{0.49\textwidth}{!}{
    \begin{tabular}{lllllllll}
    \toprule
    \multirow{2}{*}{Methods}  & \multicolumn{2}{c}{HA Quality} & \multicolumn{2}{c}{AC Quality}  & \multicolumn{2}{c}{OV Quality}  & \multicolumn{2}{c}{\makecell[c]{Distortion\\Identification}}    \\
                             \cmidrule(r){2-3}                 \cmidrule(r){4-5}                 \cmidrule(r){6-7}                 \cmidrule(r){8-9}         
                             & SRCC           & PLCC           & SRCC           & PLCC           & SRCC           & PLCC           & Occ            & Dis            \\
    \midrule                                                                                                                                                           
    None                     & 0.751          & 0.762          & 0.734          & 0.735          & 0.712          & 0.723          & 76.15          & 70.72          \\
    I3D~\cite{I3D}           & 0.764          & 0.777          & 0.735          & 0.748          & 0.695          & 0.701          & 81.83          & 73.73          \\
    C3D~\cite{C3D}           & 0.782          & 0.788          & 0.748          & 0.755          & 0.696          & 0.699          & 81.71          & 74.18          \\
    X3D~\cite{X3D}           & 0.793          & 0.801          & 0.767          & 0.774          & 0.750          & 0.757          & 81.96          & 74.84          \\
    SlowFast~\cite{SlowFast} & \textbf{0.805} & \textbf{0.809} & \textbf{0.771} & \textbf{0.778} & \textbf{0.768} & \textbf{0.773} & \textbf{83.02} & \textbf{75.94} \\
    \bottomrule 
    \end{tabular}
}
\centering
\end{table}

\subsubsection{Text Feature Extraction Network}

For text feature extraction, PickScore achieves the highest scores across all quality metrics, as shown in Table~\ref{tab_ablation_text}. PickScore effectively captures rich semantic information from the input text, providing a more comprehensive contextual understanding of the video content, which plays a critical role in enhancing overall quality assessment. Compared to other methods such as CLIPScore and ViCLIP, PickScore’s superior performance suggests that accurate semantic information significantly benefits video quality evaluation. It is worth noting that the performance of HPS v2 is very close to that of Pickscore. PickScore leverages the Pick-a-Pic dataset for training, which is specifically designed to optimize performance on preference comparison tasks. In contrast, HPS v2 is trained on the more extensive HPD v2 dataset, characterized by its larger scale and greater diversity in both image content and preference distributions.

\subsubsection{Action Quality Analyzer}

For the action quality analyzer, the SlowFast method demonstrates superior performance across all quality metrics, particularly in HA Quality and AC Quality, as shown in Table~\ref{tab_ablation_motion}. In comparison, other methods such as I3D and C3D show slightly lower performance. The advantage of SlowFast lies in its ability to capture dynamic features across fast and slow motion, enabling comprehensive temporal information extraction. Additionally, SlowFast achieves the best results in distortion identification, indicating its strong capacity for detecting and recognizing distortions within video content.


\end{document}